\documentclass[a4paper,fleqn]{cas-sc}

\usepackage{float}
\usepackage{graphicx}

\usepackage[para]{threeparttable}
\usepackage{multicol}
\usepackage{longtable}
\usepackage{array}
\usepackage{rotating}

\usepackage[numbers]{natbib}
\usepackage{subfig}
\usepackage{caption}
\usepackage[para]{threeparttable}
\usepackage{multicol}
\usepackage{enumitem}

\def\tsc#1{\csdef{#1}{\textsc{\lowercase{#1}}\xspace}}
\tsc{WGM}
\tsc{QE}

\begin{document}
\let\WriteBookmarks\relax
\def\floatpagepagefraction{1}
\def\textpagefraction{.001}

\shorttitle{Person Re-id: A Retrospective on Domain Specific Open Challenges and Future Trends}    

\shortauthors{A. Zahra et.al.}

\title [mode = title]{Person Re-identification: A Retrospective on Domain Specific Open Challenges and Future Trends}  



%

\author[1]{Asmat Zahra}
\affiliation[1]{organization={National University of Sciences and
              Technology (NUST)},
            city={Islamabad},
            postcode={44000}, 
            country={Pakistan}}
            
\author[1]{Nazia Perwaiz}

\author[2]{Muhammad Shahzad}
\affiliation[2]{organization={Technical University of Munich},
            addressline={D-80333 Munich}, 
            country={Germany}}
  
\author[1,3]{Muhammad Moazam Fraz}[orcid=0000-0003-0495-463X]
\cormark[1]
\ead{moazam.fraz@seecs.edu.pk}
\affiliation[3]{organization={The Alan Turing Institute},
            addressline={British Library, 96 Euston Road}, 
            city={London NW1 2DB},
            country={United Kingdom}}



\begin{abstract}
Person re-identification (Re-ID) is one of the primary components of an automated visual surveillance system. It aims to automatically identify/search persons in a multi-camera network having non-overlapping field-of-views. Owing to its potential in various applications and research significance, a plethora of deep learning based re-Id approaches have been proposed in the recent years. However, there exist several vision related challenges, e.g., occlusion, pose scale \& viewpoint variance, background clutter, person misalignment and cross-domain generalization across camera modalities, which makes the problem of re-Id still far from being solved. Majority of the proposed approaches directly or indirectly aim to solve one or multiple of these existing challenges. In this context, a comprehensive review of current re-ID approaches in solving theses challenges is needed to analyze and focus on particular aspects for further advancements. At present, such a focused review does not exist and henceforth in this paper, we have presented a systematic challenge-specific literature survey of 230+ papers between the years of 2015-21. For the first time a survey of this type have been presented where the person re-Id approaches are reviewed in such solution-oriented perspective. Moreover, we have presented several diversified prominent developing trends in the respective research domain which will provide a visionary perspective regarding ongoing person re-Id research and eventually help to develop practical real world solutions.
\end{abstract}




\maketitle


\section{Introduction}
\label{sec:introduction}

In recent years, person re-identification has received much interest owing to its widespread application prospects in numerous fields including intelligent video surveillance \cite{wang2015zero},  robotics \cite{wang2019real} and human-computer interaction \cite{wang2016human} \emph{etc}. Specifically, it is one of the fundamental components of an automated visual surveillance system where for public safety and security in a smart environment, an individual person may be automatically identified and tracked in videos (or images) acquired through multiple non-overlapping cameras installed on public places like airports, banks, cantonments, parks, streets, educational institutes etc. Since it is simply not feasible to rely on manual human intervention to identify a person of interest in huge amount of video data collected on daily basis, therefore a plethora of approaches have been proposed by vision researchers that aim to automate this highly challenging problem. 

Methodologically, person re-id refers to identifying and tracking a person in multi-network non-overlapping cameras installed in indoor and outdoor environments. Given an image of a person captured from one camera, the task of person re-id is to identify this person from a pre-stored gallery set captured by other multiple cameras. 
 

Despite growing trend in the number of publications appearing in top venues achieving increasingly higher accuracy on the existing benchmark datasets, the problem is still far from being solved to be translate into real world settings. This can be attributed to the number of challenges (e.g., occlusion, variations in person pose, viewpoint variations, misalignment, poor resolution \emph{etc}.) that makes the problem extremely hard and needs to be resolved to bridge the performance gap between research (benchmark specific) and real-world environments.

\subsection{Scope/Objective of the Review}
\label{subsec:objective}

\begin{figure*}
    \begin{center}
    \includegraphics[width=\linewidth]{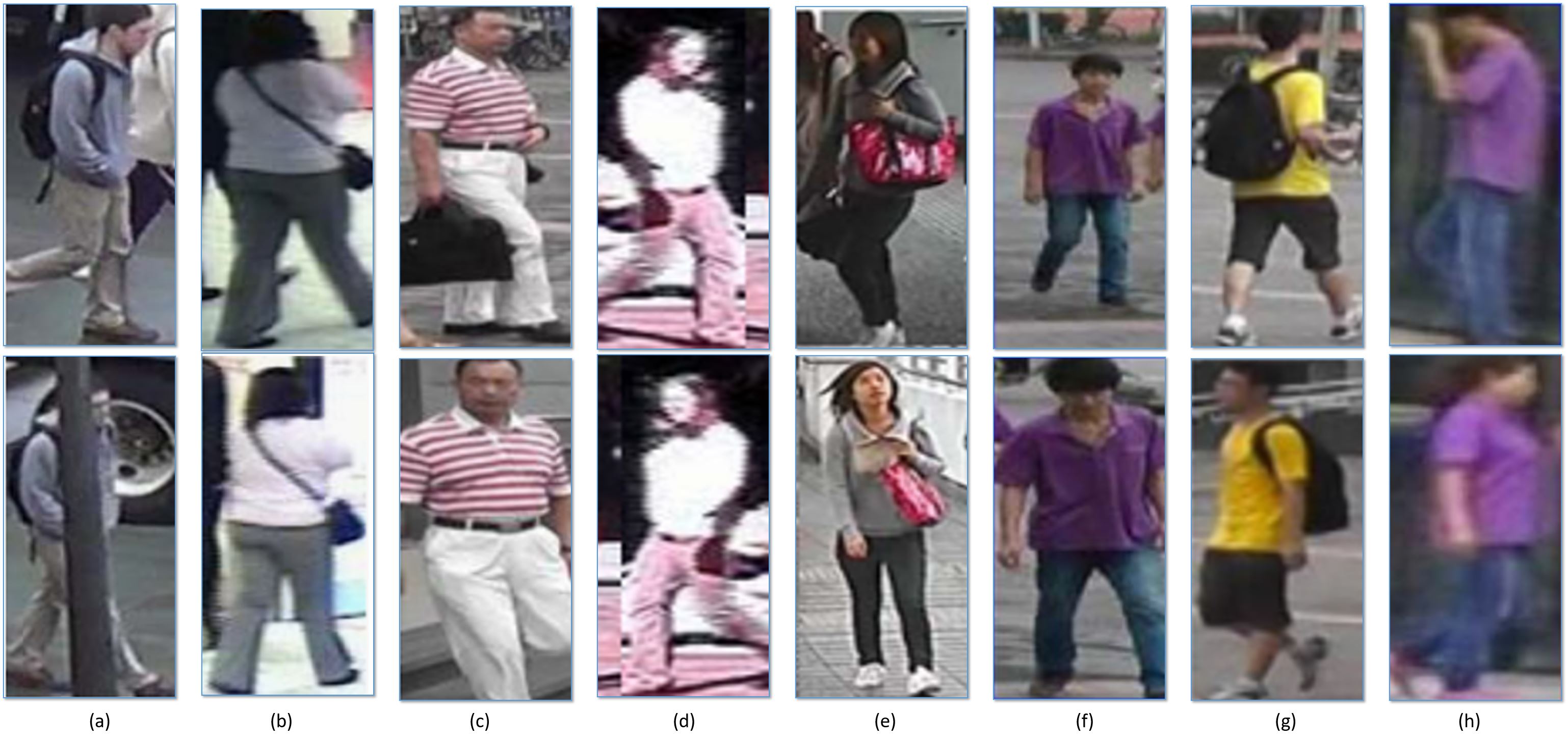}
    \caption{Graphical view of enlisted challenges.(From left to right) (a) Occlusion, (b) Illumination variance, (c) Pose variance, (d) Background clutter, (e) Misalignment, (f) Scale difference, (g) Viewpoint variance and (h) Low Resolution}
    \label{fig:challenges}
    \end{center}
\end{figure*}

In this paper, we have targeted the most popular challenges in person re-id to perform systematic challenge-wise review of the published approaches. In this context, we have collected papers from top conferences and journals for the years from 2015 to 2021. The progress in papers addressing each challenge and its influence on published results is comprehensively reviewed. The specific challenges in re-id that are mainly considered in this review include: occlusion, pose variance, background clutter, misalignment, scale difference, illumination variance, viewpoint variance, low resolution and cross-domain or generalization. Particularly, the proposed review makes many-fold contributions. For instance, we provide in-depth analysis on impact of most popular re-id challenges by discussing the work on each challenge in top computer vision conferences and journals. This provides insights on complexities that arise due to each challenge in the whole re-id process. Moreover, based on reviewed progress, the best performing architectures achieving state-of-the-art (SOTA) results on each challenge (as shown in the Fig. \ref{fig:challenges}) are highlighted and critically analyzed. Furthermore, we attempt to make future directions for researchers by comprehensively reviewing publications relevant to each challenge and discuss the limitations and benefits to lessen the gap between close-world and real-world implementations. Lastly, in addition to reporting trends and highlighting interesting approaches, we distil our analysis into few recommendations in the hope of fostering reproducible and efficient research in the field. For the readers from any of these scenario, this survey also present comprehensive information of how the challenges have been addressed in past and how various components of deep learning (DL) can be utilized to contribute in improving the person re-id considering the influence of each individual challenge.

\subsection{Comparison with Previous Reviews}
\label{subsec:Comparison}

There already exist few review articles on person re-id \cite{zheng2016person, wang2018survey, wu2019deep, ye2020deep, leng2019survey}. Each one of them focus on different aspect of the re-id problem. For instance, in \cite{zheng2016person} both the hand-crafted and deep learning approaches based on image and video data have been reviewed. Similarly, advantages and disadvantages of the traditional and deep learning based approaches are critically analyzed in \cite{wang2018survey}. \cite{leng2019survey} focused on application-driven methods that are designed for specific applications and defined as generalized open-world re-id. In \cite{wu2019deep} six different learning methods including identification, verification, distance metric learning, part-based, video-based and data augmentation based deep models are comprehensively reviewed. Recently \cite{ye2020deep} person re-id is reviewed using open and closed world setting while keeping challenges of re-id perspectives in view. These different perspectives for close world setting includes feature representation, metric learning and ranking optimization. For open world setting, heterogeneity, end-to-end, semi or unsupervised learning, robust model learning with noisy data annotations and open-set person re-id are the considered perspectives.

\begin{figure*} [t]
    \begin{center}
    \includegraphics[width=\linewidth]{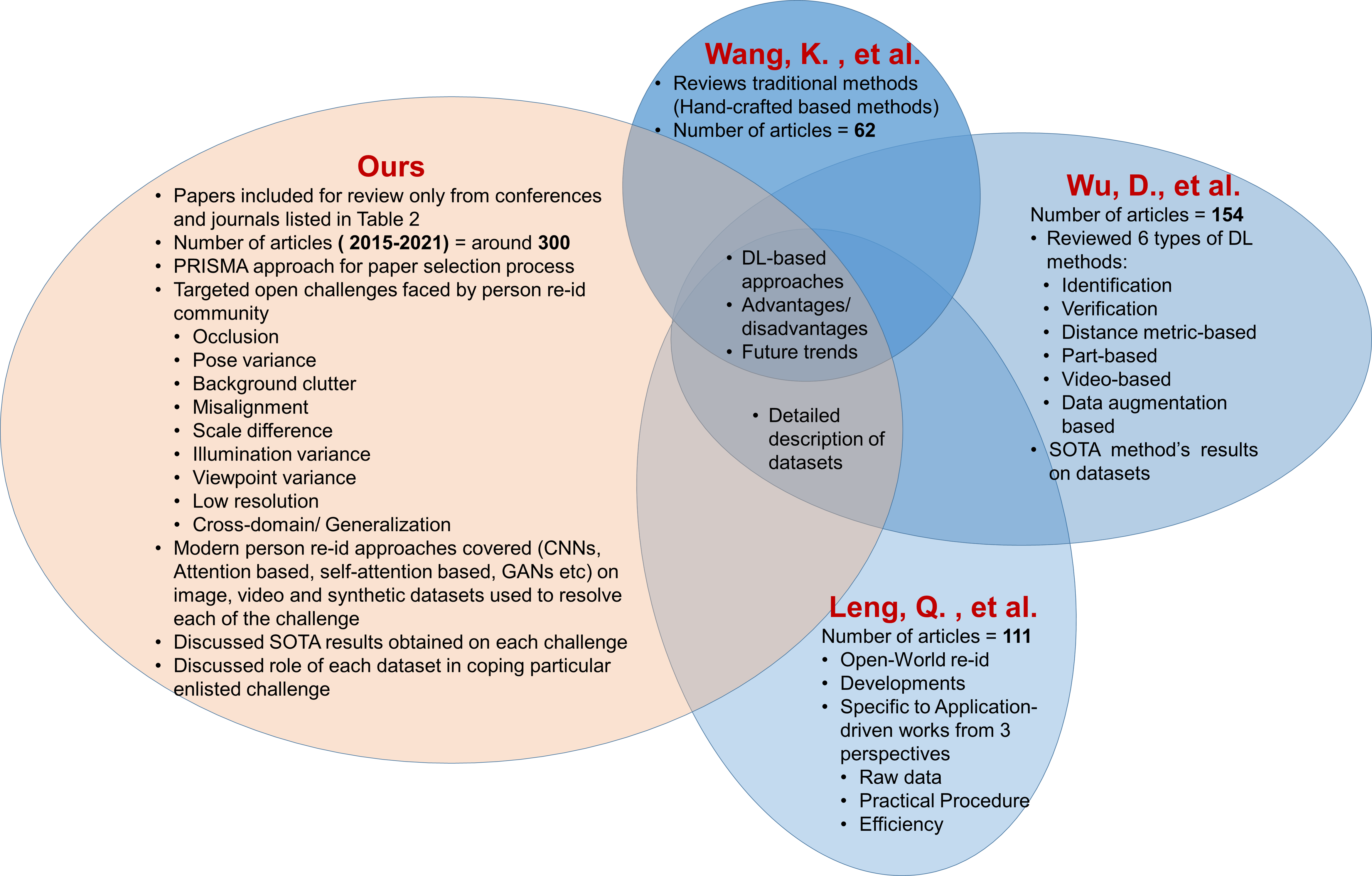}
    \caption{Comparison with recent published reviews on person re-id}
    \label{fig:gapfromprevstudies}
    \end{center}
\end{figure*}

All the aforementioned reviews have comprehensively provided the short-comings and benefits of considered methods and settings. Moreover, they have also provided the insightful future directions. However, none of them have systematically reviewed the influence of popular challenges on performance results and the role of datasets in resolving these challenges. Specifically, no review exists that comprehensively addressed the challenges (mentioned in section \ref{sec:introduction}) of the person re-id and their proposed DL based solutions. From 2015 to 2021 numerous articles have been published and each of these addressed the person re-id challenge in a specific way. Some of challenges are open to the re-id world and still not addressed properly. This motivated us to write this survey article which have comprehensively reviewed how all open challenges are addressed in past and how are results getting improved with respect to each challenge using DL based methods. The difference of our survey from the existing surveys can be visualized in Fig. \ref{fig:gapfromprevstudies}. 

\section{Survey Methodology}
\label{sec:organization}
This survey paper is organized in six sections, Section \ref{sec:introduction} introduces the domain, discusses the scope, objective and rationale of this paper, and highlights main contribution of this review by providing a detailed comparison with recent survey articles on person re-id. Section \ref{sec:organization} illustrates the data collection methodology used for selection the articles include din this review. A comparative account on the publicly available datasets and performance  metrics used to report result of person re-id methodologies is given in Section \ref{sec:datasetsAndPerformanceEvaluation}. Deep learing based approaches are used to solve person re-id challenges in the recent years. For the readers from inter disciplinary domain the deep learning approaches are briefly described in section \ref{sec:deeplearning}. In Section \ref{sec:challenges}, the open challenges in person re-id are discussed and the methodologies proposed to address these challenges are critically analyzed.
Section \ref{sec:discussion} overviews the impact of challenges on results. Role of datasets in resolving the identified challenges along with limitation and benefits. And then it concludes the paper with possible future directions.
\begin{figure}[hbt!]
\begin{center}
    \includegraphics[width=0.65\linewidth, height=\linewidth] {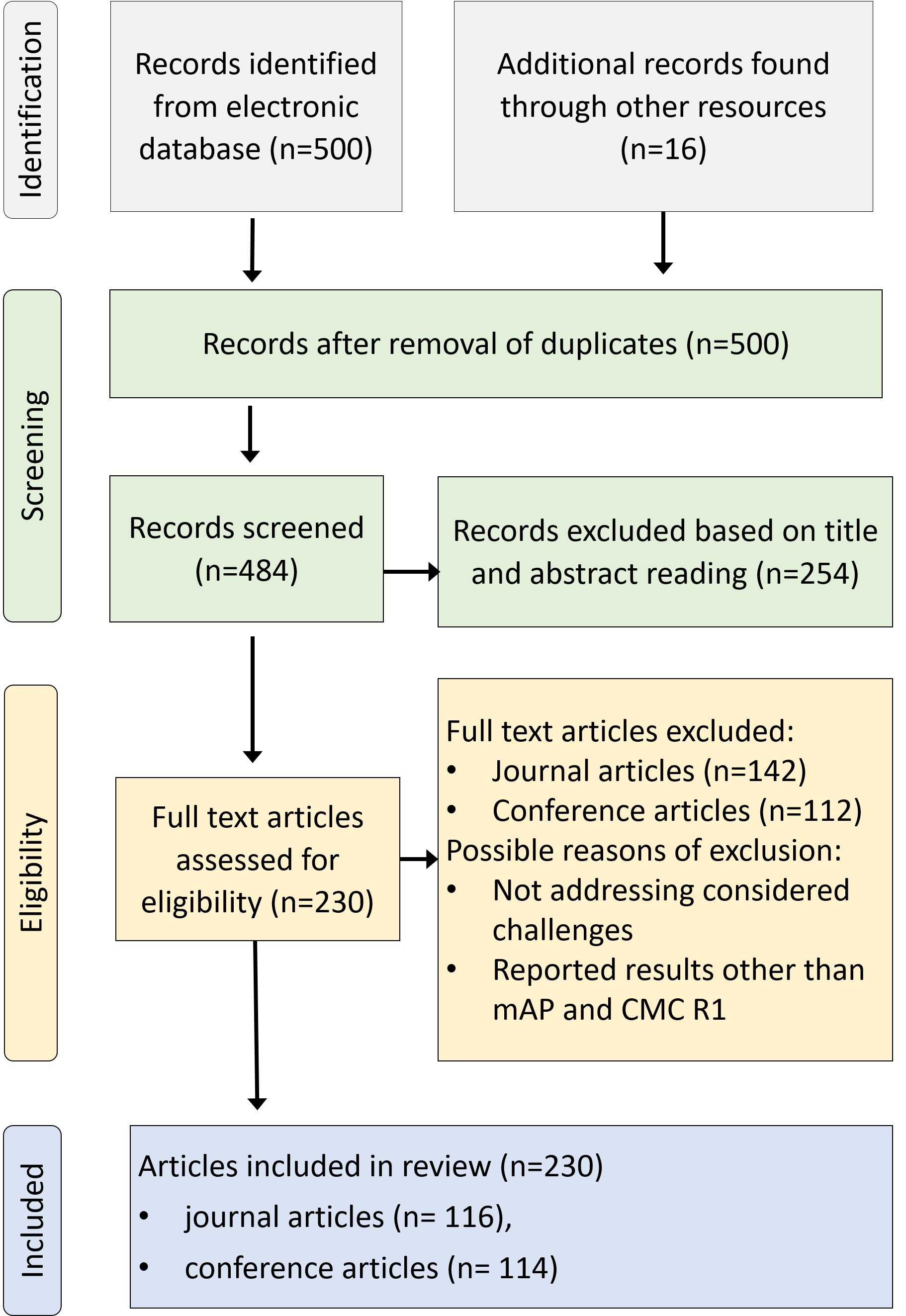}
    \caption{Summary of paper selection process (PRISMA).}
    \label{fig:summarypaperselection}
\end{center}
\end{figure}

\subsection{Study Selection}
\label{subsec:studyselection}
For study selection we used PRISMA (Preferred Reporting Items for Systematic Reviews and Meta-Analyses) \cite{PRISMAwebsite}. Although PRISMA is used primarily in medical domain but because of its vast benefits we have used it for our review as well and obtained valuable results. Fig. \ref{fig:summarypaperselection} shows the summary of paper selection process. As per PRISMA guidlines, our selection of articles for this review consists of two stages. In first stage all the articles that did not meet the eligibility criteria were excluded. While in the second stage we have studied full text reports to find the relevant articles. After shortlisting all the articles, we have excluded those paper which have reported results only through graphs and on such performance measure and datasets which were not so widely adopted by research community. In complex scenario where paper selection become difficult due to some ambiguity a discussion with senior member is organized to reach mutual final decision.
\begin{center}
    \begin{table}
    \caption{Inclusion and exclusion criteria of selected articles}
    \label{table:inclusionExclusionCriteria}
    \begin{tabular}{m{18em} m{18em}}
        \hline\noalign{\smallskip}
        Inclusion criteria & Exclusion criteria \\
        \hline\noalign{\smallskip}\hline\noalign{\smallskip}
        -Articles that address the identified challenges enlisted in table \ref{table:SummaryOfDataItems} & -Articles in which     qualitative evaluation of results are missing\\
        -Provide detailed summary of proposed architecture including training parameters & -Papers that does not address the identified challenges enlisted in table \ref{table:SummaryOfDataItems} \\
        -Articles that are based on deep learning techniques & -Survey papers \\
        -Articles that are published in journals and conferences enlisted in  table \ref{table:summaryOfArticlesCollected} between January 2015 to October 2021  & -Reported results on metrics other than Rank-1 and CMC\\
            & -Papers that have used datasets that were not so widely adopted by research community \\
        \hline\noalign{\smallskip}
    \end{tabular}
    \end{table}
\end{center}

\subsection{Data Extraction Methods}
\label{subsec:methods}
We have collected an initial list of articles from Springer, Google Scholar, IEEE Xplore and Elsevier.
We have considered top seven journals and three conferences held from 2015 to 2021 for review.
We have used the following terms (or matching to these) to search relevant articles: 
    \begin{enumerate}[label=(\alph*)]
        
            \item Person Re-Identification
            \item Deep learning
            \item Supervised person re-id
            \item Semi-supervised person re-id
            \item Unsupervised person re-id
            \item Pose variations
            \item Body misalignment
            \item Attention based approach
            \item Camera View(Viewpoint)
            \item End-to-end learning
        
    \end{enumerate}
Search results are further enhanced by combining the mentioned terms using logical operators in a way like: ('a') AND ('b' OR 'c' OR 'd' OR 'e' OR 'f' OR 'g' OR 'h' OR 'i' OR 'j') .The papers were then excluded or included based on the criteria listed in table \ref{table:inclusionExclusionCriteria}.
We first read the title of article for final selection. In case title does not clearly fall in our inclusion and exclusion criteria then we also read the abstract and conclusion section of the article as well. After that we start full reading for data collection. And articles that does not match with our criteria were not include in this review.

\begin{table} [hbt!]
\caption{Summary of data-items extracted from each article.}
\label{table:SummaryOfDataItems}
    \begin{tabular}{m{7em} m{30em}}
    \hline\noalign{\smallskip}
    Category & Data Item \& Description  \\
    \hline\noalign{\smallskip}\hline\noalign{\smallskip}
    Origin of article & Conference/ Journal Name \\
     & Publication year \&  venue \\
     & Article Title  \\
     & Image/Video dataset or both\\
     & CNN, ViT, GCN, GAN based approaches used in the paper\\
    \hline\noalign{\smallskip}
    Challenges & Occlusion : Blockage or hiding of target person in image \\
     & Pose Variation : Particular person appeared in different positions\\
     & Background clutter : A pattern present in background resembles with pattern of person's wearing in image \\
     & Body Misalignment : Person not aligned according to viewing angle in image \\
     & Scale : Person appearing in different sizes in an image \\
     & Illumination : Light variations in an image \\
     & Viewpoint variation : Change in position of capturing camera \\
     & Resolution : Clarity/detail in input image \\
     & Cross-domain /Generalization : Images belong to multiple domains\\
     \hline\noalign{\smallskip}
     Proposed Methodology & Description of proposed method in article\\
    \hline\noalign{\smallskip}
    Implementation Details & Implementation Framework \& Platform used\\
       & Base Model : DL baseline model used as backbone \\
      & Training approach of model \\
      & Take single/multiple images as query \\
      & Data-set used in pre-trained model \\
      & Batch Normalization Scheme used \\
      & Batch size considered for training \\
      & Type of pooling used in the article \\
      & Learning rate-decay \\
      & Data Augmentation technique used \\
    \hline\noalign{\smallskip}
    Reproducible & Code Availability for public use or not\\
    \hline\noalign{\smallskip}
    Results & Performance Metric used for reporting the results\\
    \hline\noalign{\smallskip}
    Data-sets & Name(s) of data-sets used \\
    \hline\noalign{\smallskip}
    \end{tabular}
\end{table}

\subsection{Data Synthesis}
\label{subsec:datasynthesis}
To make our review more useful in a sense that other researcher can contribute into it in future for the purpose for extension in review in multiple perspective, a data extraction sheet was developed that describe the multiple rel event data items to be extracted from the articles. Around 30 data items were used for metadata extraction from each article. These data items were classified in seven categories: Origin of the article, The challenges addressed, Details on the proposed methodology, Implementation detail, Reproduciblity and code availability, The performance metrics and the Datasets used. Table \ref{table:SummaryOfDataItems} shows the category wise distribution and description of each of the data item. The results will be stored in a spreadsheet which will be made public for interested researchers intended to extract more information or analyse a different perspective.

\begin{figure*}
    \begin{center}
    \includegraphics[width=0.9\textwidth, height=12cm]{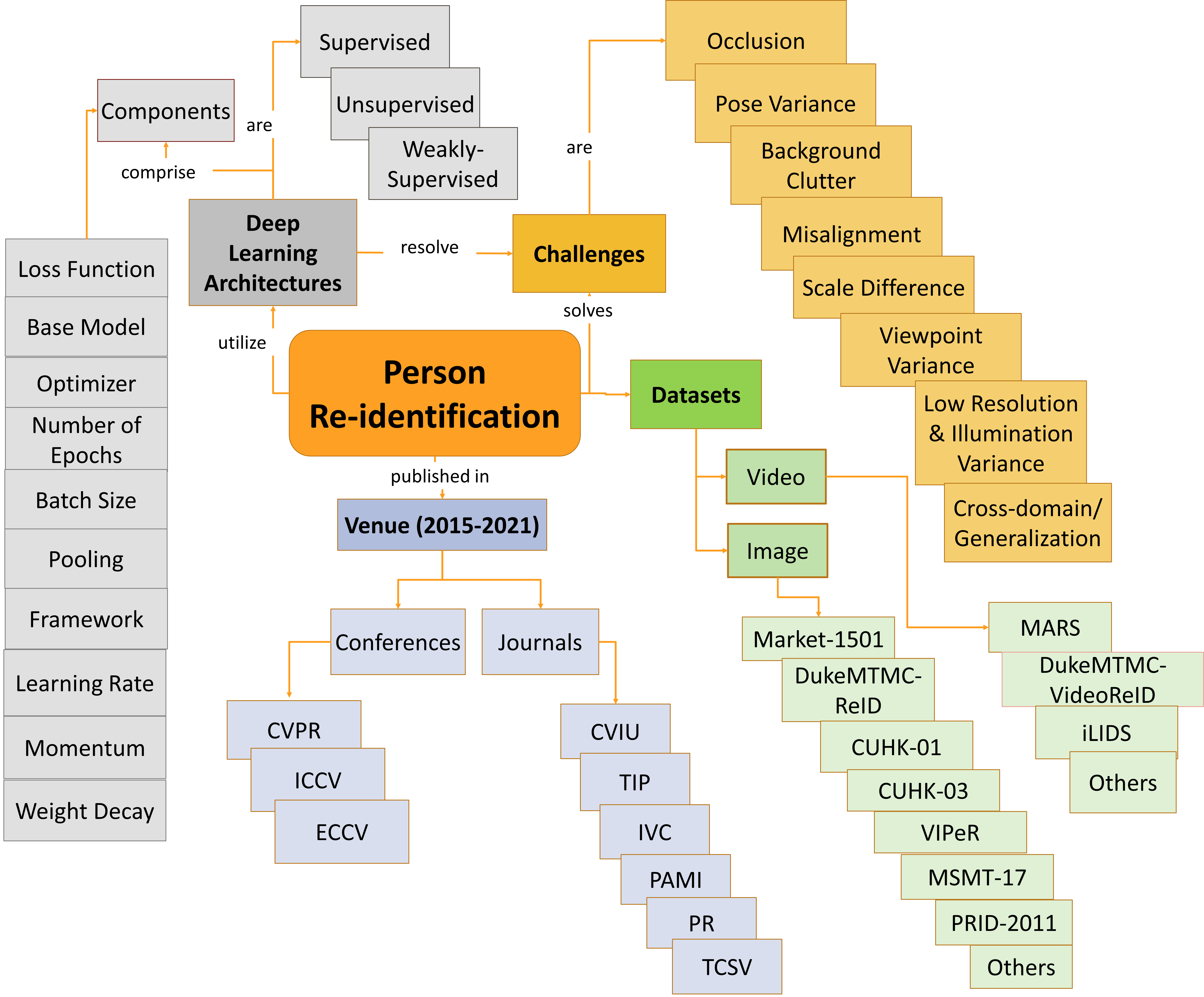}
    \caption{Taxonomy of the Review}
    \label{fig:taxanomy}
    \end{center}
\end{figure*}

The first category extracts the details about the origin of the article whether the article is published in a conference or a journal, the year of publishing, the title of article, which problem the particular article is addressing, image based or video based and finally which methodology is used i.e. CNN, Vision Transformer (ViT), GANs or Graph CNNs etc. In the second category, the details of enlisted person re-id challenge (Poor resolution, background \emph{etc.}) addressed by the article is explored. In the third category, ,a brief extract of the proposed methodology in the article is analysed.
The fourth category comprehensively extracts the implementation detail including implementation framework, deep learning based methodology, base model, training approach, batch normalization etc. The fifth category tells about public availability of implementation. The sixth category gathers the information about the quantitative performance measures of proposed work. Finally, the details about the datasets used for evaluation is reported in the last category. The Taxonomy of our review is shown in Fig. \ref{fig:taxanomy}.

\subsection{Results}
\label{subsec:results}
\begin{figure} [b]
    \centering
    \includegraphics[width=0.70\linewidth]{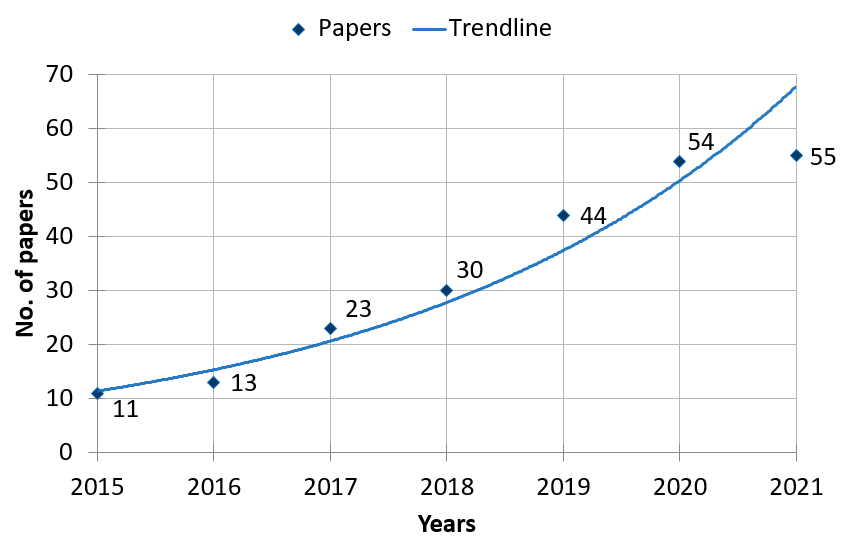}
    \caption{The yearly frequency of selected re-id papers for this review}
    \label{fig:PaperTrendPerYear}
\end{figure}

There were 516 articles collected using the selection criteria.
After removing irrelevant articles database contain 500 articles. According to our inclusion and exclusion criteria more 254 articles were excluded and therefore we selected 230 articles (116 Journal \& 114 Conference articles) for review. The table \ref{table:summaryOfArticlesCollected} depicts the publication venue wise summary of the articles selected for this review, and Fig. \ref{fig:PaperTrendPerYear} illustrates the growing trend of person re-id publications each year in reputed venues.

\begin{table}
\caption{Summary of articles collected}
\label{table:summaryOfArticlesCollected}
\begin{tabular}{ m{3em}  m{15em} m{5em}  m{5em} m{5em} }
    \hline\noalign{\smallskip}
    Sr.No & Venues (2015-2021) & Found & Filtered & Reviewed \\
    \hline\noalign{\smallskip}\hline\noalign{\smallskip}
    1 & Computer Vision and Pattern Recognition (CVPR) & 131 & 72 & 59 \\
    2 & IEEE International Conference on Computer Vision (ICCV) & 50 & 18 & 32 \\
    3 & European Conference on Computer Vision (ECCV) & 45 & 22 & 23 \\
    4 & Elsevier Computer Vision and Image Understanding (CVIU) & 6 & 0 & 6 \\ 
    5 & IEEE Transactions on Image Processing (TIP) & 91 & 46 & 45 \\
    6 & Image and Vision Computing (IVC) & 8 & 3 & 5 \\
    7 & IEEE transactions on pattern analysis and machine intelligence (PAMI) & 42 & 27 & 15 \\
    8 & International Journal of Computer Vision (IJCV) & 8 & 4 & 4 \\
    9 & Pattern Recognition (PR) & 42 & 27 & 15 \\
    10 & IEEE Transactions on Circuits and Systems for Video Technology (TCSVT) & 51 & 31 & 20 \\
    \hline\noalign{\smallskip}
       & Total & 484 & 254 & 230 \\
    \hline\noalign{\smallskip}
\end{tabular}
\end{table}

\section{Datasets and performance evaluation}
\label{sec:datasetsAndPerformanceEvaluation}
Several image based and video based datasets for person re-identification models has been released. In past few years several review papers review the available datasets. But they have not reported the progress on each challenge against each dataset. We review several image and video based datasets by reporting the progress of each challenge on each dataset and how dataset support the challenge to be addressed effectively. The attributes of each of the public datasets is summarized in the table \ref{table:DatasetProperties}. 
\subsection{Performance Evaluation Metrics}
For evaluation purpose, person re-id models use two widely known measurements named Cumulative Matching Characteristics (CMC) also known as rank-k accuracy and mean Average Precision (mAP).  
CMC represents the probability that accurate match of image appears in the top-k ranked retrieved results. CMC will be considered accurate when only one ground-truth exists for each query, since it only considers the first match in evaluation process. However, the gallery set usually contains multiple ground truths in a large camera network, and CMC cannot completely reflect the discriminability of a model across multiple cameras. The other widely used metric, mAP measures the average retrieval performance with multiple ground- truths. It is originally widely used in image retrieval process. 

\subsection{Image based datasets}
\label{subsec:ImageBasedDatasets}  
Several image based person re-id datasets are available. We have considered top 6 image based person re-id datasets \emph{i.e.} Market-1501, DukeMTMC-ReID, CUHK03, CUHK01, VIPeR and MSMT-17 for our review on which person re-id models achieved SOTA performance. Major person re-id challenges are addressed by using these mentioned datasets. Summary of number of papers on major challenges is shown in the Fig. \ref{fig:paperCountonMajorchallenges}.

\subsubsection{Market-1501}
\label{subsec:DatasetMarket1501}
The Market-1501 \cite{zheng2015scalable} dataset is specifically for person re-id and proposed in 2015. It was collected in front of a supermarket in Tsinghua University. A total of six cameras were used, including five high-resolution cameras, and one low resolution camera. Overlap exists among different cameras. Overall, this dataset contains 32,668 annotated bounding boxes of 1,501 identities. Each annotated identity is present in at least two cameras, so that cross-camera search can be performed.
Market-1501 is one of the most trending and widely used dataset by researchers to address multiple person re-id challenges like illumination, viewpoint, pose variation and body misalignment. In 2019, this dataset helped to boost the performance of person re-id models. Market-1501 dataset provides the viewpoint angle and this property help the deep learning models in learning the deep features to resolve the viewpoint and pose variation challenge. In last five years many person re-id algorithms achieved the SOTA results by using Market-1501 dataset.

\subsubsection{DukeMTMC-ReID}
\label{subsec:DatasetDukeMTMC-ReID}
DukeMTMC-ReID \cite{ristani2016performance} has significant potential, it provides access to details like frame level, ground truth, full frames and calibration information \emph{etc.}). It correspond to images of 1,852 people existing across all the 8 cameras. It covers 1,413 unique identities with 22,515 bounding boxes that appear in more than one cameras. It also consists of 439 distractor identities with 2,195 bounding boxes that appear in only one camera. The size of the bounding box varies from 72*34 pixels to 415*188 pixels.

\subsubsection{CUHK01}
\label{subsec:DatasetCUHK01}
The image quality of CUHK01 \cite{li2012human} datasets is relatively good and this benefits the person re-id models to achieve good results and perform well in real world scenarios. This dataset was published in 2012 and consists of 3884 images of 971 people. Two disjoint cameras were used to capture different views. Each camera captures two images for each person, total of four images of a person.

\subsubsection{CUHK03}
\label{subsec:DatasetCUHK03}
CUHK03 \cite{li2014deepreid} dataset is one of the largest dataset in 2014 and proved good for person re-id and deep learning models to report effective results. Dataset comprised of 13,164 images of 1,360 people.Images are captured by using six cameras. Each identity appears in two disjoint camera views (\emph{i.e.} in each view there are 4.8 images on average). In CUHK03 bounding boxes are manually labeled and detected from Deformable Part Models (DPM).

\begin{table}
\caption{Properties of each dataset.}
\label{table:DatasetProperties}

    \begin{tabular}{ m{1em}  m{5em} m{2em} m{4em} m{4em} m{2em} m{4em} m{4em} m{4em} m{4em}}
    \hline\noalign{\smallskip}
    Sr.No & Dataset & Year & Environment & Identities & Cameras & Resolution & Label & BBoxes & Challenging Attributes\\
    \hline\noalign{\smallskip}\hline\noalign{\smallskip}
    1 & VIPeR \cite{gray2007evaluating} & 2007 & Campus & 632 & 2 & 48×128 & Hand & 1,264 & VV\tnote{1}, IV\tnote{2} \\
    2 & PRID-2011 \cite{hirzer2011person} & 2011 & Outdoor & 200 & 2 & 64×128 & DPM/ GMMCP/ Hand & 40,000 & IV, VV,,BC\tnote{3}\\
    3 & CUHK01 \cite{li2012human} & 2012 & Campus & 971 & 2 & 60×160 & Hand & 3,884 & VV,OCC\tnote{4}\\
    4 & CUHK03 \cite{li2014deepreid} & 2014 & Campus & 1,360 & 10 & Vary & DPM/Hand & 13,164 & VV,OCC \\
    5 & iLIDS-Vid \cite{wang2014person} & 2014 & Airport & 300 & 2 & Vary & Hand & 42,495 & VV,IV,BC,OCC\\
    6 & Market-1501 \cite{zheng2015scalable} & 2015  & Campus & 1,501 & 6 & 64×128 & DPM/Hand & 32,688 & VV, PV\tnote{5}, RES\tnote{6}\\
    7 & MARS \cite{zheng2016mars} & 2016 & Campus & 1,261 & 6 & 256×128 & DPM/ GMMCP & 1,067,516 & PV,IV,RES\\
    8 & DukeMTMC-ReID \cite{ristani2016performance} & 2017 & Campus & 1,404 & 8 & Vary & Doppia/ Hand & 36,411 & VV,IV,BC,OCC\\
    9 & DukeMTMC-Video ReID & 2017 & Campus & 1,404 & 8 & Vary & Hand & 36,411 & VV,IV,BC,OCC \\
    10 & MSMT-17 \cite{wei2018person} & 2018 & Campus & 4,101 & 15 & Vary & Faster RCNN & 12,6441 & VV,IV\\
    \hline\noalign{\smallskip}
    \end{tabular}
    
    \begin{tablenotes}
        \item[1] Viewpoint Variation
        \item[2] Illumination Variation
        \item[3] Background Clutter
        \item[4] Occlusion
        \item[5] Pose Variation
        \item[6] Resolution
        \item[7] Deformable Part Model \cite{felzenszwalb2009object} (A Pedestrian detector)
        \item[8] GMMCP- Generalized Maximum Multi Clique problem \cite{dehghan2015gmmcp} (A tracker)
    \end{tablenotes}

\end{table}

\begin{figure*} [b]
    \centering
    \includegraphics[width=0.8\linewidth]{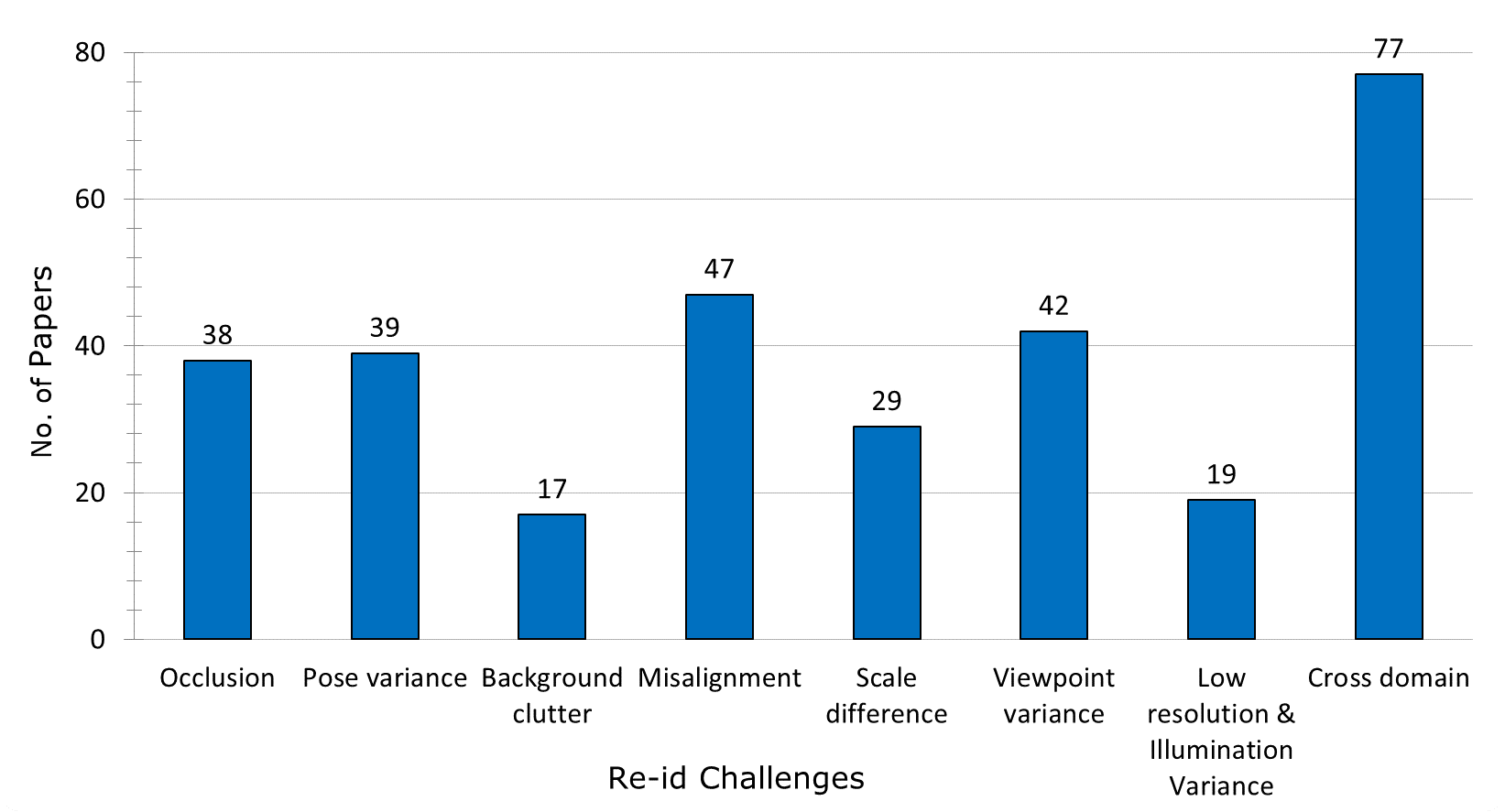}
    \caption{Count of papers on each enlisted challenge}
    \label{fig:paperCountonMajorchallenges}
\end{figure*}

\subsubsection{VIPeR}
\label{subsec:DatasetVIPeR}
VIPeR \cite{gray2007evaluating} is one the most challenging dataset, several researchers tested it and reported very widely and interesting results specially by addressing viewpoint variation challenge. As this dataset is one of the oldest dataset and still in trending for person re-identification models. It contains around 632 identities and images are captured by two cameras one image per person. VIPeR also facilities with the viewpoint angle of each image. In our review we have observed that the in last five years most journal paper on person re-identification test their model on VIPeR dataset to address pose variation and viewpoint challenge.

\subsubsection{MSMT-17}
\label{subsec:DatasetMSMT-17}

MSMT-17 \cite{wei2018person} is one of the largest image based dataset containing 126441 images and 4101 identities. Images are captured in morning, noon and afternoon in campus. In our review we have observed that MSMT-17 is widely used dataset, although it contains similar viewpoint with Market-1501 dataset but this data commonly known for capturing the most complicated scenarios that’s why several and almost all recent person re-id models test their models on this dataset and reported SOTA results. It mostly adopted to address pose variation, viewpoint and body misalignment challenges.

\subsection{Video based datasets}
\label{subsec:DatasetvideoBasedDatasets}

Several video based person re-id datasets are available. We have considered top 3 video based person re-id dataset MARS, DukeMTMC-Video re-id and iLIDS for our review on which person re-id models achieved best performance. Major person re-id challenges are also addressed by using these video based datasets.

\subsubsection{MARS (Motion Analysis and Re-identification Set)}
\label{subsec:DatasetMARS}

MARS \cite{zheng2016mars} dataset is an extended version of Market-1501 dataset. It is largest video based dataset having 1191003 images and with maximum crop size 256*128 among all other video based datasets. In our review we have found that MARS is specifically used to deal with pose variation, viewpoint and similarity measure challenges. Paper published in almost all top journals based on video based person re-id very few of them used this dataset to test their models but these effective results explain that the MARS is significant dataset because all the tracklets and bounding box are generated automatically. This automatic generation makes learning faster.

\subsubsection{iLIDS-VID}
\label{subsec:DatasetiLIDS-VID}

iLIDS \cite{wang2014person} contain 42495 images with 300 identities. iLIDS dataset contain heavy occlusion in the captured images that’s why it mostly used in person re-id models which address occlusion challenge. In our survey we have found that iLIDS majorly help to achieve good results in pose variation, viewpoint and similarity measure challenges. Therefore, in last five years very few journal papers that address these challenges used iLIDS dataset.

\section{Deep Learning Approaches}
\label{sec:deeplearning}

In recent years, neural network-based deep learning algorithms become a popular branch of machine learning. Deep learning algorithms attempt to model high-level abstractions in data by using multiple processing layers with complex structures and are able to outperform state-of-the-art methods in many tasks in the fields of computer vision, natural language processing, robotics, etc. For the interdisciplinary readers, in this section we will present an overview of deep learning architectures used to address various challenges of person re-identification

\subsection{Convolutional Neural Network}
\label{sec:CNN}

For the computer vision task, the input data are images of sizes usually ranging from several hundreds to several tens of thousands of pixels. If a neural network processed this input matrix with only fully connected neurons, the number of parameters to train would be very large, leading to a high risk of overfitting. The convolutional layer was invented to deal with this problem and became the first successfully trained deep neural network. The convolutional layer uses two basic ideas: local receptive fields and shared weights to reducing the complexity of the neural network. A CNN receives a matrix as the input, but connects a hidden node to only a small region of nodes in the input layer, since the spatial correlation is local. This region is called the local receptive field for the hidden node. Moreover all the mappings between a local receptive field and a hidden node share the same weights, since the features are not specific to some regions in an image. These two properties reduce dramatically the numbers of parameters of the network. By applying these two principals, a fully-connected neuron layer is transformed into a  convolutional layer as follows:

\begin{equation}
      y = \alpha \Big( W * X + b \Big) 
      \label{eq:232}
\end{equation}

where X is the input image W is the weight matrix, also called a ’filter’ or ’kernel’. b
is the bias term. The function $\alpha$ is an activation function and the operator * represents the discrete convolution operation. For a two-dimensional image X as input, the convolution operator is defined as:

\begin{equation}
      (W * X)(i,j) = \sum_{m} \sum_{n} X (m,n) W (i-m, j-n)  
      \label{eq:233}
\end{equation}

Intuitively, the output of the convolutional layer is formed by sliding the weight matrix
over the image and computing the dot product. The resulting matrix is called “activation map” or “feature map”. In image processing, convolution operations can be employed for edge detection, image sharpening and blurring just by using different the numeric values of the filter matrix. This means that different filters can detect different features
from an image and capture the local dependencies in the original image. In convolutional
layers, the convolutional kernel or filter, i.e. the coefficients of the weight matrix W, are learnt automatically by the backprop algorithm, and one layer usually contains several such convolution kernels and resulting feature maps.
 
In a CNN architecture, the image is passed through a series of convolutional, nonlinear, pooling layers and fully connected layers, and then generates the output. The Convolution layer is always the first layer where the image is entered. The reading of the input matrix begins at the top left of image with the help of a smaller matrix, which is called a filter. The filter produces convolution, i.e. moves along the input image. The filter multiplies its values by the original pixel values. All these multiplications are summed up. One number is obtained in the end. Since the filter has read the image only in the upper left corner, it moves further and further right by 1 unit performing a similar operation. After passing the filter across all positions, a matrix is obtained, but smaller then a input matrix. This operation, from a human perspective, is analogous to identifying boundaries and simple colours on the image. But in order to recognize the properties of a higher level such as the trunk or large ears the whole network is needed.
 
The network will consist of several convolutional networks mixed with nonlinear and pooling layers. When the image passes through one convolution layer, the output of the first layer becomes the input for the second layer. And this happens with every further convolutional layer. The nonlinear layer is added after each convolution operation. It has an activation function, which brings nonlinear property. Without this property a network would not be sufficiently intense and will not be able to model the response variable (as a class label).
 
The pooling layer follows the nonlinear layer. It works with width and height of the image and performs a downsampling operation on them. As a result the image volume is reduced. This means that if some features (as for example boundaries) have already been identified in the previous convolution operation, than a detailed image is no longer needed for further processing, and it is compressed to less detailed pictures.
 
After completion of series of convolutional, nonlinear and pooling layers, it is necessary to attach a fully connected layer. This layer takes the output information from convolutional networks. Attaching a fully connected layer to the end of the network results in an N dimensional vector, where N is the number of classes from which the model selects the desired class.
 
Initially, the deep learning based re-id research utilized the ImageNet pre-trained weights to fine tune the most popular deep architecture ResNet on the person re-id or pedestrian datasets \cite{v36}, \cite{v42}.

\subsubsection{Global Features Learning}
\label{sec:section_241}

In the past few years, with the rise of deep learning for image classification tasks, the development of various deep architectures and the availability of their pre-trained weight, various classification loss functions are proposed for deep learning base person representations \cite{He2019} \cite{Wu2019}. Furthermore self-learning of person features is endorsed very first time by using pairwise Siamese loss \cite{li2014deepreid} and triplet loss \cite{hermans2017defense} where features selection brings similar identities closer and different identities apart. Person attributes and the salient features in a person image are also worked out to focus on salient regions of a person images \cite{Ou2019} \cite{LiT2018} \cite{nperwaiz19} \cite{vg3} \cite{vg4}. A technique for hard-identity mining is proposed to learn global features in an efficient way by Ristani et al in \cite{IA18}, where a weighted triplet loss function is introduced and a robust two streams metric learning model is proposed for person tracking and person re-id in parallel. In \cite{IA19} a restraint and relaxation iteration (RRI) training scheme is used to propose SVDNet for person re-id where an Eigenlayer is introduced to find the correlation among the global features before fully connected layer of the deep model. However the spatial information loss of local cues need to be handled effectively for improvement of re-id performance.

\subsubsection{Local Features Learning}
\label{sec:section_242}

In addition to learning global visual descriptors for person representation, the re-id research deals with the local vli2014deepreidisual cues as well. A famous strategy for working on the local/ part based person features is to divide the image in multiple local parts or strips. In the re-id approaches \cite{li2014deepreid}\cite{IA8}\cite{IA9}\cite{IA10}\cite{IA11} the person cropped image is divided into various horizontal strips to emphasize on the local discriminative regions in an image. The methodology proposed in \cite{li2014deepreid} divides the person image into three fixed horizontal parts and assumes that the top part of the image contains head, middle one contains torso and the bottom one contains legs. \cite{IA8}, \cite{IA9}, \cite{vg1}, \cite{vg2} use another fixed number of local parts for local features whereas in \cite{IA10} a divide and fuse strategy is adopted to focus on local parts of image, however the association among local parts of images is not catered for in any of these works.

One of the major limitation of all these convolution based attention networks is that these networks learn the dependency among immediate neighborhoods both at the initial layers of the network and at the deeper layers as well \cite{v14}. The structure and working mechanism of CNNs do not exhibit the learning of attention dependency at distant positions of an image or feature maps at a particular level. This arises the need to learn self-attention or intra-attention at each learning level (layer). The self-attention learns the associations among different parts of images and embed this information in the global representations.

\subsection{Attention Based Deep Network} 
\label{sec:attention}

In the convolution based deep architectures, the attention regions of the images are captured at small receptive fields of the input image and are propagated towards deeper layers of the deep networks to have their aggregated perception \cite{krizhevsky2012imagenet} \cite{v18} \cite{vg5} \cite{vg6} \cite{vg7}. This cumulative attention information provides good intuition about the regions of the images which can play an important role for the task of person re-id. But the deeper layers of CNNs are bound to view whatever the initial layers of a deep architecture have forwarded to them hence losing the global context to greater extent.
 
Over the last couple of years, with the advancement of hardware architectures and the availability of large-scale datasets, the attention based deep architectures are being explored extensively for the person re-identification ongoing research  \cite{v20}\cite{v21}. Attention mechanisms get intuition from the human vision system and focus on the most attentive regions of the image for decision making and suppress the less attentive regions. 
 
The attention operations map the queries (Q) and a set of key-values (K, V) to the output vector. The queries, keys and values are the matrices and make the input for an image for which attention is to be computed. Two common kinds of attention functions are the multiplicative attention function i.e. scaled dot product attention, and the additive attention function. The most common method is the multiplicative function to compute the computationally efficient attention matrices instead of the additive attention method. The additive attention function is computationally expensive and uses a feed-forward network with a single hidden layer i.e 3 layer network (input + hidden + output), which involves one multiplication of the input vector by a matrix, then by another matrix, and then the computation of resultant vector. In contrast, the smart implementation of the scaled dot product attention computation does not break out the whole matrix multiplication algorithm and basically is a tight, easily parallelized operation.
 
By using variants of CNNs as a backbone, several re-id networks are designed that explicitly embed higher level attention in addition to the local attentions learnt by CNNs base architectures. Most of these attention learning networks opt multi-stream structures to learn the regional/ spatial attention along with deep convolutional representations \cite{v20}\cite{v21}. In addition of learning the spatial attentions, the attentive but diverse re-id model \cite{chen2019abd} and critical attention learning mechanism \cite{chen2019self} also focus to learn channel wise attentions in order to extract the most significant channels only as all the channels do not contain significant information. In addition to attentive channel information, \cite{chen2019abd} learns correlation among the attentive channels as well.
 
Another multi-task and attention aware learning network \cite{v24} proposed by Chen et al. demonstrated an explicit holistic attention branch to learn global attention in addition to a partial attention learning branch to learn local attention, however the network has additional requirements of the key-points for its local attention branch and learning self-attention within sparse parts of the image.
 
The above mentioned methods learn hard-level attention to perform person re-id, but the computation of only hard attention makes these networks less generalized. Harmonious attention networks \cite{li2018harmonious} handle multi-level attention i.e. both of the hard attention (regional salience) and the soft attention (pixel level salience). It retains the generalization of CNNs as well but it still fails to capture the relationships among distant attentive regions within an image. All these multi-stream attention learning mechanisms need huge computational resources. Moreover, these networks do not learn the associations among far-distant attention regions of an image which are apparently dispersed all over the image but can provide better intuition to re-identify a person. 

\subsection{Self-Attention Based Deep Network}
\label{sec:selfattention}

One of the major limitation of all these convolution based attention networks is that these networks learn the dependency among immediate neighborhoods both at the initial layers of the network and at the deeper layers as well \cite{v14}. The structure and working mechanism of CNNs do not exhibit the learning of attention dependency at distant positions of an image or feature maps at a particular level. This arises the need to learn self-attention or intra-attention at each learning level (layer). 
 
The self-attention learns the associations among different parts of images and embed this information in the global representations. Transformer is a deep learning model for learning self-attentions among a given sequence of inputs and is currently state-of-the-art for solving the language problems \cite{v14}.  However due to the large number of the pixels in images, it is not practical to use a standard transformer in its default mode for the images based classification tasks. A standard transformer encoder intakes a sequence of one dimensional data, learns self-attentions among the given sequences and embeds this self-attention information into the final representations of the data. The self attention computation (SA) is given in the equation \ref{eq:3}.

\begin{equation}
    SA(Q,K,V) = \left(\frac{\exp({QK}^T)}{\sum_j \exp({QK}^T)_j} * {\frac{1}{\sqrt {d_k}}}\right)V 
    \label{eq:3}
\end{equation}
In these methods, multiple parallel self-attention heads are used to run multiple parallel attention functions. The main objective is to learn attention from different positions and representation spaces and jointly update learnable parameters of each attention head as shown in equation \ref{eq:5}. This mechanism efficiently learns the self-attention and associations among local parts at distant positions in an image. The weights of the attention are set on the basis of pairwise similarity among the patches’ sequences. These learnt self-attentions are further refined through the depth of transformer layers.

\begin{equation}
    h_i = SA_{mul}\left(Q{W_i}^Q,K{W_i}^K,V{W_i}^V\right)  
    \label{eq:4}
\end{equation}
where, $ h_i $ is the attention computed by ith head with resultant trainable parameter matrices i.e.  $ {W_i}^Q,{W_i}^K,{W_i}^V $, as shown in equation \ref{eq:4} 

\begin{equation}
    MSA = F_x\left(\sum_i h_i\right)W^{o}   
    \label{eq:5}
\end{equation}
Finally, the multi-head self-attention module integrates the attention computed by each head using the function (F\textsubscript{x}) of concatenation. It limits the information loss through a residual connection i.e. \(W^0\), which contains the normalized output vector of the preceding layer. The \(W^0\) is integrated with a layer's output before submitting it to the next modules. The attention learning mechanism aims to learn the attention and its interdependence in a global manner.
 
For vision tasks, one potential way is to take the pixels sequences as one dimensional input data. However, typically the large number of pixels in images makes the self-attention computation across the whole set of pixels computationally very costly and thus consequently limits the use of transformers particularly for vision applications. Recently, \cite{v15} proposes an effectual pre-processing pipeline to handle the massive pixels of images data, which enables the use of transformers for vision based problems. 
 
Luo et al. first time use the transformer based network for person re-id. \cite{20stnreid} integrates the deep convolutions based re-id module with the pair-wise spatial transformer networks (STN) module to perform the partial person re-id. The spatial transformer networks module samples an affined image (a semantically corresponding patch) from the holistic image to match the partial image. The re-id module learns the embedding of holistic, partial and affined images, the STN module performance is influenced by the re-id module. 
 
Taking inspiration from the state-of-the-art self-attention methods for natural language problems solution, the researchers explored the self-attention impact for person re-id task. In \cite{Li17}, the multi-scale convolutions are applied to the entire image and to the pre-defined three local parts of the image, i.e. upper, middle and bottom. The latent parts localization is performed by using spatial transformer networks to learn the self-attention. However, this work needs the positions of local parts and the value range constraint on the scale parameter as prerequisites. 

\section{Person Re-id Challenges and State-of-the-Art Results}
\label{sec:challenges}
In recent years, person re-id has gained impact full attention in the community of intelligent system and computer vision for various decisive applications. Although person re-id is comprehensively studied by researchers globally but still it is a challenging issue. When images were captured by non-over-lapping cameras under dynamic-environment are of low quality and in some images face or some other important features are not covered comprehensively. Conventional methods based on hand crafted algorithms and small-scale evaluation are not feasible because of their limited applications. In past, results based on subtracting background from frame to frame in multi-camera tracking are not enough due to variations in viewpoint, domain and illumination \emph{etc.} However, the reliable re-id mainly involves the accurate response that is near to real time. This requires the availability or selection of good quality images to cope the challenges. Therefore, bunch of challenges are still not fully addressed \emph{i.e.} occlusion,pose variations, background clutter, misalignment, scale, illumination, viewpoint changes, poor resolution and cross-domain or generalization.
 
In our review we have provided detailed progress on each of the challenges in last six years as shown in Fig. \ref{fig:paperCountonMajorchallenges}. How the results are getting better yearly and which datasets are mostly used to solve these challenges. How deep learning techniques have improved results on each challenge. 

We have categorized the papers of each challenge on the basis of adopted algorithmic techniques \emph{i.e.} CNN, attention and self-attention based approaches.
\begin{itemize}
    \item \textbf{CNN based Re-id Solutions:} CNNs are widely used because of their in-depth learning. They consists of number of convolutional layers. Mainly used for image processing, segmenation, classification and other correlated data. A sliding filter is used to convolve over the entire image to gradually learn the portions of the image and its surroundings.
    \item \textbf{Attention based Re-id solutions:} Attention-based approaches are also in wide use due to their focusing behaviour of learning specific attributes. There use enhances the focus on important part of data while ignoring the unused background information.
    \item \textbf{Self-Attention based Re-id Solutions:} While transformers are newly introduced in vision processing tasks to boost their performance at next level. They use multi-head self-attention mechanism to learn image embedding. Transformers can also be used in conjunction with CNNs however pure transformers applied directly to image patches can also produce best results.
\end{itemize}

\subsection{Occlusion}
    \label{subsec:occlusion}
Occlusion is caused by any overlapping object that may lead to wrong results. Persons are occluded by various environmental objects (traffic sign board, trees in parks, vehicles in parking) or by other pedestrians in person retrieval scenarios and make it difficult to track the movement of people. Visual illustration of occlusion is illustrated in the Fig. \ref{fig:challenges}.
 
When a person is occluded, the features extracted from the whole image may contain the distracted information and leads to wrong results if the model is not able to differentiate between the occluded region and person region. Recently \cite{hermans2017defense,sun2018beyond,zheng2015partial} few of them solved the occlusion challenge but did not consider the situation where the person is occluded due to different obstacles like cars, walls, shelves, poles and other people. In previous works to reduce the effect of obstructions, the occluded target person in the probe images is cropped manually and then use the non-occluded part as the new probe image. There are limitations in such approaches because it require more processing time due to manual cropping. 
 
Some of the works on occlusion \cite{sun2018beyond,he2019foreground} have achieved better performance by using part based models via part to part matching. It’s not ideal to discard the occluded part from the target image, recently attention mechanism has been introduced in person re-id models which pay more attention to the non-occluded part during feature construction and preserve the most discriminating and informative appearance information and leads to better person retrieval results. In the literature there were many approaches reported to handle the occlusion. But we have divided the collected papers in three sub-categories: CNN-based, Attention-based and Transformer-based.
    
    \begin{figure} [hbt!]
        \includegraphics[width=\textwidth]{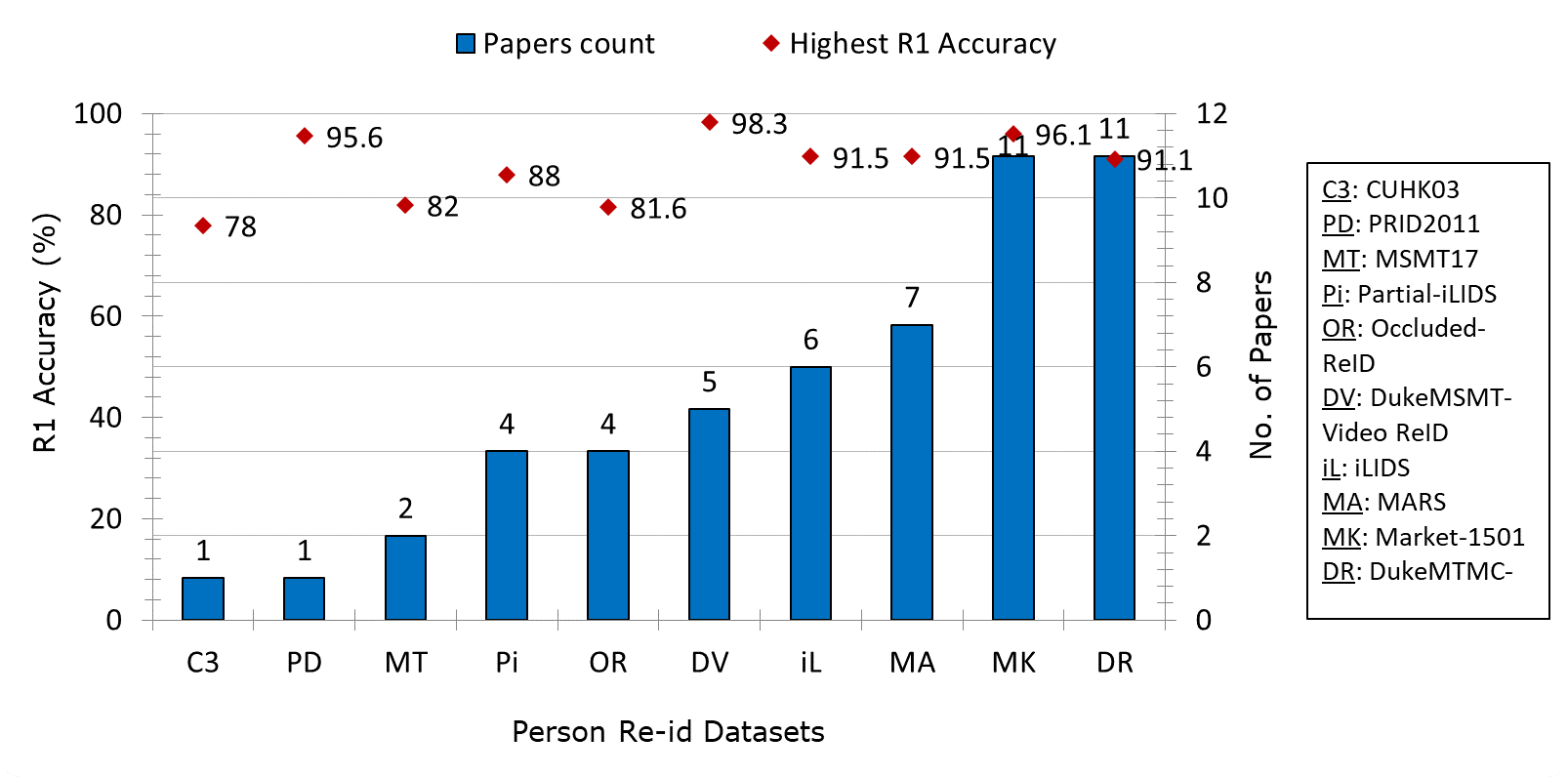}
        \caption{Progress on the challenge of occlusion for person re-id benchmarks}
        \label{fig:progressOnOcclusion}
    \end{figure}
    
    \subsubsection{CNN-based approaches:}
    \label{subsubsec:occlusionCNNApproaches}
    In \cite{he2020guided} crowded scenes are challenged. For handling partial occlusion a simple occlusion-aware approach is presented. Fully convolutional network was used to obtain spatial features. In order to make the features more discriminative a salience heat-map is created using combination of mask-guided and pose-guided layer. Salience heat-maps are then used to guide the adaptive spatial matching. Adaptive matching is for assigning the larger weights to foreground human parts, so to obtain effective results as compared to existing state of the art approaches.
    Lingxiao He \emph{et al.} \cite{he2019foreground} designed an occlusion and alignment free framework to obtain spatial pyramid features at multiple levels/scales. Fully convolution network was used to generate discriminative feature maps. No prior information about alignment was used.
    A CNN based estimation model was used to extract the semantic information in \cite{wang2020high}. An adaptive graph based direction providing layer is introduced with a purpose to pass the information of relation to the nodes. This layer also stops the passing of irrelevant information by analysing how strong the link of nodes is, which is determined by amount of shared information. Moreover Cross graph layer was added to embed and learn the topology and alignment information of group of local features using graph matching technique.
     
    A new large scale occluded person re-id dataset was introduced in\cite{yan2021occluded}. Proposed framework was tested not only on existing benchmark datasets but also on the newly proposed occluded dataset. In the presented framework single-scale discriminative features are learnt without using any auxiliary module. A bounded distance loss makes the approach unique as this enables the model to learn discriminative features explicitly from occlusion-based augmented data.
    Jnrui Yang \emph{et al.} in \cite{yang2021learning} a simple and effective mechanism was presented to elevate the occlusion challenge. Pose information was discretized to the level of visibility. In this manner the impact of occluded body regions is suppressed that has eventually helped in learning the robust and effective pose information on occluded person re-id.
     
    Cairong Zhao \emph{et al.} in \cite{zhao2021incremental} presented an Incremental Generative Occlusion Adversarial Suppression (IGOAS) network. The proposed model consists of two blocks \emph{i.e.} a generative block and a global block. Former block generates the easy-to-hard occluded data samples to make the model robust to occlusion challenge while the later block usually extract the global features, suppresses the impact of occluded part and strengthens the focus on non-occluded part of body regions. Results are then concatenated to obtain the discriminative feature representation.
    An encoder-decoder based approach to alleviate the occlusion challenge was proposed in \cite{hou2021feature}. Region encoder was responsible for building a correlation among occluded and non-occluded regions. while the region decoder uses the spatial correlation to recover the occluded regions in particular frames. Moreover, in order to refine the spatial region feature completion a temporal region feature completion module was plugged in that was responsible to create the long-term temporal contexts.
     
    Another network was presented in \cite{zhang2020semantic}. They have combined the intrinsic relationship between the tasks of person re-id and semantic segmentation to alleviate occlusion. Proposed network consists of three branches \emph{i.e.} semantic, local and global branch. Local branch obtains the part level features, global branch was responsible to obtain features robust to occlusion and the semantic branch generates the foreground-background mask of a person image that basically leverages the non-occluded regions of human body. These three branches were trained collectively to obtain the discriminative representation of pedestrian image.
    In \cite{zheng2015partial} a matching framework was proposed that consists of two modules \emph{i.e.} a local and a global. Local matching was based on patch-level while the global matching was part-based to provide the spatial information. Presented approach was tested modified existing dataset (explicitly included partial person images) and a new partial person re-id dataset.
    
    \subsubsection{Attention-based approaches:}
    \label{subsubsec:occlusionAttentionApproaches}
    Occlusion as an explicit challenge was handled in \cite{miao2019pose}. Attention mechanism was used to learn more about non-occluded parts instead of occluded parts. At the matching stage only the visible shared regions based on pose landmarks are compared to obtain the efficient results without any manual cropping. Human pose landmarks worked as guidance to learn only the non-occluded regions. They have generated new occluded dataset named Occluded-DukeMTMC from DukeMTMC to test the proposed framework. While testing, both gallery and probe dataset contains occluded images to make it consistent with real world scenario.
    Another robust framework to resolve the challenge of partial occlusion was proposed in \cite{hou2019vrstc}. Spatio-temporal information was used explicitly to recover the occluded frames based on the visible body parts. Neighbouring frames were used to recover the information about occluded parts that had resulted to obtain accurate temporal information. A temporal attention layer was introduced that accurately learned the missing information from adjacent layers efficiently.
     
    To address the problem of occlusion in more crowded situations a novel deep network named PISNet is presented in \cite{zhao2020not}. It is comprised of Query-Guided Attention Block that enhances the feature learning process of the target image in the gallery under the direction of query. Because of the improved location accuracy and attention based distinctive feature learning improved results are obtained even in occluded regions.
    Yingquan wang \emph{et al.} in \cite{wang2021pyramid} targeted the challenge of occlusion. In the proposed novel framework both short and long-term temporal information was learned using attention mechanism in a hierarchical manner. Learned spatio-temporal representation is then aggregated by an aggregation block that strengthens the learned features and makes them more discriminative to produce useful results on benchmarks.
     
    In \cite{zhang2019scan} a non-parametric attention mechanism was adopted that takes the video pairs as input and provides the matching score. Attention mechanism helped in refining the intra and inter sequence representation of input videos and outputs self and collaborative feature representation of each video. A generalized pairwise similarity measurement was also presented that calculates the pairwise similarity measurement of feature representation. Finally the matching result was obtained using binary classifier.
    Guangyi Chen \emph{et al.} in \cite{chen2019spatial} videos are sliced into different spatial-temporal units. Purpose was to preserve the body-structure information. Quality scores of these spatio-temporal units was obtained using attention model.
     
    In \cite{aich2021spatio} introduced discriminative explicit pathways to learn unique temporal and spatial features using existing 3D convolution network architecture. Each proposed path is responsible to learn motion (dynamic features) and appearance (static features) information of specific person. Moreover local and global features are also learnt to resolve the occlusion and misalignment. These features together make the approach unique and effective using attention maps. Improved results are obtained when tested on video based benchmark datasets.
    Another approach is \cite{eom2021video} to handle the occlusion challenge based on Spatial and Temporal Memory Networks (STMN). Spatial part stores the information about spatial distractors while temporal memory stores the attention information in person videos. Based on collected information aggregated features are learned to obtain the representation that itself is refined at frame-level of person.
    
    \subsubsection{Self-Attention based approaches:}
    \label{subsubsec:occlusionTransformerApproaches}
    An end-to-end part aware transformer was proposed in\cite{li2021diverse} in a weakly supervised manner. Self-attention mechanism was used to capture the context information of full image. Transformer based encoder decoder architecture was used for pixel based context and diverse part-level discovery.
     
    Fig. \ref{fig:progressOnOcclusion} summarized the reported SOTA results on occlusion. While Table \ref{table:OcclusionResults} gives a brief overview of the published work done on occlusion challenge.
 
{\small
\begin{longtable} { m{1.5em}  m{9em} m{10em} m{5em}  m{4.5em}}
    \caption{Results obtained on occlusion challenge against each dataset. Results in bold are the highest. \label{table:OcclusionResults}}\\
    
    
    
    \hline\noalign{\smallskip}
     SN & Paper & Dataset & R1/mAP & Code availability\\
    \hline\noalign{\smallskip}\hline\noalign{\smallskip}
    1 & STRF, 2021, \cite{aich2021spatio}  & MARS & 90.3/86.1 & No\\
      &      & DukeMTMC-VidReId & 97.4/96.4 &        \\
      &      & iLIDs & 89.3/--  &        \\
    \hline\noalign{\smallskip}
    2 & STMN, 2021, \cite{eom2021video} & MARS & 90.5/84.5 & Yes\\
      &      & DukeMTMC-VidReId & 97.0/95.9 &        \\
      &      & iLIDs & 82.1/69.2 &       \\    
    \hline\noalign{\smallskip}
    3 & PSTA, 2021, \cite{wang2021pyramid} & MARS & \textbf{91.5}/85.8 & Yes \\
      &      &  DukeMTMC-VidReId & \textbf{98.3}/97.4 &    \\
      &      &  iLIDs  & 91.5/-- &        \\
      &      &  PRID-2011 & 95.6/-- &        \\
    \hline\noalign{\smallskip}
     4 & SGPR, 2021, \cite{yan2021occluded} & Market-1501 & 96.1/89.3 & No \\
       &      & DukeMTMC-ReId & 91.1/81.3   &        \\
       &      & Occluded-ReId & 78.5/72.9  &        \\
       &      & Occluded-Duke & 69.0/57.2  &        \\
     \hline\noalign{\smallskip}
     5 & PE-PGFA, 2021, \cite{yang2021learning} & Occluded-ReId & 81.0/71.0 & No\\
     &      & Occluded-Duke & 62.2/46.3  &        \\
     &      & Partial-iLIDs & 80.7/85.7  &        \\
     \hline\noalign{\smallskip}
     6 & FPR, 2019, \cite{he2019foreground} & Market-1501 & 95.42/86.58 & No\\
     &      & DukeMTMC-ReId & 88.64/78.42  &        \\
     &      & CUHK-03  & 76.08/72.31        &        \\
     \hline\noalign{\smallskip}
     7 & PGFA, 2019, \cite{miao2019pose} & Market-1501 & 91.2/76.8 & No\\
     &      & DukeMTMC-ReId & 82.6/65.5  &        \\
     &      & Occluded-Duke & 51.4/37.3  &        \\
     \hline\noalign{\smallskip}
     8 & AMC-SWM, 2015, \cite{zheng2015partial} & Partial-ReId & 53.14/-- & No\\
     \hline\noalign{\smallskip}
     9 & PISNet, 2020, \cite{zhao2020not} & Market-1501  & 95.6/87.1 & No\\
     &      & DukeMTMC-ReId  & 88.8/78.7  &        \\
     \hline\noalign{\smallskip}
     10 & GASM, 2020, \cite{he2020guided}  & Market-1501  & 95.3/84.7 & Yes\\
     &      & DukeMTMC-ReId & 88.3/74.4  &        \\
     &      & MSMT-17       & 79.5/52.5  &        \\
     \hline\noalign{\smallskip}
     11 & PAT, 2021, \cite{li2021diverse} & Market-1501 & 95.4/88.0 & No \\
     &      & DukeMTMC-ReId & 64.5/53.6  &        \\
     &      & Occluded-ReId & 81.6/72.1  &        \\
     &      & Partial-iLIDs & 88.0/76.5  &        \\
     \hline\noalign{\smallskip}
     12 & HOReID, 2020, \cite{wang2020high} & Market-1501 & 94.2/84.9 & Yes\\
     &      & DukeMTMC-ReId  & 86.9/75.6  &        \\
     &      & Occluded-ReId  & 80.3/70.2  &        \\
     &      & Occluded-Duke  & 55.1/43.8  &        \\
     &      & Partial-iLIDs  & 72.6/85.3  &        \\
     \hline\noalign{\smallskip}
     13 & VRSTC, 2019, \cite{hou2019vrstc} & MARS & 88.5/82.3 & No\\
     &      & DukeMTMC-VidReId & 95.0/93.5  &        \\
     &      & iLIDs            & 83.4/--    &        \\
     \hline\noalign{\smallskip}
     14 & ATNet, 2019, \cite{liu2019adaptive} & Market-1501 & 45.1/24.9 & No \\
     &      & DukeMTMC-ReId & 55.7/25.6  &        \\
     \hline\noalign{\smallskip}
     15 & IGOAS, 2021, \cite{zhao2021incremental} & Market-1501 & 93.4/84.1 & No \\
     &      & DukeMTMC-ReId & 86.9/75.1  &        \\
     \hline\noalign{\smallskip}
     16 & SCAN, 2019, \cite{zhang2019scan}  & MARS & 87.2/77.2 & No\\
     &      & iLIDs       & 88.0/89.9    &        \\
     \hline\noalign{\smallskip}
     17 & STAL, 2019, \cite{chen2019spatial} & MARS & 71.5/50.8 & No\\
     &      & iLIDs  & 76.7/--    &        \\
     \hline\noalign{\smallskip}
     18 & SRFC, 2021, \cite{hou2021feature} & MARS & 95.2/89.2 & No\\
     &      & DukeMTMC-ReId   & 90.7/80.7    &       \\
     &      & CUHK-03         & 78.0/81.1    &        \\
     &      & MSMT-17         & 82.0/60.2    &        \\
     &      & MARS            & 90.7/86.3    &        \\
     &      & DukeMTMC-VidReId & 97.6/97.0   &        \\
     \hline\noalign{\smallskip}
     19 & SORN, 2021, \cite{zhang2020semantic} & Market-1501 & 94.8/84.5 & No\\
     &      & DukeMTMC-ReId  & 86.9/74.1     &        \\
     \hline\noalign{\smallskip}
\end{longtable}
}

 \subsection{Pose Variance}
    \label{subsec:posevariations}
    Despite of hundreds of research papers, person re-id is still challenging to be solved mainly due to complex view variations and pose variations in the person images as shown in Fig. \ref{fig:challenges}. The problem of body misalignment caused due to pose/viewpoint variations and occlusion results in imperfect detection.
    
    \subsubsection{CNN-based Approaches:}
    \label{subsubsec:PoseVariationCNNBasedApproaches}
    Spatial Interaction and Aggregation (SIA) module was introduced in \cite{hou2019interaction} to increase the feature learning capability of CNN. The module adapts receptive field according to pose and scale of input image and hence resolves the issue of large body variations. Another module named Channel Interaction and Aggregation (CIA) was also used to semantically aggregate the similar channel features to make the better feature representation even for small visual cues.
    In \cite{papandreou2018personlab} part based model bottom-up approach was used in a fully conventional way to improve the long-range predictions. Once the localization of keypoints is done then greedy decoding process was used to group the instances. Proposed approach was confident in detection as the detection process starts from distinguishing key-points \emph{i.e.} nose to produce best results even in clutter and with variant poses.
     
    In \cite{su2017pose} presented a novel Pose-driven Deep Convolutional model that learns the global and local representation simultaneously. Softmax loss was used to learn the global (whole body) representation while a Feature Embedding subnet was implemented to learn the local (body-part) representation. This local representation makes possible to learn the affine transformation and relocate the regions for easy recognition across multiple cameras. Pose Transfer Network further makes it possible to handle the pose variations. Similarity measure is then fed with this effective fusion of features, to obtain the effective results. It is noted that model is end-to-end and model weights and representations are learned jointly.
     
    Muhammed Kocabas \emph{et al.} \cite{kocabas2018multiposenet} has presented his work to learn multi-person pose estimation using 4x faster bottom-up approach. Their proposed model is multi-task. Person detection, segmentation and pose estimation are jointly learned using shared backbone. Network have the key-points and after person detection it forms the pose by assigning key-points information to each detected person. Their method can also be extended for other tasks \emph{i.e.} person segmentation.
    Yeong-Jun Cho \emph{et al.} in \cite{cho2016improving} analysed camera viewpoints and person poses. Captured images are calibrated and their respective pose is estimated. Proposed multi-pose model gives representative features that are clustered into four groups according to poses \emph{i.e.} left, right, front and back. In a weighted summation aspect, matching scores are calculated between multi-pose models to produce effective results.
     
    In \cite{quispe2019improved} challenge of extreme changes in pose and view point is observed.  A novel unified framework that combines the saliency and semantic parsing to enhance the performance results. Saliency is about focusing at the points that are at the first glance by human eye. Global representation is carried out by saliency and semantic parsing masks.These generate the two types of complementary feature maps that improve the results. Semantic parsing is used to encode all parts of the person and this results in overcoming the challenges of occlusion and misalignment in the bounding box. Hence each sub-net stream learns to resolve different scenarios of the problem.
    In \cite{chen2018deep} attribute based approach was used to approach the challenge of large visual variations and spatial shifts. Large spatial shifts are caused by different pose variations and camera views. Pedestrian attribute assisted CNN-based framework was consists of two parts. In the first part attributes are learned. LOMO features are extracted that are specifically in use to make the model viewpoint invariant. These hand crafted low level features are combined with high-level learned CNN features to obtain robust and diverse feature representation model. The learned embeddings from this first step was then used at second step. In this second step two neural networks fuse together. One is pre-trained network on attribute labels and the other is CNN pre-trained on person re-id labels. These networks are integrated into triplet architecture. Hence optimal fusion parameters are learned in this manner.
     
    Niall McLaughlin \emph{et al.} in \cite{mclaughlin2016person} proposed a strategy to tackle the challenge of appearance changes. In the proposed unified framework invariant feature extraction and supervised learning are combined. To overcome the problem of over-fitting several techniques were used including multitask learning, data augmentation and dropout. Siamese network architecture was used to extract the useful features. Siamese architecture helps to train the network to generate low dimensional feature representation. Diverse images of same person are mapped onto same location in feature space wheres images of different people are mapped onto different locations in the feature space. This network setting helps to learn the subtle cues from diverse set of images.
    In \cite{wang2016deeplist} Jin Wang \emph{et al.} proposed to learn the deep representation with  a novel adaptive margin list-wise loss. In training ranking lists are introduced to replace image pairs. Ranking lists resolve the issue of data imbalance. And Adaptive margin in the list-wise loss function helps in assigning larger margin to hard negative samples. Four convolutional neural networks were combined in one architecture, each network takes input of different body parts or scales. In training stage network was jointly optimized with similarity layer that can be separated at testing stage; this separation accelerates the computation.
     
    To overcome the challenge of pose variation ensemble of invariant features were proposed in \cite{lee2016ensemble}. Ensemble of invariant features helped to achieve the robustness against multiple challenges includes partial occlusion, camera color changes, pose and viewpoint variations. Ensemble of invariant features includes pose-invariant features of specific regions and of holistic image. Invariant features are extracted using DCNN. Region based features are extracted from different body parts via Gaussian Mixture Models on color histograms. Each Gaussian distribution represents the dominant color mode.
    Main goal of \cite{li2019pose} was to come up with a solution that is robust in handling both pose variations and misalignment. Pose Guided Representation composed of two components \emph{i.e.} Pose Invariant Features (PIF) and Local Descriptive Features (LDF). PIF handles the pose variations, it approximates pose invariant representation via pose estimation and normalization. While LDF resolves the misalignment errors, it predicts the discriminative representation learning via body region segmentation. To achieve the robust results at test time, the pose estimation and region segmentation only applies during training time.
     
    Shoubiao Tan \emph{et al.} addressed the challenge of variation in pose and viewpoint across multiple camera views in paper \cite{tan2016dense}. To model the variations of poses a ranking based strategy was proposed that fuses dense invariant features. Images are first divided into dense sampled patched inspired by the fact that local features perform better than global features. To achieve viewpoint invariant representation, it is considered that discriminative parts of a person appear in different regions and in different views. Hence corresponding patch of first image is searched in the neighbourhood of second image. Moreover based on the dense invariant features Support Vector Ranking was used further to learn the transformation across the views and hence results are improved.
    A network based on Inception architecture was presented in \cite{barbosa2018looking} to handle multiple challenges \emph{i.e.} pose, view and illumination. They have made the network to learn the structural aspects of human body. The network was trained and tested not only on real datasets but on a newly formed synthetic datasets as well.
     
    In \cite{li2021person} teacher-student model was used in which teacher acts a guider to extract the global features. These global features are then used to train the other branch of the model that is local feature branch. Moreover a part prediction alignment module was also introduced that align the different parts of same person that resulted in effective estimation of person pose and alignment.
    To alleviate the pose variation problem a fine-grained person re-id solution was proposed by Jiahang Yin \emph{et al.} in \cite{yin2020fine}. The proposed solution was focused on dynamic pose features. Two types of pose features are incorporated \emph{i.e.} motion-attentive local dynamic pose feature and joint-specific local dynamic pose feature. Motion and global person features are incorporated in the model to obtain the discriminative features. Moreover a new dataset was also proposed to better resolve the issues specific to fine-grained person re-id problem.
     
    In \cite{zhou2020fine} a channel parse block was responsible to extract the pose information at pixel level. The required information was obtained by suppressing the background to avoid the inaccurate detection due to occlusion. Moreover a patch level alignment mode was also proposed to handle the misalignment at local level in a fine-grained way.
    In \cite{wang2021horeid} a high order re-id framework was proposed to handle the pose alignment problem using semantic fine-grained part details of multi-level feature maps. These multi-level feature similarity maps were then used to find the difference of similarity among aligned and misaligned parts of person images. As the similarity information among two person images found reduced so the proposed framework successfully increase the robustness of pose features to rectify the misaligned pose information.
     
    A unified end-to-end trainable framework to handle the challenge of pose and viewpoint variation was proposed in \cite{shen2021person}. For accurate ranking of gallery images a novel Kronecker Product Matching operation and group shuffling random walk operation was presented. Together these modules are responsible to accurately rank the probe to gallery and gallery to gallery affinities. This accurate ranking resulted in improved learning of features when embedded with deep learning frameworks.
    Mid-level multi-type human attributes were learned in \cite{su2018multi} in weakly supervised manner.These human attributes were then used to learn features of images with large visual variance for person matching. Contextual cues among attributes were considered to play the vital role in boosting the accuracy of the model in a progressive way.
     
    In \cite{saquib2018pose} (both for image and video) challenge of pose variation and alignment was targeted using both fine and coarse information of acquired person image. They have adopted an unsupervised re-ranking framework, produced results are re-ranked based on simple aggregate of Euclidean distances among the provided gallery and probe images.
    Pose Invariant and Embedding (PIE) \cite{zheng2019pose} resolves the challenge of pose changes and errors in pose estimation. At first place PosBox is designed to formulate a bounding box around each individual. Due to the careful use of pose estimators \cite{wei2016convolutional}, PoseBox produces well-aligned person images so that discriminative and up to the mark features could be learned even in intensive scenarios of posel changes. Secondly Pose Box Fusion (PBF) CNN is added during formulation of PoseBox to reduce the impact of information loss and errors in pose estimation. PBF-CNN takes three streams of input \emph{i.e.} Original image, PoseBox and confidence score of pose estimation. PBF achieves globally optimized results on PoseBox as compared to original image. This PIE plays its role as FC activation of PBF network.
     
    Pose variation is a complex challenge and is targeted in \cite{liu2018pose}, GAN framework for transferable pose variation was based on sample augmentations.
    Xuecheng Nie \emph{et al.} \cite{nie2018pose} challenged pose variations to achieve efficient results on person re-id. They have tried to overcome the limitations of top-down and bottom-up approaches by using unified model based on regression process. The proposed joint framework simultaneously models the detection and joint partition. This has helped to precisely detect with partitioned joints using candidate voting scheme, that further speed-ups the pose estimation. Feed-forward pass helped in achieving higher results as compared to other top-down approaches for achieving optimal improved performance results on multi-person joint configurations pose estimations.
    Another representation learning scheme was proposed in \cite{wu2020adaptive} for video person re-id that is based on adaptive graphs. Pose alignment connection was built to have an adaptive structure aware graph. In order to refine the regional features, feature propagation was performed on adjacency graph in an iterative manner. To make the representation more discriminative temporal resolution aware regularization scheme was also evolved that ensures the consistent temporal resolution for alike identities.
     
    Xiaoqiang Hu \emph{et al.} in \cite{hu2021hypergraph}. In the presented hypergraph video pedestrian re-id method a posture structure and action relationship was explored. It makes full use of the walking posture of a pedestrian. They have used graph convolution network to preserve the structure information obtained from a pedestrian image. Structural relationship was then formed by detecting the joint point regions of the pedestrian. Moreover action hypergraph was also constructed by using the action information obtained by tracking the moving position of the detected joint points. Structure and action information together forms the saliency score which is then converted into a distribution problem to obtain the final results.
    In \cite{chen2016similarity} a unified procedure is describe to handle the challenge of pose variation by stripping down image horizontally in spatial domain to maintain the geometric structure. Feature maps are obtained using \cite{chen2015similarity}. Feature maps of each stripped part were compared with corresponding part in other image on the basis of learnt similarity measure. Each similarity measured score is combined to achieve flexibility. This approach is more robust to occlusion and hence less risky of mismatching.
     
    Le An \emph{et al.} \cite{an2015person} challenged the pose variations caused due to multiple reasons \emph{i.e.} illumination changes, appearance changes and occlusion. For re-id gallery and probe images are projected onto RCCA subspace. In RCCA subspace reference descriptors of probe and gallery images are generated via measuring the similarity among images and the reference data. Identity of probe image is determined via comparison of probe reference descriptor and gallary reference descriptor. Saliency based matching is further used to add re-ranking step that improve the results further.
    Yifan Sun \emph{i.e.} in \cite{sun2019learning} designed a discriminative feature descriptor named as Part-based Convolutional Baseline that is declared general as it is able to handle the challenges like pose estimation, human parsing and uniform part partitioning. Other component named as refined part pooling is based on feature descriptor and is used to precisely locate the parts. Their idea is based on within-part consistency which states that pixels in a well located part are more alike to each other and more dissimilar from other pixels of other parts. And if a pixel in a part is more alike to the pixels of other parts then it is considered as outlier and this refers to inefficient partitioning. Refined Part Pooling (RPP) handles this outlier issue by reallocating the pixel to its closest matching part. RPP trained in semi-supervised way and needs no part labeling. This approach increased the generality of model and boosted the performance as well.
     
    A pose-aware multi-shot matching strategy was presented in \cite{cho2018pamm}. It estimates the pose information using pose estimation methods and then multi-shot matching is performed. The approach analysis both camera viewpoint and person poses. Four image clusters are formed on the basis of front, back, left and right poses. Multi-pose model is generated based on the four feature descriptors formed on the basis of image clusters. Matching scores are then calculated between generated multi-pose models. Additional cues like person poses and 3D scene information makes the proposed model more tractable.
    
    \subsubsection{Attention-based Approaches:}
    \label{subsubsec:PoseVariationAttentionBasedApproaches}
    Challenge of pose variation was addressed \cite{yang2019towards} in an end-to-end way using overlapped activation penalty. Activation penalty pays more attention to the less activated regions of the image. Diverse complementary features were learnt effectively with less penalty.
    Challenges of background clutter, alignment, occlusion and pose are resolved in a unified framework \cite{xu2018attention}. In the proposed framework attention-aware pose guided network is used to handle the challenge of occlusion, misalignment and pose. And noise was excluded via filtering finer parts based on attention mechanism.
    Attention mechanism was adopted in \cite{zhang2020relation} to tackle the problem of pose variation via efficiently capturing the global structure information to better formulate the attention based discriminative feature learning for person re-id task. Global structural information that is kind of clustering information is better to have the contextual information . This global relation helps to leverage discriminative attention on human body regions. In this way both global (semantic) and local (human body regions) are learned to produce the effective results.
     
    An end-to-end Comparative Attention Network (CAN) based framework presented in \cite{liu2017end}.Model is inherent with comparative components. They have tackled appearance variances caused due multiple factors mainly includes different poses and camera views. Proposed model adaptively finds the multiple local regions comprised of discriminative information via set of glimpses. At each glimpse (or region of attention) model creates different parts. Model takes location of previous glimpses and raw person images as input, and formulates next glimpse of local regions as output. These glimpses formulate a kind of dynamical pooling feature, resulted in enhanced performance as compared to conventional pooling features.Which is then integrated to further improve the performance.
     
\begin{figure} [hbt!]
        \includegraphics[width=\textwidth]{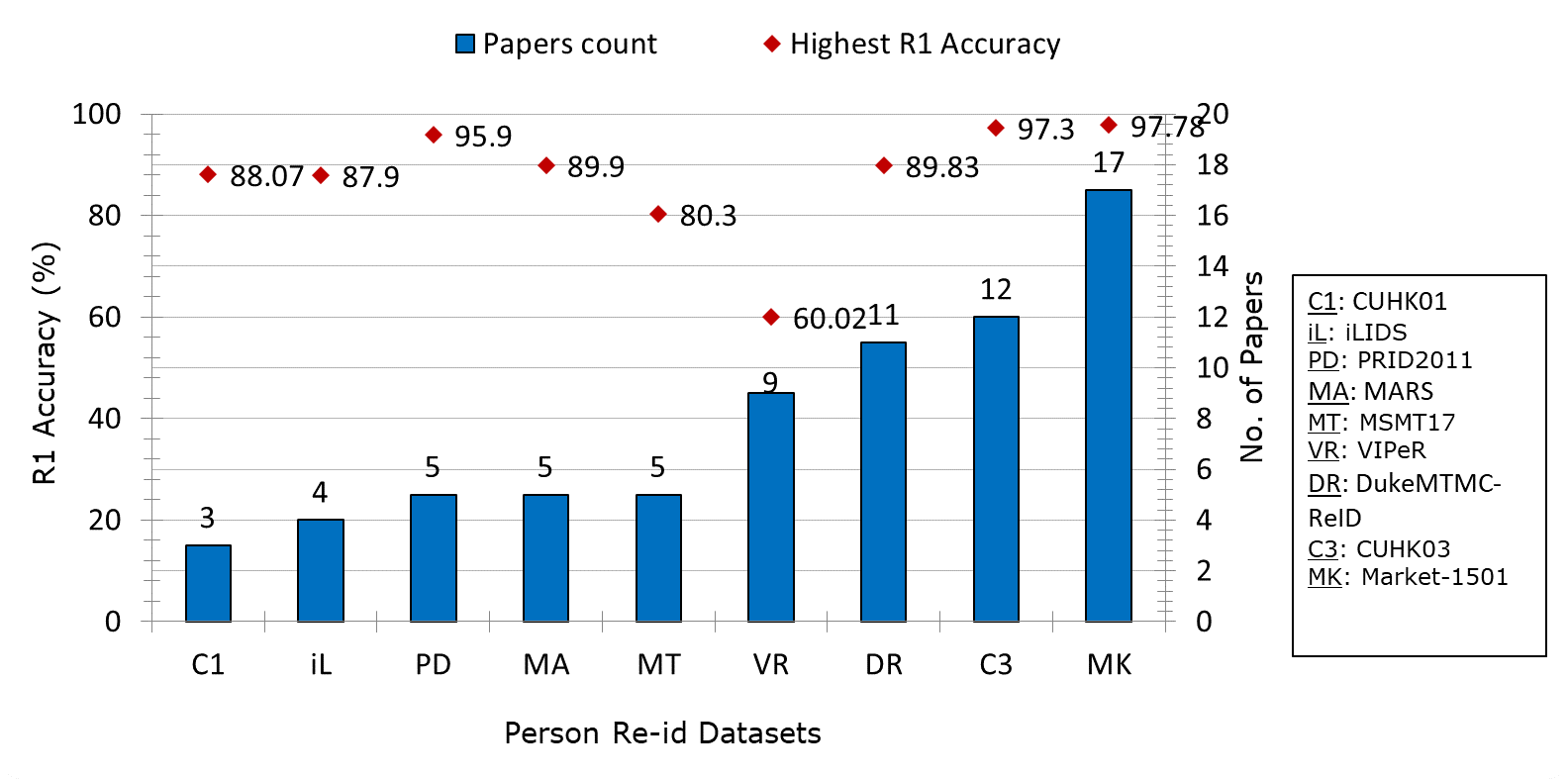}
        \caption{Progress on the challenge of pose variation for person re-id benchmarks}
        \label{fig:progressOnPosevariation}
    \end{figure}

    An end-to-end pose-guided framework was introduced in \cite{gao2020pose} to tackle the challenge of occlusion. In the jointly learned visibility prediction model pose-guided attention and part visibility models are combined. Graph matching technique was then adopted to match the self-learned features of  visibility cues and self-mine visibility score of the gallery and test image accordingly. Matching holistic images with occluded images is an inefficient approach and leads to wrong results as well. Hence occlusion was dealt explicitly.
Fig \ref{fig:progressOnPosevariation} shows the progress on pose variation in literature since 2015.
    
    {\small
    \begin{longtable} {m{1.5em}  m{9em} m{10em} m{5em}  m{4.5em}}
    \caption{Results obtained on Pose-variation challenge against each dataset. Results in bold are the highest.\label{table:PoseVariationResults}}\\
    
    \hline\noalign{\smallskip}
     SN & Paper & Dataset & R1/mAP & Code availability\\
    \hline\noalign{\smallskip}\hline\noalign{\smallskip}
    1 & PDC, 2017, \cite{su2017pose}  & Market-1501 & 84.14/63.41 & No\\
      &      & CUHK-03 (Labeled) & 88.7/-- &        \\
      &      & CUHK-03 (Detected) & 78.29/--  &        \\
      &      & VIPeR & 51.27/--  &        \\
    \hline\noalign{\smallskip}
    2 & RGA, 2020, \cite{zhang2020relation} & Market-1501 & 96.1/88.4 & Yes\\
      &      & CUHK-03 (Labeled) & 81.1/77.4 &        \\
      &      & CUHK-03 (Detected) & 79.6/74.5  &        \\
      &      & MSMT-17 & 80.3/57.5 &        \\  
    \hline\noalign{\smallskip}
    3 & SIA+CIA, 2019, \cite{hou2019interaction} & Market-1501 & 94.4/83.1 & No \\
      &      &  DukeMTMC-ReId & 87.1/73.4 &    \\
      &      &  CUHK-03 (Labeled)  & 92.4/-- &  \\
      &      &  CUHK-03 (Detected)  & 90.1/-- &  \\
      &      &  MSMT-17 & 75.5/46.8 &        \\
    \hline\noalign{\smallskip}
     4 & CAM+RAM, 2019, \cite{yang2019towards} & Market-1501 & 94.7/84.5 & No \\
       &      & DukeMTMC-ReId & 85.5/72.9   &       \\
       &      & CUHK-03 (labeled) & 70.1/66.5  &    \\
       &      & CUHK-03 (Detected) & 66.6/64.2  &    \\
     \hline\noalign{\smallskip}
     5 & AACN, 2018, \cite{xu2018attention} & Market-1501 & 88.69/82.96 & No\\
     &      & DukeMTMC-ReID & 76.84/59.25  &        \\
     &      & CUHK-01(Detected) & 88.07/--  &        \\
     &      & CUHK-03(Labeled) & 81.86/81.61  &        \\
     \hline\noalign{\smallskip}
     6 & PSE, 2018, \cite{saquib2018pose} & Market-1501 & 90.3/84 & Yes\\
        &      & DukeMTMC-ReId & 85.2/79.8  &        \\
        &      & MARS  & 76.7/71.8        &        \\
     \hline\noalign{\smallskip}
     7 & PaMM, 2016, \cite{cho2016improving} & iLIDS & 30.3/-- & No\\
        &      & PRID-2011 & 45.0/--  &        \\
     \hline\noalign{\smallskip}
     8 & SCSR, 2016, \cite{chen2016similarity} & VIPeR & 53.54/-- & Yes\\
        &       & Market-1501 & 51.9/-- &   \\
     \hline\noalign{\smallskip}
     9 & SOMAset, 2018, \cite{barbosa2018looking} & Market-1501  & 81.29/56.98 & No \\
     &      & CUHK-03 (Labeled)  & 85.9/--  &        \\
     \hline\noalign{\smallskip}
     10 & PPA, 2021, \cite{li2021person}  & Market-1501  & 92.4/79.6 & Yes\\
     &      & DukeMTMC-ReId & 85.1/71.8  &        \\
     &      & CUHK-03(Labeled)  & 69.2/66.3  &    \\
     &      & CUHK-03(Detected)  & 65.5/62.4  &    \\
     \hline\noalign{\smallskip}
     11 & PIE, 2019, \cite{zheng2019pose} & Market-1501 & 87.33/69.25 & No \\
     &      & DukeMTMC-ReId & 80.84/64.09  &        \\
     &      & CUHK-03(Detected) & 45.88/41.21  &    \\
     \hline\noalign{\smallskip}
     12 & CAN, 2017, \cite{liu2017end} & Market-1501 & 72.1/47.9 & No\\
     &      & CUHK-01(Labeled)  & 87.2/--  &        \\
     &      & CUHK-03(Labeled)  & 77.6/--  &        \\
     &      & CUHK-03(Detected)  & 69.2/--  &        \\
     &      & VIPeR  & 54.1/--  &        \\
     \hline\noalign{\smallskip}
     13 & PaMM, 2018, \cite{cho2018pamm} & MARS & 66.3/-- & No\\
     &      & iLIDs            & 57.3/--    &        \\
     &      & PRID-2011        & 79.4/--    &        \\
     \hline\noalign{\smallskip}
     14 & PAFAM, 2020, \cite{wu2020adaptive} & MARS & 89.8/81.1 & Yes \\
     &      & iLIDS & 84.5/--  &        \\
     &      & PRID-2011 & 94.6/--  &        \\
     \hline\noalign{\smallskip}
     15 & FGSAM, 2020, \cite{zhou2020fine} & Market-1501 & 91.5/85.4 & No \\
     &      & DukeMTMC-ReId & 85.9/74.1  &        \\
     \hline\noalign{\smallskip}
     16 & HOReID, 2021, \cite{wang2021horeid}  & Market-1501 & 97.78/93.94 & No\\
     &      & DukeMTMC-ReId      & 89.83/82.16    &        \\
     &      & CUHK-03(Labeled)      & 96.12/--    &        \\
     &      & MSMT-17      & 78.42/54.77    &        \\
     \hline\noalign{\smallskip}
     17 & PBCNN, 2018, \cite{chen2018deep} & CUHK-03(Detected)  & 65.0/-- & No\\
     \hline\noalign{\smallskip}
     18 & PCB, 2019, \cite{sun2019learning} & Market-1501 & 93.8/81.6 & No\\
     &      & DukeMTMC-ReId   & 84.5/71.5    &       \\
     &      & CUHK-03  & 63.7/57.5  &        \\
     &      & MSMT-17         & 69.8/43.6    &        \\
     \hline\noalign{\smallskip}
     19 & KPMM, 2021, \cite{shen2021person} & Market-1501 & 93.1/84.7 & Yes\\
     &      & DukeMTMC-ReId  & 71.3/84.9     &        \\
     &      & CUHK-03  & \textbf{97.3}/96.4     &        \\
    \hline\noalign{\smallskip}
     20 & PGR, 2019, \cite{li2019pose} & Market-1501 & 93.87/77.21 & No\\
     &      & DukeMTMC-ReId  & 83.63/65.98     &        \\
     &      & CUHK-03(Labeled)  & 92.15/--     &        \\
     &      & CUHK-03(Detected)  & 89.61/--     &        \\
     &      & VIPeR  & 60.02/--     &        \\
     &      & MSMT-17  & 66.02/37.87     &        \\
     \hline\noalign{\smallskip}
     21 & PGR, 2019, \cite{li2019pose} & Market-1501 & 93.87/77.21 & No\\
     &      & DukeMTMC-ReId  & 83.63/65.98     &        \\
     &      & CUHK-03(Labeled)  & 92.15/--     &        \\
     &      & CUHK-03(Detected)  & 89.61/--     &        \\
     &      & VIPeR  & 60.02/--     &        \\
     &      & MSMT-17  & 66.02/37.87     &   \\
     \hline\noalign{\smallskip}
     22 & FGPR, 2020, \cite{yin2020fine} & MARS & 82.9/72.7 & No\\
     \hline\noalign{\smallskip}
     23 & PA-HVPReid, 2021, \cite{hu2021hypergraph} & MARS & 89.9/79.6 & No\\
     &      & iLIDS  & 87.9/--     &        \\
     &      & PRID-2011  & \textbf{95.9}/--     & \\
     \hline\noalign{\smallskip}
     24 & WSMTAL, 2017, \cite{su2018multi} & Market-1501 & 56.6/31.2 & No \\
     &      & VIPeR  & 39.7/--     &        \\
     &      & PRID-2011  & 24.2/--     &  \\
     \hline\noalign{\smallskip}
     25 & WSMTAL, 2017, \cite{su2018multi} & Market-1501 & 56.6/31.2 & No \\
     &      & VIPeR  & 39.7/--     &        \\
     \hline\noalign{\smallskip}
     26 & PReID-RS, 2015, \cite{an2015person} & CUHK-01 & 31.1/-- & No  \\
     &      & VIPeR  & 33.29/--     &        \\
     \hline\noalign{\smallskip}
     27 & DIFs, 2016, \cite{tan2016dense} & CUHK-01 & 39.46/-- & No \\
     &      & VIPeR  & 29.35/--     &        \\
     \hline\noalign{\smallskip}
     28 & SNML, 2016, \cite{mclaughlin2016person} & VIPeR & 33.6/-- & No \\
     \hline\noalign{\smallskip}
     29 & DL-ReID, 2016, \cite{wang2016deeplist} & CUHK-01(Detected) & 57.02/-- & Yes \\
     &      & CUHK-03(Labeled)  & 55.89/--     & \\
     &      & CUHK-03(Detected)  & 50.67/--     & \\
     &      & VIPeR  & 40.51/--     & \\
     \hline\noalign{\smallskip}
     \end{longtable}
     }
\subsection{Background Clutter}
    \label{subsec:background}
    It’s very basic and important problem in person re-id. Example of background variations can be seen in Fig. \ref{fig:challenges}. 
     
\begin{figure} [h]
        \includegraphics[width=\textwidth]{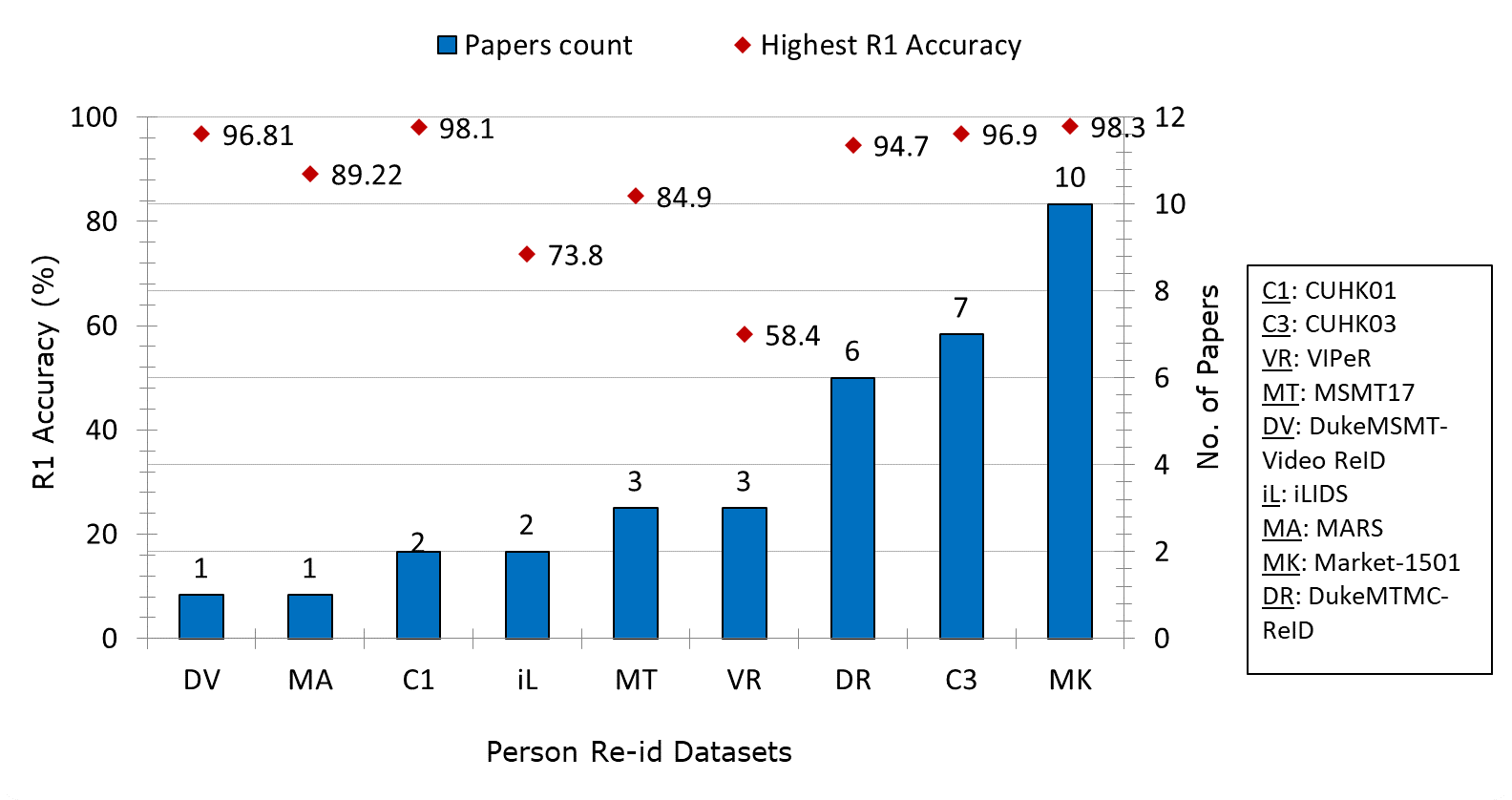}
        \caption{Progress on the challenge of background clutter for person re-id benchmarks}
        \label{fig:progressOnBackgroundClutter}
    \end{figure}

    Complex background makes the detection of pedestrian difficult. Existing methods are not capable to comprehensively address this challenge mainly due to two reasons. Firstly, available datasets have multiple images of one person with same background taken by less number of cameras like dataset CUHK01 and CUHK03 were collected by using 2 cameras, and Market-1501 dataset collected by using 6 cameras. When deep learning based models trained on single datasets they perform usually poor on the other available datasets due to different backgrounds. Secondly, (in the cases where background is complex) most of the existing method ignore the background they just only focus on the visual appearance of the person. Progress of last 6 years is shown in Fig. \ref{fig:progressOnBackgroundClutter}.
    
    \subsubsection{CNN-based Approaches:}
    \label{subsubsec:BackgroundCNNBasedApproaches}
    The challenge of variations in background was investigated in \cite{tian2018eliminating}. In the proposed framework human parsing maps are calculated to cater the background influence. New dataset with different backgrounds was created. Data augmentation technique was used  on existing datasets  to generate images.
    In \cite{pham2017fully} fully annotated person re-id is proposed that is divided into two phases \emph{i.e.} detection and person re-id. In the first human ROIs are automatically extracted using adaptive Gaussain Mixtrure Model method and HOG-SVM detector. Casting shadows are removed using density based score matching. In this scheme both chromatic and physical based features of shadow regions are taken into account. Good performance results are obtained even in the presence of clustered background with occlusion.
     
    In Deep Person model\cite{bai2020deep} contextual information of body parts   \emph{i.e.} from head to foot was integrated to strongly relate the one part with another. LSTM was used in an end-to-end manner, this resulted in enhancing the discriminative features of local parts that will align better to full person (due to prior information of body structure). Global full body representation was also adopted. For identification task both global and part-based features are fed into two separate network branches. Difference of combining features from other approaches lies in ranking task branch that used triplet loss to  learn the similarity measure explicitly.
     
    In \cite{li2020long} long-short temporal spatial clues are used to obtain the robust representation of features. Proposed network does so by combining the motion appearance and motion refinement features. Motion appearance provide the temporal clues (person specific features) by suppressing the background clutter present in multiple scenes. While motion refinement part (activates the person specific features) incorporates the motion refinement layers that are to be executed in multiple motion-excitation blocks of CNN. In this manner the network can differentiate the persons in multiple scenes.
    A self-paced learning framework was presented in \cite{zhou2018deep}. In order to learn the discriminative features and distance metric, a self-paced constraint and a regularization technique was implemented. A part-based deep network was also built part features are learnt in lower convolutional layers and then embedded into higher layers to have the rich and discriminative features.     
    
   \subsubsection{Attention-based Approaches:}
   \label{subsubsec:BacgroundAttentionBasedApproaches}
   In \cite{zhou2019discriminative} an end-to-end Foreground Attentive Neural Network (FANN) is built to distinguish the individuals across non-overlapping camera views. In this attention based approach network focuses on foreground persons. Each image was first passed through encoder, decoder network. Encoder network extract features out of whole image, output of encoder is then further taken for learning discriminative features. Decoder network rebuilds the binary mask for each person in present in foreground of an image. Due to the regularization of decoder and use of local regression loss function encoder slowly start paying attention. Novel triplet loss is used to effectively learn the discriminative features. Use of triplet loss intra-class distances were minimized and inter-class distances were maximised at the same time. 
    
   An extended version of \cite{zhou2019discriminative} was presented in \cite{zhou2019discriminativejournal}. It is an end-to-end foreground attentive neural network with symmetric triplet loss function. Framework is comprised of three sub-networks \emph{i.e.} Foreground attentive sub-network, Body-part sub-network and Feature fusion sub-network. Foreground attentive sub-network comprised of encoder decoder network that takes input of images and focuses its attention on foreground part of input images. Body-part network takes encodes feature maps, slice them and learn features. Resulting feature maps are then fused in third and last sub-network. Finally normalized feature vectors are passed to symmetric triplet loss function.
   A soft mask based end-to-end foreground aware network was presented in \cite{liu2021end}. In order to model the background both pedestrian and camera ID were used. A target attention loss helped in focusing on foreground pedestrian features by reducing the negative impact of changes in background. It is noted that as compared to existing approaches no additional dataset was required to train the model. Improved results have demonstrated the effectiveness of the proposed model.
    
   A multi-level attention and fusion model was proposed in\cite{sun2021memf}. Multi-level attention module has helped in learning the global level features while the multi-layer fusion module has helped in increasing the feature expression at fine granular level. Implementation of the presented model has improved the results as compared to existing.
   To remove the background interference Xin Ning \emph{et al.} designed a feature refinement approach in \cite{ning2020feature}. Instead of directly focusing on high response features, complete features of a person was extracted and then highly valuable features were identified using multi-branch attention network that has resulted in increased performance of the model on benchmark re-id datasets.
    {\small
    \begin{longtable} {m{1.5em}  m{9em} m{10em} m{5em}  m{4.5em}}
    \caption{Results obtained on Background-clutter challenge against each dataset. Results in bold are the highest.\label{table:BackgroundClutterResults}}\\
    
    \hline\noalign{\smallskip}
     Sr.No & Paper & Dataset & R1/mAP & Code availability\\
    \hline\noalign{\smallskip}\hline\noalign{\smallskip}
    1 & CAR, 2019, \cite{zhou2019discriminative}  & Market-1501 & 96.1/84.7 & No \\
      &      & DukeMTMC-ReID & 86.3/73.1 &        \\
      &      & CUHK-03 (Labeled) & 96.9/--  &        \\
      &      & CUHK-03 (Detected) & 93.2/--  &        \\
    \hline\noalign{\smallskip}
    2 & PRGP-DNN, 2018, \cite{tian2018eliminating} & Market-1501 & 81.2/-- & No \\
      &      & CUHK-01 (Detected) & 80.2/-- &        \\
      &      & CUHK-03 (Labeled) & 92.5/--  &        \\
      &      & VIPeR & 51.9/--     &        \\  
    \hline\noalign{\smallskip}
    3 & FANN, 2019, \cite{zhou2019discriminativejournal} & Market-1501 & 94.4/82.5 & No \\
      &      &  DukeMTMC-ReId & 85.2/70.2 &    \\
      &      &  CUHK-01 (Labeled)  & 98.1/-- &  \\
      &      &  CUHK-01 (Detected)  & 81.2/-- &  \\
      &      &  CUHK-03 (Labeled)  & 92.3/-- &  \\
      &      &  CUHK-03 (Detected)  & 70.2/-- &  \\
      &      &  VIPeR & 58.4/-- &        \\
    \hline\noalign{\smallskip}
     4 & TEM-ReID, 2021, \cite{liu2021end} & Market-1501 & 95.0/84.6 & No \\
       &      & DukeMTMC-ReId & 88.7/77.0   &       \\
       &      & MSMT-17 & 76.8/51.0  &    \\
     \hline\noalign{\smallskip}
     5 & HOG–SVM, 2017, \cite{pham2017fully} & iLIDS & 73.8/-- & No \\
     \hline\noalign{\smallskip}
     6 & SBSGAN, 2021, \cite{huang2021unsupervised} & Market-1501 & 87.9/80 & No\\
        &      & DukeMTMC-ReId & 79.7/71.5  &        \\
     \hline\noalign{\smallskip}
     7 & LSTS-NET, 2020, \cite{li2020long} & Market-1501 & 95.8/93.0 & No\\
        &      & CUHK-03(Labeled) & 78.23/72.3  &        \\
        &      & CUHK-03(Detected) & 70.11/67.9  &        \\
        &      & MARS & 89.22/83.12  &        \\
        &      & DukeMTMC-VideoReID & 96.81/93.91  &        \\
        &      & iLIDS & 60.92/--  &        \\
     \hline\noalign{\smallskip}
     8 & LSTM-PDReID, 2020, \cite{bai2020deep} & Market-1501 & 94.48/85.09 & Yes\\
       &       & DukeMTMC-ReID & 80.9/64.8 &   \\
       &       & CUHK-03(Labeled) & 91.5/-- &   \\
       &       & CUHK-03(Detecte) & 89.4/-- &   \\
     \hline\noalign{\smallskip}
     9 & DSPL, 2018, \cite{zhou2018deep} & Market-1501  & 87.05/-- & No \\
     &      & CUHK-01   & 81.33/--  &        \\
     &      & CUHK-03   & 73.16/--  &        \\
     &      & VIPeR   & 56.32/--  &        \\
     \hline\noalign{\smallskip}
     10 & MEMF, 2021, \cite{sun2021memf}  & Market-1501  & 96.11/89.45 & No\\
     &      & MSMT-17 & 82.89/59.8  &        \\
     \hline\noalign{\smallskip}
     11 & WFCB-ReID, 2021, \cite{ning2020feature} & Market-1501 & 98.3/94.2 & No \\
     &      & DukeMTMC-ReId & \textbf{94.7}/90.3  &        \\
     &      & CUHK-03(Labeled) & \textbf{88.6}/84.9  &    \\
     &      & CUHK-03(Detected) & 84.2/80.6  &    \\
     &      & MSMT-17 & 84.9/66.7  &    \\
     \hline\noalign{\smallskip}
    \end{longtable}
    }
\subsection{Misalignment}
    \label{subsec:bodymisalignment}
    In person re-id, body misalignment (the problem in which body parts of person are spatially misaligned) is the essential challenge. The form of it is shown in Fig. \ref{fig:challenges}.
     
\begin{figure}[h]
        \includegraphics[width=\textwidth]{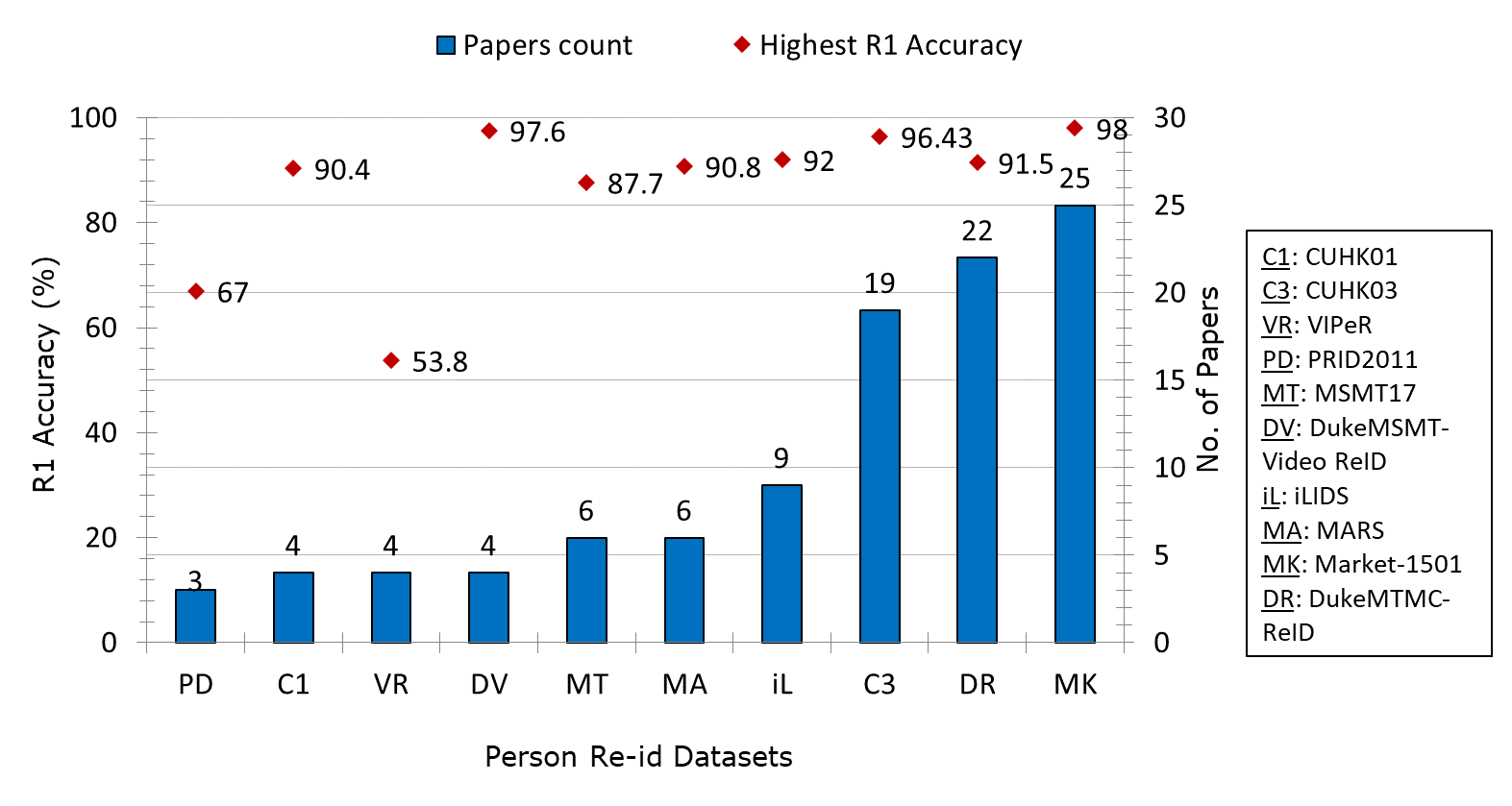}
        \caption{Progress on the challenge of misalignment for person re-id benchmarks}
        \label{fig:progressOnMisalignment}
        \end{figure}

    Human detection becomes imperfect when the body parts were not perfectly aligned and it also makes the matching of person difficult between the probe and gallery image in person re-id. To build an efficient model which provides supervision in part alignment is still challenging task. Fig. \ref{fig:progressOnMisalignment} describe the progress on body misalignment in each of the image based datasets.
       
    In past years, to resolve this challenge conventional neural network based approaches have been proposed and addressed it by using part based alignment and by focusing on shared regions in the images. As part based alignment has shown essential role in the generalization of re-id model. 
    \subsubsection{CNN-based Approaches:}
    \label{subsubsec:MisalignmentCNNBasedApproaches}
    One of the framework to handle the challenge of misalignment was presented in \cite{zhang2019densely}. Person image was divided into spatially semantically aligned 24 parts to learn the canonical surface based representation in UV space.
    Lingxiao He \emph{et al.} \cite{he2018deep} designed an end-to-end framework to leverage feature maps of different sizes without alignment problems as the matching is at pixel level.
    Yifan Sun \emph{et al.}  \cite{sun2018beyond} proposed a supervised framework that consists of two sub networks with purpose of well aligned parts. One subnetwork was based on part-based convolution that generates a descriptor consists of uniform part-level convolutional discriminative features. Other adaptive subnetwork assigns outliers (out while doing uniform partitioning by first subnetwork) to nearest matching similar parts results in consistent refined parts.
     
    In \cite{suh2018part}part-maps are used to learn the human poses. These part-maps are then joined with appearance maps so that part-aligned representation can be computed. In the proposed model two streams are used to generate appearance maps and body part maps separately. Aggregation module then generates the feature maps that are part-aligned. As the computation of body-part information is not relative hence challenge of misalignment is reduced.
    In \cite{zhao2017spindle} structure of human body parts was used to better align the features of body region. These features are obtained using ROI pooling framework. And different semantic level features are obtained separately. These semantic level features are combined with local body region features using tree structured fusion network.
     
    In \cite{zhu2020identity} weakly supervised approach semantic parsing was used to address the misalignment. Proposed framework identifies the person body parts and belongings at pixel level to achieve aligned person re-id. Pseudo-labels of human body parts were generated and refined using iterative mechanism. The framework can effectively identify the occluded parts in an image using foreground and background clustering approach. In this way only features of visible body parts are learned that leads to fine feature matching.
    Yifan Sun \emph{et al.} \cite{sun2019perceive} proposed a fully convolutional network- ResNet-50 \cite{he2016deep} named Visibility-aware Part Model (VPM) inspired from holistic person re-id \cite{sun2018beyond} \cite{kalayeh2018human}. Proposed design was self-supervised, focuses the challenge of partial person re-id, misalignment and scalability. It learns the features of shared region among two images using region locator and region extractor. In VPM holistic images are used for end-to-end training while at test time distance/visibility score of shared region-to-region and whole image is computed to obtain the meaningful results on synthetic as well as realistic datasets.
     
    In \cite{luo2019alignedreid++} Dynamically Matching Local Information (DMLI) aligns the horizontal strips without any explicit information of pose estimation. It also resolves the challenge of pose variation caused due to inaccurate detection, occlusion and change in viewpoints. Later they have combined the local branch that is DMLI with Aligned re-id++ (combination of local and global features) to learn the global features. Shortest path distance was used to align the local parts. In this manner local branch directs the global branch to learn more discriminative global features that results in improved performance. Irrespective of the fact that global level features have nearly global receptive field, they found that high-level feature maps are more suitable for aligning local parts.
     
    A framework named RankSVM proposed in \cite{zhao2016person}.This patch matching based computational model handles  the misalignment challenge caused due to factors like poses, viewpoints and illumination. This patch matching is integrated with saliency matching to increase the discriminative power and robustness. In this paper salience refers to regions with distinct attributes and are reliable to find same person in multiple camera views.They have transformed the original high dimensional visual featur space into saliency feature space that is 80 times less in dimension, this helped to attain high efficiency and resolve overfitting. Features of each  10*10 patch were extracted using Dense color histogram and Dense SIFT that is computed around local region.
     
    Sheng Li \emph{et al.} in \cite{li2017person} addressed the misalignment caused due to pose variations and viewpoint. In the proposed cross-view dictionary learning model, representation power of learned features is improved by learning dictionaries at multi-level \emph{i.e.} image-level, horizontal part-level, patch-level and is considered as a general solution to multi-level learning problem. Main idea is to exploit and captures the details at multiple levels that captures both local and global characteristic of an image, it jointly enhanced the performance results on challenges like misalignment, pose variations.
     
    Changxing Ding \emph{et al.} in \cite{ding2020multi} presented a multi-task part-aware network to handle the challenge of misalignment. Semantically aligned features of aligned parts are extracted in training phase. Model learns the representation specific to part by using regularization technique that result in selection of part specific channels. Global max pooling helped in making the learned features invariant to scale and translation. That fixed channels specified for each body part makes the approach robust at inference level to cope with blur and clutter may present in the image.
    Guanshuo Wang \emph{et al.} in \cite{wang2020receptive} focused on stripe-based approach to learn the features at multi-granular level. Instead of performing partition at input images or output features they have applied partition on intermediate representation to retain the local association among parts. Moreover, to deal with misalignment challenge random shifting augmentation was applied within bounding boxes appeared around detected persons in the image.
     
    Poor spatial alignment for video sequences was targeted in \cite{dai2018video}. Proposed framework was consists of two modules \emph{i.e.} temporal residual module and spatial-temporal transformer network module. Temporal module was responsible to obtain both generic and specific features present in consecutive frames while in second module semantic information of specific frame and its temporal context with adjacent frames were recorded. Model effectively align the person with major changes in appearance and hence outperforms the existing approaches.
    In \cite{li2020attributes} attributes are learned per patch. As a way local part features are extracted to handle the misalignment challenge. For final feature representation local features are fused with holistic features. hence effective results are obtained.
     
    Wei Shi \emph{et al.} in \cite{shi2022image} a deep Image to Video re-id pipeline was proposed. Fine-grained features are learnt using a three dimensional semantic appearance alignment module. Module extracts the images that are aligned with local appearance. These aligned images were then aggregated with another multi-branch network that helped in weakening the influence of occluded body parts. Moreover, another module was responsible to handle the modality misalignment problem, it ensures the interaction among global representation of an image and video streams as a result discriminative fine-grained features were learned to produce the outperforming results.
    An unsupervised way to tackle the challenge of unaligned video based person re-id was presented in \cite{ma2017person}. A video matching algorithm select and then match the image sequences that are inaccurate or incomplete. Proposed approach does not rely on external labelling hence applicable to large scale unseen data to produce better results.
     
    Patch wise metric learning approach proposed in \cite{bkak2017deep}. Appearance measures of each patch was learned and then combined using deformable models. Patches were allowed to change the locations so to resolve the correspondence issue. Even if patch locations were found different but each have the same metric score hence the proposed method effectively helped to increase the training data and produce the effective results.
    In addition to spatial alignment temporal alignment was challenged in \cite{liu2015spatio}. First they extracted the walking cycles (can be referred as gait) of a person in a chunk of video sequence. Each chunk was then divided into spatial and temporal data. Temporal sequence was then further divided into segments based on different walking cycle phases. While in the spatial domain image was divided into different body parts. Body-action unit was then formed on the basis of discriminative information obtained from temporal and spatial domains. Fisher vectors were extracted from body-action units to obtain a sort of generalized Bag-of-Words feature. Obtained features are then combined to form a vector that shows the visual/ appearance of walking person.
     
    Yang Shen \emph{et al.} \cite{shen2015person} presented a novel framework that handles the spatial misalignment caused due to multiple factors like camera-view and pose variations. Proposed framework learns the correspondence structure that include probabilities of matching patches among pair of cameras, this correspondence structure learning allows the better handling of misalignment. One-to-many graph in each correspondence structure of each patch allows to tackle the pose variations within each camera view, this graph shows the weights that depicts the matching probabilities of patches. Not only local but global context was also considered to achieve more reliable scores.
    Yang Shen \emph{et al.} \cite{shen2015person} presented a novel framework that handles the spatial misalignment caused due to multiple factors like camera-view and pose variations. Proposed framework learns the correspondence structure that include probabilities of matching patches among pair of cameras, this correspondence structure learning allows the better handling of misalignment. One-to-many graph in each correspondence structure of each patch allows to tackle the pose variations within each camera view, this graph shows the weights that depicts the matching probabilities of patches. Not only local but global context was also considered to achieve more reliable scores.
     
    Inter and intra local relationship among extracted features was maintained simultaneously in \cite{zhang2021part} using part-guided graph convolution network. For optimization purpose two types of graphs were formed that describe the relationship among adjacent parts. First is inter-local graph of same parts of person image and other one is intra-local graph of variant parts of a person image. At the graph convolution operation was performed to inject the representation of person images.
    In \cite{he2021dense} decoder and encoder based approach was used to handle the challenge of misalignment. Proposed model takes the advantage of both CNN and attention based architecture. On the basis of these architectures multi-grained spatio-temporal and positional spatio-temporal features are effectively learned to handle the misalignment.
    
    \subsubsection{Attention-based Approaches:}
    \label{subsubsec:MisalignmentAttentionBasedApproaches}
     Wei Li \emph{et al.} \cite{li2018harmonious} proposed a framework to resolve the challenge of alignment and representation learning. Proposed framework jointly learns the hard region-level attention along with soft pixel-level attention in specified bounding boxes to efficiently learn the feature representation.
     In \cite{chen2019abd} misalignment and background clutter challenges are addressed. They have combined multiple modules into a single module to extract the better feature maps. Their proposed design is able to obtain both channel wise and position/spatial information using channel and position attention modules. 
     In \cite{fang2019bilinear} a deep bilinear attention framework was formulated to handle the challenge of misalignment and representation learning. Two attention modules are introduced, decision for outer attention module \emph{i.e.} where to focus is taken by inner attention module. The module uses channel wise second order information and hence interdependency among global and local features is formed, in the meanwhile spatial information is also preserved. 
      
     Jianyuan Guo \emph{et al.} \cite{guo2019beyond} presented a supervised human body parts parsing approach to achieve efficient results on a challenge of alignment in person re-id. In the proposed framework information from both human and non-human parts was extracted. Parsing model was used to extract human parts aligned information while self-attention mechanism was used to group alike pixels. Parsing and attention modules are then combined to learn discriminative features for accurate human parts and coarse non-human parts.
     A unified deep supervised network to resolve the challenge of misalignment was presented in \cite{wang2018mancs}. Novel fully attention based block can be inserted into any convolution neural network to obtain the deep features. Channel-wise and spatial-wise attention information was extracted using fully attention block to better extract the useful multi-scale features. The problem of vanishing gradient  was also resolved because of deep supervision.
     Body part misalignment problem was also handled in \cite{zhao2017deeply}. They have decomposed the body parts into regions and computes the representation accordingly. This computation is discriminative in itself, hence helpful for person matching. The model learning was inspired by attention mechanism.
      
     In \cite{yang2019attention} challenge of misalignment was addressed with the help of pose estimation in a multi-branch network architecture. In the proposed attention based network, representation of body-part and whole-body is learnt. This representation is then fused on the basis of their contribution towards feature matching. Intra-attention directs to precisely learn the discriminative features of whole-body and body-part images. Whereas intra-attention simultaneously learn the optimal feature representation and attention maps for whole body and interested part of body.
     In \cite{wang2019cdpm} a novel end-to-end model for part alignment problem is proposed. Model not only detects the particular body parts but it also extracts the discriminative representation at part-level. They have divided the image vertically and horizontally to obtain the structural information without any alignment and clutter issue. Vertical module detects the body-parts while horizontal module learns the part representation using attention mechanism.
      
     In order to learn semantically aligned part-level features a simple batch-driven approach was proposed in \cite{wang2021batch}. Two modules were used \emph{i.e.} a guided attention channel and pair of regularization term. First module highlights the channel responsible for each part of person image in the output of the deep network while the other regularizer term that maintains the consistency among batches and hence make the process coherent and robust.
     In \cite{zhang2020learning} a novel triplet loss was proposed that not only considers the alignment but also pays attention to salient parts of the person image. It measure how much effort is deployed to align two distributions. Distribution of local parts were formed using attention technique. Weights are assigned to each part according to learned distribution. Novel loss term rectifies the assigned weights and hence an elegant solution to misalignment was presented.
      
     An attention mechanism was adopted in \cite{zheng2018pedestrian} that exhibit its focus on body parts rather than background. Alignment of person images was learned using identification procedure. Proposed mechanism effectively located the person in an image by placing a bounding box based on attention mechanism.
     Xinqian Gu \emph{et al.} in \cite{gu2020appearance} tried to resolve the temporal appearance misalignment in video based person re-id. Their appearance preserving framework is comprised of two parts. First part preserves the appearance at pixel level and other one is 3D convolution kernel that helps to model the temporal information. To ensure the temporal appearance alignment adjacent feature maps are reconstructed according to cross-pixel semantic similarity. An attention mask was also learned that finds the unmatched regions among reconstructed and central feature map. Learned attention mask was then imposed to avoid error propagation.
    {\small
    \begin{longtable} {m{1.5em}  m{9em} m{10em} m{5em}  m{4.5em}}
    \caption{Results obtained on Misalignment challenge against each dataset. Results in bold are the highest.\label{table:MisalignmentResults}}\\
    
    \hline\noalign{\smallskip}
     Sr.No & Paper & Dataset & R1/mAP & Code availability\\
    \hline\noalign{\smallskip}\hline\noalign{\smallskip}
    1 & STRF, 2021, \cite{aich2021spatio}  & MARS & 90.3/86.1 & No\\
      &      & DukeMTMC-VideoReID & 97.4/96.4 &        \\
      &      & iLIDS & 89.3/--  &        \\
    \hline\noalign{\smallskip}
    2 & DenseIL, 2021, \cite{he2021dense} & MARS & 90.8/87.0 & No\\
      &      & DukeMTMC-VideoReID & 97.6/97.1 &        \\
      &      & iLIDS & \textbf{92.0}/--  &        \\
    \hline\noalign{\smallskip}
    3 & ABD-Net, 2019, \cite{chen2019abd} & Market-1501 & 95.6/82.28 & Yes \\
      &      &  DukeMTMC-ReId & 89.0/78.59 &    \\
      &      &  MSMT-17  & 82.3/60.8 &  \\
    \hline\noalign{\smallskip}
     4 & BAT-net, 2019, \cite{fang2019bilinear} & Market-1501 & 95.1/87.4 & No \\
       &      & DukeMTMC-ReId & 87.7/77.3   &       \\
       &      & CUHK-03(Labeled) & 78.6/76.1  &        \\
       &      & CUHK-03(Detected) & 76.2/73.2  &        \\
       &      &  MSMT-17  & 79.5/56.8 &  \\
     \hline\noalign{\smallskip}
     5 & P2-Net, 2019, \cite{guo2019beyond} & Market-1501 & 95.2/85.6 & Yes \\
       &      & DukeMTMC-ReId & 86.5/73.1   &       \\
       &      & CUHK-03(Labeled) & 78.3/73.6  &        \\
       &      & CUHK-03(Detected) & 74.9/68.9  &        \\
     \hline\noalign{\smallskip}
     6 & PAHR, 2017, \cite{zhao2017deeply} & Market-1501 & 81.0/63.4 & No\\
        &      & CUHK-03 & 85.4/90.9  &        \\
        &      & VIPeR & 48.7/--  &        \\
     \hline\noalign{\smallskip}
     7 & BBA+PWM, 2015, \cite{shen2015person} & iLIDS & 44.3/-- & No\\
       &       & PRID-2011 & 64.1/-- &   \\
     \hline\noalign{\smallskip}
     8 & AP3D, 2020, \cite{gu2020appearance} & MARS & 90.7/85.6 & Yes\\
       &       & DukeMTMC-VideoReID & 97.2/96.1 &   \\
       &       & iLIDS & 88.7/-- &   \\
     \hline\noalign{\smallskip}
     9 & ISP, 2020, \cite{zhu2020identity} & Market-1501  & 95.3/88.6 & Yes \\
     &      & DukeMTMC-ReId   & 89.6/80  &        \\
     &      & CUHK-03(Labeled)  & 76.5/74.1  &       \\
     &      & CUHK-03(Detected)  & 75.2/71.4  &       \\
     \hline\noalign{\smallskip}
     10 & PCB, 2018, \cite{sun2018beyond}  & Market-1501  & 93.8/81.6 & No\\
     &      & DukeMTMC-ReId   & 83.3/69.2  &        \\
     &      & CUHK-03(Detected)  & 63.7/57.5  &       \\
     \hline\noalign{\smallskip}
     11 & MANCS, 2018, \cite{wang2018mancs} & Market-1501 & 93.1/82.3& No \\
     &      & DukeMTMC-ReId & 84.9/71.8  &        \\
     &      & CUHK-03(Labeled) & 69.0/63.9  &    \\
     &      & CUHK-03(Detected) & 65.5/60.5  &    \\
     \hline\noalign{\smallskip}
     12 & PABR, 2018, \cite{suh2018part} & Market-1501 & 95.4/93.1 & Yes \\
     &      & DukeMTMC-ReId & 84.4/69.3  &        \\
     &      & CUHK-01(Labeled) & 80.7/--  &    \\
     &      & CUHK-01(Detected) & 90.4/--  &    \\
     &      & CUHK-03(Labeled) & 91.5/--  &    \\
     &      & CUHK-03(Detected) & 88.0/--  &    \\
     &      & MARS & 84.7/75.9  &    \\
     \hline\noalign{\smallskip}
     13 & VAPM, 2019, \cite{sun2019perceive} & Market-1501 & 93.0/80.8 & No \\
     &      & DukeMTMC-ReId & 83.6/72.6  &        \\
     \hline\noalign{\smallskip}
     14 & EANet, 2019, \cite{huang2018eanet} & Market-1501 & 94.6/85.6 & Yes \\
     &      & DukeMTMC-ReId & 87.5/74.6  &        \\
     &      & CUHK-03(Detected) & 72.5/66.8  &    \\
     \hline\noalign{\smallskip}
     15 & DSA-reID, 2019, \cite{zhang2019densely} & Market-1501 & 95.7/87.6 & No \\
     &      & DukeMTMC-ReId & 86.2/74.3  &        \\
     &      & CUHK-01 & 90.4/--  &    \\
     &      & CUHK-03(Labeled) & 78.9/75.2  &    \\
     &      & CUHK-03(Detected) & 78.2/73.1  &    \\
     \hline\noalign{\smallskip}
     16 & DSR, 2018, \cite{he2018deep} & Market-1501 & 83.58/64.25 & No \\
    \hline\noalign{\smallskip}
     17 & HA-CNN, 2018, \cite{li2018harmonious} & Market-1501 & 91.2/75.7 & No \\
     &      & DukeMTMC-ReId & 80.5/63.8  &        \\
     &      & CUHK-03(Labeled) & 44.4/41  &    \\
     &      & CUHK-03(Detected) & 41.7/38.6  &    \\
     \hline\noalign{\smallskip}
     18 & SPReID, 2018, \cite{kalayeh2018human} & Market-1501 & 94.63/90.96 & Yes \\
     &      & DukeMTMC-ReId & 88.96/84.99  &        \\
     &      & CUHK-03 & 96.22/--  &    \\
     \hline\noalign{\smallskip}
     19 & Spindle Net, 2017, \cite{zhao2017spindle} & Market-1501 & 76.9/-- & Yes \\
     &      & CUHK-01 & 79.9/--  &    \\
     &      & CUHK-03 & 88.5/--  &    \\
     &      & iLIDS & 66.3/--  &    \\
     &      & PRID-2011 & 67/--  &    \\
     \hline\noalign{\smallskip}
     20 & CDPM, 2020, \cite{wang2019cdpm} & Market-1501 & 92.2/86.0 & No \\
     &      & DukeMTMC-ReId & 88.2/77.5  &        \\
     &      & CUHK-03(Labeled) & 81.4/77.5  &    \\
     &      & CUHK-03(Detected) & 78.8/73.3  &    \\
     \hline\noalign{\smallskip}
     21 & BCD-Net, 2020, \cite{wang2021batch} & Market-1501 & 97/92.7 & No \\
     &      & DukeMTMC-ReId & 91.1/81.6  &        \\
     &      & CUHK-03(Labeled) & 86.2/81.6  &    \\
     &      & CUHK-03(Detected) & 84.2/78.7  &    \\
     &      & MSMT-17 & 84.1/63.7  &    \\
     \hline\noalign{\smallskip}
     22 & PCB, 2020, \cite{zhang2020learning} & Market-1501 & 95.5/88.2 & No \\
     &      & DukeMTMC-ReId & 89.1/79.8  &        \\
     &      & CUHK-03(Labeled) & 72.5/69.8  &    \\
     &      & MSMT-17 & 81.4/59.7  &    \\
     \hline\noalign{\smallskip}
     23 & RMGL, 2020, \cite{wang2020receptive} & Market-1501 & 96.2/90.1 & No \\
     &      & DukeMTMC-ReId & 90.7/81.5  &        \\
     &      & CUHK-03(Labeled) & 20.8/18.7  &    \\
     \hline\noalign{\smallskip}
     24 & ST2N, 2019, \cite{dai2018video} & MARS & 80.5/69.1 & No \\
     &      & iLIDS & 57.7/--  &        \\
     \hline\noalign{\smallskip}
     25 & MPN, 2020, \cite{ding2020multi} & Market-1501 & 96.3/89.4 & No \\
     &      & DukeMTMC-ReId & 91.5/82.0  &        \\
     &      & CUHK-03(Labeled) & 85.0/81.1  &    \\
     &      & CUHK-03(Detected) & 83.4/79.1  &    \\
     \hline\noalign{\smallskip}
     26 & ADP, 2018, \cite{yang2019attention} & Market-1501 & 94.99/86.47 & No \\
     &      & DukeMTMC-ReId & 86.04/74.57  &    \\
     &      & CUHK-03(Labeled) & 96.43/--  &    \\
     &      & CUHK-03(Detected) & 93.58/--  &    \\
     \hline\noalign{\smallskip}
     27 & AlignedReID++ , 2019, \cite{luo2019alignedreid++} & Market-1501 & 92.8/89.4 & Yes \\
     &      & DukeMTMC-ReId & 86.2/82.8  &    \\
     &      & CUHK-03(Detected) & 67.9/70.7  &    \\
     &      & MSMT-17 & 69.8/43.7  &    \\
     \hline\noalign{\smallskip}
     28 & APDR, 2020, \cite{li2020attributes} & Market-1501 & 94.4/90.0 & No \\
     &      & DukeMTMC-ReId & 87.3/83.2  &    \\
     \hline\noalign{\smallskip}
     29 & 3D-SAA+CMIL, 2021, \cite{shi2022image} & MARS & 81.3/72.6 & No \\
     &      & DukeMTMC-VideoReId & 82.8/81.0  &    \\
     &      & iLIDS & 54.7/--  &    \\
     \hline\noalign{\smallskip}
     30 & PGCN, 2021, \cite{zhang2021part} & Market-1501 & 98.0/94.8 & No \\
     &      & DukeMTMC-ReId & 91.1/85.2  &    \\
     &      & CUHK-03(Labeled) & 86.7/83.6  &    \\
     &      & MSMT-17 & \textbf{87.7}/72.7  &    \\
     \hline\noalign{\smallskip}
     31 & TS-DTW, 2017, \cite{ma2017person} & iLIDS & 31.5/-- & No \\
     &      & PRID-2011 & 41.7/--  &    \\
     \hline\noalign{\smallskip}
     32 & DPML, 2017, \cite{bkak2017deep} & CUHK-01 & 75.9/-- & No \\
     &      & CUHK-03 & 84.0/--  &    \\
     &      & VIPeR & 51.7/--  &    \\
     &      & iLIDS & 82.2/--  &    \\
     \hline\noalign{\smallskip}
     33 & PAN, 2019, \cite{zheng2018pedestrian} & Market-1501 & 82.81/63.35 & No \\
     &      & DukeMTMC-ReId & 71.59/51.51  &    \\
     &      & CUHK-03(Labeled) & 36.86/35.03  &    \\
     &      & CUHK-03(Detected) & 36.29/34.0  &    \\
     \hline\noalign{\smallskip}
    \end{longtable}
    }

\subsection{Scale Difference}
    \label{subsec:scale}
     In real time scenario objects may appear in smaller or larger form. Their appearance may also depend on camera setting that makes the scale a complex challenge. Learning most discriminative features is the primary objective of re-id. These features should be computed at multiples scales so the re-id model become capable to differentiate two persons. Existing re-id model are based on fixed scale approach. A fixed scale representation makes most informative features blur due to this re-id model performance suffer. Existing approaches mostly resize all the pedestrian bounding box images into single scale. Fixed or single scale approach is not so optimal and explicitly multi scale representation becomes necessary as shown in Fig. \ref{fig:challenges}.
     
\begin{figure}[h]
        \includegraphics[width=\textwidth]{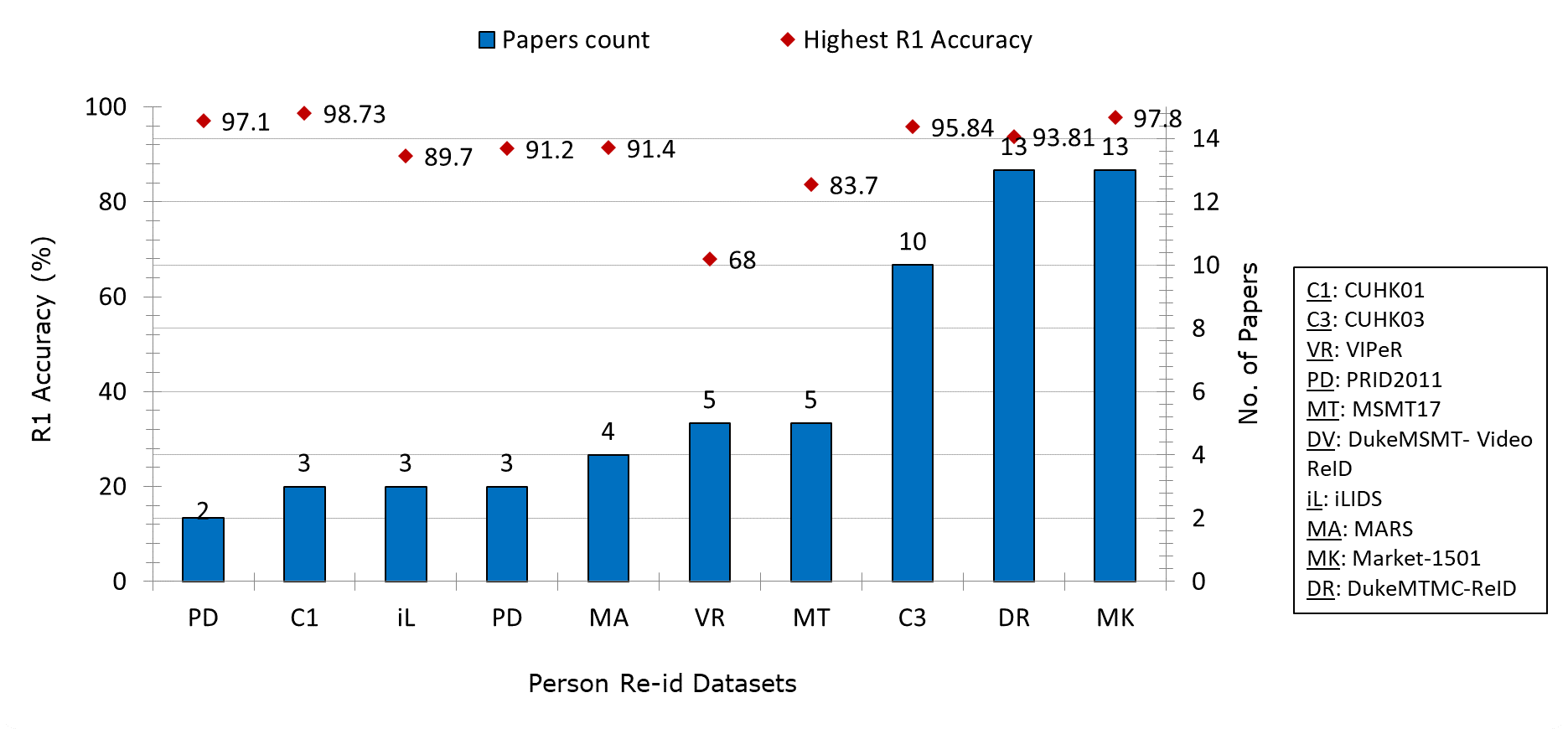}
        \caption{Progress on the challenge of scale variations for person re-id benchmarks}
        \label{fig:progressOnScale}
        \end{figure}

     People can be easily distinguished by using global features like gender and detecting local images patches but optimal matching only become possible if the features are computed at multiple scales and combined \cite{chen2017person}. In open surveillance scene images are captures at an arbitrary scale, this make it challenging to learn correlations among features of different scales. \cite{qian2017multi} based on Siamese network and capable to learn most discriminative and informative features at different scales and evaluate their importance for cross-camera matching.  In past five years progress achieved on this challenge is shown in Fig. \ref{fig:progressOnScale}
   
    \subsubsection{CNN-based Approaches:}
    \label{subsubsec:ScaleCNNBasedApproaches}
    A coarse-to-fine model \cite{zheng2019pyramidal} based on 3 dimensional sub-maps captures the discriminative information at different scales hence scalability challenge is addressed effectively.
    In \cite{zhou2019omni} both inter-class and intra-class variations are taken as challenge and resolved using omni-scale feature extraction based on different receptive fields. Proposed network is lightweight due to usage of point-wise and depth-wise convolutions instead of standard convolution. Unified Aggregation (AG) Gate dynamically combines the generated channel wise weights of multi-scale feature maps. AG share parameters along all the streams and hence network enables to learn discriminative features at heterogeneous scales.
     
    In \cite{chen2017person} multi-scale feature learning problem is resolved using an end-to-end Deep Pyramidal Feature Learning CNN architecture. Model concurrently learns the scale-specific deep features. Effective results are obtained using multiple classification losses. There are scale specific branches that are correlated with each other (to maximise the scale specific discriminative features) and regularisation mechanisms that makes the model more effective for multi-scale issues.
    In \cite{liu2021spatial} issue of multi-scale along with pose estimation was resolved using novel Spatial-Temporal Correlation and Topology Learning framework (CTL). The work was done for video-based person re-id. In order to obtain the diverse discriminative semantics local features were learnt at multi-granularity levels.
     
    For video-based person re-id a novel 3D multi-scale convolution layer was introduced in \cite{li2020multi} that can be inserted into any existing 2D convolution network. Depending on the location of insertion of 3D layer it splits into two variants \emph{i.e.} local and global. Local learns the spatial-temporal cues among adjacent 2D feature maps while global learns the temporal relation among features of adjacent frames at global level. Together these features form the strong multi-scale combination and hence model got the immersive capability to learn discriminative features.
    An unsupervised generalized cross-dataset omni-scale approach was presented in \cite{zhou2021learning}. In the proposed framework an omni-scale network learn the features at multiple spatial scales with assigned channel-wise weights to produce outperforming results.
     
    In \cite{zhao2020similarity} a novel similarity learning model was presented. In the model they have combined the feature optimization using multi-view visual words and metric optimization. k-means clustering method was used to capture he multi-view visual words. A similarity function was then used to combine the common subspaces.
    Ancong Wu \emph{et al.} \cite{wu2019distilled} presented the scalable adaptive framework to address the challenge of scalability in unsupervised manner. In the presented approach source dataset was trained on labelled dataset however no labelling was done for practical testing. Teacher student transfer learning technique was adopted to obtain the effective results from trained model.    
                 
    \subsubsection{Attention-based Approaches:}
    \label{subsubsec:ScaleAttentionBasedApproaches}
    In \cite{guo2018efficient} visual similarities among images have been measured at different scales in an end-to-end manner. Moreover meaningful information have been extracted from feature map of an image using attention based spatial transformer Siamese network.
    A multi-scale model was presented in \cite{yan2021bv} that is cross attention-based. It is able to learn the discriminative information of different body parts of specific identity from multiple views. A new large scale Bird-View dataset was also proposed and tested along with existing benchmark datasets.
     
    In \cite{qian2020leader} deep re-id network is proposed. Two novel layers are introduced to handle multi-scale challenge. Multi-scale deep learning layer learns the discriminative features at multiple scales. And leader based attention learning layer takes information of multiple scales and use this to selectively learn the optimal weightage assigned to each scale. Moreover pair of classification losses are used that strengthen the process of feature learning at multi-scale both at local and global level. Proposed methodology was useful due to the use of features that were extracted at multiple scales and locations and this makes the model generalised as some of the features become transferable.
     
    Wei Zhang \emph{et al.} in \cite{zhang2019multi} proposed a framework for video based person re-id. In presented framework local regions are taken into account at multiple scales. Weights are assigned to local regions using attention mechanism at both spatial and temporal levels. These local features are aggregated to form rich representation of video, resulted in effective final outcome.
    A video-based person re-id approach was presented in \cite{yang2021two}. A thorough spatial and temporal representation was obtained using two branches \emph{i.e.} pyramid dilated convolution and pyramid attention pooling. Pyramid based strategy helped in extracting multi-scale features that has helped to mitigate the quality problems might present in the video \emph{i.e.} partial occlusion.
    In \cite{martinel2020deep} multi-scale pooled regions are fed into a novel deep architecture to extract the discriminative features at multiple scale semantic levels. Approach was made possible by using attention mechanism and was inspired by pyramid based methods.
     
    Another multi-scale attention pyramid method to mitigate the scale challenge was presented in \cite{chen2021person}. First features were divided into multiple local parts and then learned using attention mechanism. Then these features were merged and stacked using residual connection to form an attention pyramid. This attention pyramid was implemented in both channel-wise and spatial attention modules and hence reported the outperforming results as compared to existing state of the art methods.
    Another approach to learn complementary features was presented in \cite{zhong2021progressive}. Features were learnt from deep to shallow layers in a progressive manner. To focus on the layer specific features a two stage attention was also introduced to filter the noisy feature maps.
     
    An end-to-end approach \cite{huang2020multiscale} to extract holistc and local feature maps using multi-scale omnibearing attention network. Multi-sized convolutions were used to obtain the local and holistic feature maps. Two kinds of attentions \emph{i.e.} spatial and channel were also incorporated to extract the comprehensive feature representation.
    
    {\small
    \begin{longtable} {m{1.5em}  m{9em} m{10em} m{5em}  m{4.5em}}
    \caption{Results obtained on Scale challenge against each dataset. Results in bold are the highest.\label{table:ScaleResults}}\\
    
    \hline\noalign{\smallskip}
     Sr.No & Paper & Dataset & R1/mAP & Code availability\\
    \hline\noalign{\smallskip}\hline\noalign{\smallskip}
    1 & BV-Person, 2021, \cite{yan2021bv}  & Market-1501 & 96.0/89.2 & No\\
      &      & DukeMTMC-ReID & 90.5/80.6 &        \\
    \hline\noalign{\smallskip}
    2 & OSNet, 2019, \cite{zhou2019omni} & Market-1501 & 94.8/84.9 & Yes\\
      &      & DukeMTMC-ReID & 88.6/73.5 &        \\
      &      & CUHK-03 (Detected) & 72.3/67.8 &        \\
      &      & VIPeR & 68.0/--  &        \\
      &      & MSMT-17 & 78.7/52.9     &        \\  
    \hline\noalign{\smallskip}
    3 & MSDL, 2017, \cite{qian2017multi} &  CUHK-01 & 79.01/-- & Yes \\
      &      &  CUHK-03 & 75.64/-- &    \\
      &      &  VIPeR  & 43.03/-- &  \\
      &      &  iLIDS  & 41.0/-- &  \\
      &      &  PRID-2011  & 65.0/-- &  \\
    \hline\noalign{\smallskip}
     4 & DPFL, 2017, \cite{chen2017person} & Market-1501 & 92.3/80.7 & No \\
       &      & DukeMTMC-ReId & 79.2/60.6   &       \\
       &      & CUHK-03(Labeled) & 86.7/82.8  &        \\
       &      & CUHK-03(Detected) & 82/78.1  &        \\
     \hline\noalign{\smallskip}
     5 & CTL, 2021, \cite{liu2021spatial} & MARS & 91.4/86.7 & No \\
     &      & iLIDS & 89.7/--  &        \\
     \hline\noalign{\smallskip}
     6 & CFPModel, 2019, \cite{zheng2019pyramidal} & Market-1501 & 95.7/88.2 & No\\
       &      & DukeMTMC-ReId & 89.0/79.0  &        \\
       &      & CUHK-03(Labeled) & 78.9/76.9  &        \\
       &      & CUHK-03(Detected) & 78.9/74.8  &        \\
     \hline\noalign{\smallskip}
     7 & AKA, 2019, \cite{wu2019distilled} & Market-1501 & 49.7/24.6 & No \\
        &      & DukeMTMC-ReId & 47.6/31.1  &        \\
     \hline\noalign{\smallskip}
     8 & STNs, 2018, \cite{guo2018efficient} & CUHK-01 & 88.2/-- & No\\
       &       & CUHK-03(Labeled) & 87.5/-- &   \\
       &       & CUHK-03(Detected) & 86.45/-- &   \\
       &       & VIPeR   & 50.1/--  &        \\
     \hline\noalign{\smallskip}
     9 & MSTA, 2020, \cite{zhang2019multi} & MARS  & 82.28/69.42 & No \\
     &      & iLIDS   & 70.1/--  &        \\
     &      & PRID-2011   & 91.2/--  &        \\
     \hline\noalign{\smallskip}
     10 & DPRM, 2021, \cite{yang2021two}  & MARS  & 89.0/83.0 & No\\
     &      & DukeMTMC-VideoReId & 97.1/95.6  &        \\
     \hline\noalign{\smallskip}
     11 & PyrAttNet, 2020, \cite{martinel2020deep} & Market-1501 & 97.8/95.8 & Yes \\
     &      & DukeMTMC-ReId & 93.0/90.9  &        \\
     &      & CUHK-03 & 86.8/88.0  &    \\
     \hline\noalign{\smallskip}
     12 & M3D-CNN, 2020, \cite{li2020multi} & MARS & 88.87/85.46 & No \\
     &      & DukeMTMC-VideoReId & 95.49/93.67  &        \\
     \hline\noalign{\smallskip}
     13 & APNet, 2021, \cite{chen2021person} & Market-1501 & 96.2/90.5 & Yes \\
     &      & DukeMTMC-ReId & 90.4/81.5  &        \\
     &      & CUHK-03 & 87.4/85.3  &    \\
     &      & MSMT-17 & 83.7/63.5  &    \\
    \hline\noalign{\smallskip}
     14 & PFE, 2021, \cite{zhong2021progressive} & Market-1501 & 95.2/87.5 & No \\
     &      & DukeMTMC-ReId & 89.2/77.1  &        \\
     &      & CUHK-03 & 74.0/71.1  &    \\
     &      & MSMT-17 & 82.0/56.2  &    \\
     \hline\noalign{\smallskip}
     15 & PREST, 2021, \cite{zhang2021self} & Market-1501 & 82.5/62.4 & No \\
     &      & DukeMTMC-ReId & 74.4/56.1  &        \\
     \hline\noalign{\smallskip}
     16 & MuDeep, 2020, \cite{qian2020leader} & Market-1501 & 95.34/84.66 & No \\
     &      & DukeMTMC-ReId & 88.19/75.63  &        \\
     &      & CUHK-01 & \textbf{98.73}/--  &    \\
     &      & CUHK-03 (Labeled) & 95.84/--  &    \\
     &      & CUHK-03 (Detected) & 93.70/--  &    \\
     \hline\noalign{\smallskip}
     17 & OSNet, 2021, \cite{zhou2021learning} & Market-1501 & 94.8/86.7 & Yes \\
     &      & DukeMTMC-ReId & 88.7/76.6  &        \\
     &      & CUHK-01 & 86.6/--  &    \\
     &      & CUHK-03 (Labeled) & 72.3/67.8  &    \\
     &      & VIPeR & 68.0/--  &    \\
     &      & MSMT-17 & 79.1/55.1  &    \\
     \hline\noalign{\smallskip}
    18 & IALM, 2020, \cite{zhao2020similarity} & CUHK-01 & 68.44/-- & No \\
     &      & VIPeR & 56.32/--  &        \\
     \hline\noalign{\smallskip}
    19 & MOAN, 2020, \cite{huang2020multiscale} & Market-1501 & 97.45/96.42 & No \\
     &      & DukeMTMC-ReId & 93.81/92.82  &        \\
     &      & CUHK-03 (Labeled) & 90.07/90.32  &    \\
     &      & MSMT-17 & 81.53/58.02  &    \\
     \hline\noalign{\smallskip}
    \end{longtable}
    }
\subsection{Viewpoint Variance}
    \label{subsec:viewpoint}
    Viewpoint is most important challenge in person re-id problem because different views of a pedestrian across non-overlapping cameras contain different information. Viewpoint support the learning algorithm specifically in identifying the pedestrian. Finding the particular angles are important for learning models to identify the person. To develop a model with outstanding generalizability is difficult because the visual appearance of the same person varies across different views across multiple cameras. Recent studies mainly focus on the learning of view invariant features. They initially extract view-generic features and after that view invariant model is learned. It is to reduce the distance between the intra class subjects and increase the gaps between the inter class pedestrians. Fig. \ref{fig:challenges} shows the viewpoint variation challenge.
     
\begin{figure}[h]
        \includegraphics[width=\textwidth]{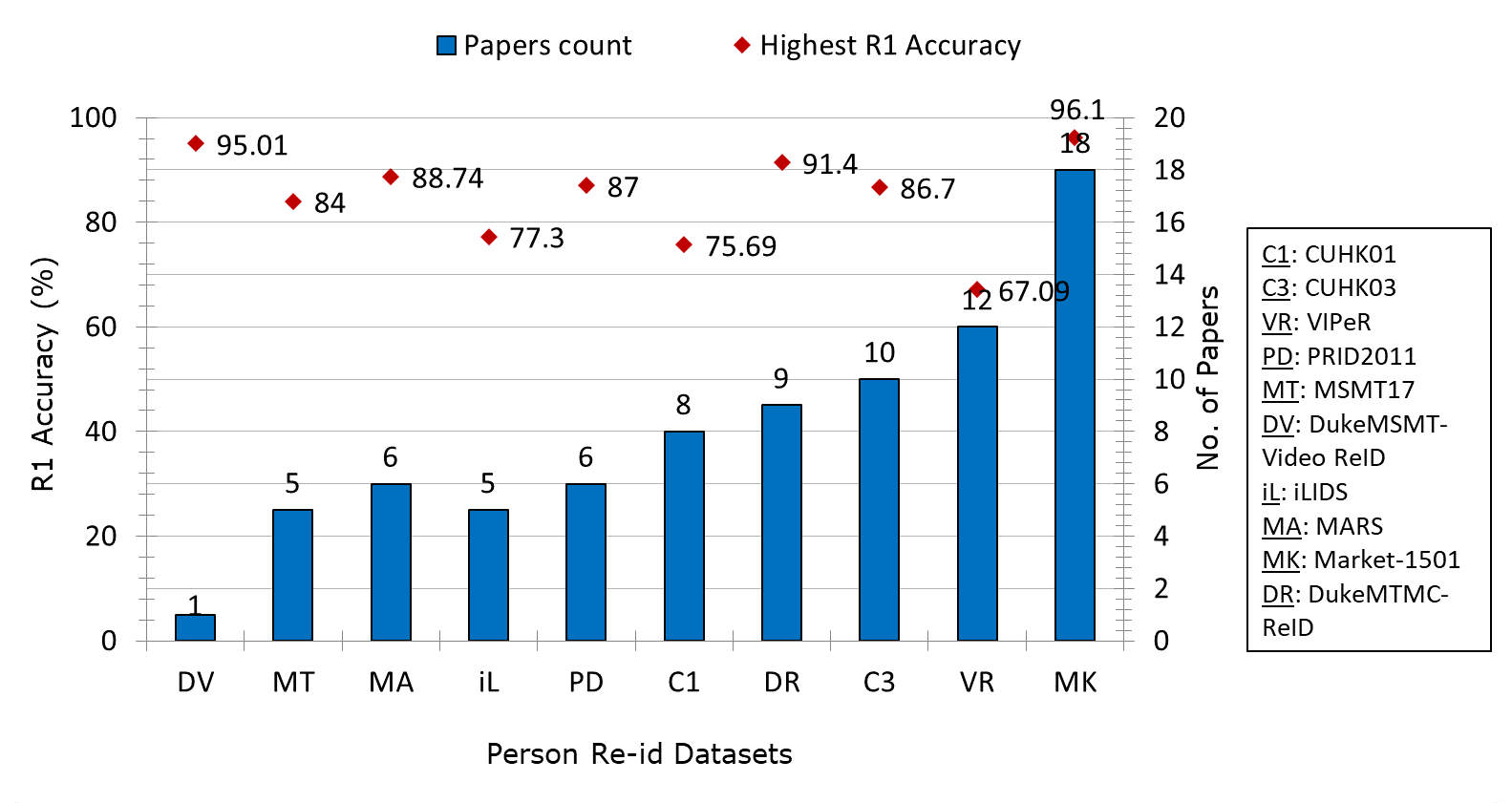}
        \caption{Progress on the challenge of viewpoint variation for person re-id benchmarks}
        \label{fig:progressOnViewpointvariation}
        \end{figure}

    But there are limitations in existing approaches, when the complex changes occur in the visual appearance between non-overlapping cameras, the view generic features become inadequate in solving person re-id task. Another problem in resolving this challenge is the lack of data and some datasets contain fixed and insufficient distribution of environmental factors \emph{e.g.} in pedestrian viewpoint, some angles might contain few or zero samples. Due to importance of this challenge each year many papers were published in journals and conferences as well. The progress of work done so far on viewpoint challenge is shown in Fig. \ref{fig:progressOnViewpointvariation}
        
    \subsubsection{CNN-based Approaches:}
    \label{subsubsec:ViewpointCNNBasedApproaches}
    Xiaoxiao sun \emph{et al.} \cite{sun2019dissecting} designed the practical solution by formulating the synthetic dataset PersonX for subjective study of various challenges \emph{i.e.} occlusion, viewpoint variations, illumination changes, poses and various backgrounds. They have used IDE+ with ResNet-50 \cite{he2016deep} as backbone with 36 angles and pre-trained ImageNet weights \cite{krizhevsky2012imagenet}. They have concluded with controlled experimentation that person with side views makes better query.
     
    In \cite{karanam2015person} single dictionary was learned for both gallery and probe images simultaneously to obtain the view-invariant feature vectors of different people. To have discriminative dictionary encoding, model is then explicitly trained by applying constraint on association of sparse representations of the feature vector. At test time gallery images are compared (using Euclidean sense) to find the closest sparse match representation.
    In \cite{liao2015person} efficient feature representation was learned. Presented method analyzes the presence of local features in horizontal direction. Occurrences of local features are maximized to make the approach viewpoint invariant. Further they have applied Retinex transform to reduce the impact of illumination and have reported the improved results.
     
    In \cite{porrello2020robust} visual variety was handled using teacher-student framework,  teacher guides the student regarding multiple views and as a result student resulted in state-of-the-art in Image-To-Video by large margin. 
    In paper \cite{dai2018cross} a novel cross-view semantic projection learning algorithm was proposed to model the feature transformation for person re-id. It retrieves the latent intrinsic invariant representation of persons. Three components are learnt simultaneously \emph{i.e.} shared basis matrix, pair of semantic projection functions and optimal association function. In training phase the shared basis matrix explores the intrinsic structure of raw descriptors that are from different camera views. The projection function maps the original hand crafted features into common semantic space in the training phase. And the optimal association function captures the best association between semantic representation of alike individuals from multiple cross-views.Algorithm was then also generalised to multiple views datasets in the same paper.
     
    Ying-Cong Chen \emph{et al.} proposed a asymmetric distance learning model \cite{chen2016asymmetric} to tackle the multi-camera view challenge. It transforms the different features that corresponds to different views to a common space. The model learns the camera-specific projection to resolve the problem feature discrepancy. Furthermore, consistency regularization helped in modeling the correlation among different views.
    Ziyan Wu \emph{}{et al.}in \cite{wu2015viewpoint} proposed a metric learning based viewpoint invariant algorithm to re-identify persons in cameras with disjoint field of view. It takes into account pose information prior from training data. After rectification color and texture histograms were used as discriptors to feature extractor from an image that was divided into six strips.
    Similarity measure is then used to assign the probe image to target class.
     
    In \cite{chen2018person} viewpoint specific approach named CRAFT was used for both view-generic and view-specific scenarios. Model was capable to adopt camera view features based on cross-view correlation and adaptive feature augmentation, this helps in transforming original features to new augmented space. Via this augmented framework view-generic can be induced to view-specific sub-models. To control the degree of correlation among sub-models camera view discrepancy regularization was also introduced in the CRAFT framework. To make the approach view invariant and generic to target datasets potential of deep learning techniques were explored and experimented.
    A view-specific model presented in \cite{feng2018learning} to handle the intra-class variations caused by viewpoints variations. Its two main contributions were: Cross-View Euclidean Constraint (CV-EC) and Cross-View Center Loss (CV-CL). CV-EC reduces the distance between features of same person from non-overlapping camera views. Whereas CV-CL was integrated to gain the high discriminative ability of view-specific deep networks. Proposed architecture can also deal with the applications using more than two cameras \emph{i.e.} multi-view version.
     
    One of the sever difficulty in re-id arises due to view variations in non-overlapping camera views. Hence a robust and discriminative descriptor is required to handle such challenge and it is resolved in \cite{chen2016deep}. This \cite{chen2016deep} novel deep learning to rank framework reduced the ranking cost of gallery images. In training phase, ranking model forms the relationship among input image pairs and their similarity scores via joint representational learning directly from raw image pixels. CNNs are used to form the relation between pair of images and their similarity score. In CNN feature representation and metric learning are integrated and generate the similarity score directly without relying on separate Euclidean or cosine distance measurement. Hence no need to compute feature representations separately. Network then learns the transformation in each ranking unit that tends to assign the highest score to the true match.
     
    In \cite{jia2017multiple} multiple metric learning method was presented that was based on cross-view quadratic discriminant analysis algorithm. The importance of each feature was decided using the discriminative power of each feature. Weights of all features are learned simultaneously using SVM learning framework. In this way gallery images were ranked based on the maximum attained score.
    Lin Wu \emph{et al.} in \cite{wu2018structured} proposed a deep hashing framework that was fast and contains the discriminative representation of pedestrians. Hamming distance was then computed to rank the similar images closer to each other. Hash codes were assigned to each cluster of similar images to make the retrieval process faster.
    In \cite{chen2021bidirectional} multiple convolution feature were extracted on the behalf of different body parts. Inter-layer interactions were also exploited to obtain the discriminative representation of person identities.
     
    Alessandro Borgia \emph{et al.} presented a metric learning approach in \cite{borgia2018cross}. More discriminative feature space was learned using novel loss function. Inter-class and intra-class relationship of person identities was steered by meta center term and centers dispersion term. 
    Inter-identities' interference was reduced to obtain a more expanded feature space.
    An unsupervised way to approach the challenge of visual ambiguities was presented in \cite{garcia2017discriminant}. They have obtained the content and context from initial ranked lists and then removed the visual ambiguities from them  to have the optimal feature representation and hence the final new rank list that was better than the initial rank list as tested on publically available datasets.
     
    A view-invariant video-based few-shot learning method was proposed in \cite{wu2020few}. Proposed method was developed on variational recurrent neural network that was trained adversarially to produce view-invariant feature space for matching persons.
    Another approach to mitigate the challenges \emph{i.e.} viewpoint variation, low resolution and pose was presented in \cite{huang2020improve}. A lightweight and labelled part segmentation head was added to the backbone of re-id during training process to obtain diverse set of features hence resulted in improved performance of re-id.
     
    In \cite{lin2020unsupervised} an unsupervised cross-camera learning mechanism was adopted to achieve the generalized results. An unsupervised style transfer model was responsible to generate the images with transferred-style and different camera styles. Similarity was measured among generated and the original image. Similar images were then grouped together to form one cluster using iterative approach. A regularization term was also invoked to balance the cluster distribution. Results on benchmark datasets demonstrates the superior performance of the proposed approach.
    An asymmetric metric learning method was proposed in \cite{meng2019deep} to alleviate the view-bias problem. It was a two stream deep neural network that jointly learns the view and feature specific transformations. Clustering based deep asymmetric metric learning method was adopted to make the solution scalable.
     
    A novel descriptor was presented in \cite{zhao2018maximal} for effective feature representation and metric learning. In order to have generalized multi-view the descriptor captures the structural information, analyzes and maximized the horizontal occurrences of multi-granularity to extract the rich feature representation even in case of drastic change in viewpoint. Besides the descriptor a metric learning method was also presented that jointly learns the multiple view-specific linear transforms to obtain the robust features. Descriptor and the metric learning method were then jointly evaluated on publicly available datasets to prove the effectiveness of the approach.
    A pragmatic semi-supervised framework was introduced in \cite{jia2020view} to address the issue of view-specific biases. Framework learns the view-specific projections against each view. Only limited labeled data was used for training purpose. To boost the performance a re-ranking strategy was also introduced in the paper that measures the similarity among probe and gallery images and re-rank them based on their overlapping ratio. Framework has yielded superior performance when tested on mostly used re-id datasets in both supervised and semi-supervised way.
     
    An adaptive multi-projection metric-learning method was introduced in \cite{hu2019adaptive} to handle the inconsistencies among different camera views. Proposed metric learning method jointly learns the different camera projections into a common feature space. Proposed method successfully adopts the newly added camera projections without updating the existing projection matrices. Notable improvements were observed when applied on major re-id datasets.
    A view-invariant subspace was learnt in \cite{wu2020cross} using adversarial approach. Specifically, coupled asymmetric mapping was learnt. View discrepancy was resolved by optimizing by cross-entropy view-specific objective. A similarity discriminator was introduced to determine the similarity value to distinguish the negative and positive pairs. To handle the imbalance of identity pairs caused due to most difficult samples adaptive weighing was also implemented.
     
    Another deep multi-view feature learning method presented in \cite{tao2017deep}. Proposed metric learning based method exploits the fusion of handcrafted and deep learning features to produce the discriminative feature representation.
    An unsupervised framework for video-based person re-id was presented in \cite{wang2020exploiting}. Relation of frame with its first neighbour was explored to form the group in each camera. Cross-view matching strategy then find the matching relationship among them. And finally metric model for each camera pair was learnt in a progressive manner.
     
    An approach was designed in  \cite{zhang2018learning} to address the variations exist in same video. Mainly a new loss term was defined comprised of of intra-video loss and Siamese loss. Intra-video loss make the video more clustered by using the mean-body of each camera viewpoint. And Siamese loss placed the wrong matches more apart. Generalization capability of the model was increased as network was trained in iterative manner and hence the mean-body weights were updated accordingly.
    In \cite{an2018multi} data discrepancy among multiple views was minimized using the proposed multi-level learning framework in iterative manner. Synthetic data was generated using already available grouping information and this data was then viewed as transitional state among original camera views. Gallery and probe images were moved into common subspace in a progressive manner to perform the matching step.
     
    A memory module was proposed in \cite{zhong2019invariance} with the purpose to make the system invariant in terms of camera viewpoint and neighbourhood changes. They have used specifically unlabelled dataset to learn unsupervised discriminative representation and to make model domain invariance. ResNet-50 \cite{he2016deep} as backbone with pre-trained weights on ImageNet \cite{krizhevsky2012imagenet}.
    An unsupervised approach to learn asymmetric learning of cross-view person images was presented in \cite{yu2017cross}. For each view model learns the specific projection, based on asymmetric clustering. In order to achieve the better matching performance, model finds the shared space with low view-specific bias.
     
    Zimo Liu \emph{et al.} \cite{liu2017end} proposed an unsupervised tracklet based framework to learn cross-view discriminative features. They have used tracklet as query in search for nearest neighbour best match after possible iteratios until the best match found. In order to reduce the impact of false positive matches they have employed hard negative mining. KNN searching was then repeated in a reverse manner to ensure the best match found. Best query matches at initial stage and at reverse stage are then used collectively to update the model.
     
    In \cite{yu2020unsupervised} an unsupervised re-id framework to cope the challenge of viewpoint variations. Deep Clustering-based Asymmetric Metric Learning (DECAMEL) learns an initial asymmetric metric using a linear unsupervised model \emph{i.e.} CAMEL. It then embeds the learned metric into deep network by jointly learning metric and features in an end to end manner. Afore mentioned steps were based on asymmetric metric clustering. Novel loss function was then applied to achieve the best results. Sub-optimality in the results was obtained on the basis of separation of metric and feature learning. Due to learning of better cross-view clusters in the shared space better cross-view matching performance was achieved.
    
    \subsubsection{Attention-based Approaches:}
    \label{subsubsec:ViewpointAttentionBasedApproaches}
    Meng Zheng\emph{et al.} \cite{zheng2019re} has proposed a Siamese architecture to address the challenge of viewpoint. In proposed joint learning end-to-end architecture a flexible attention mechanism was introduced to achieve the attention consistency among the images of same person.
    Lei Zhang \emph{et al.} in \cite{zhang2021adversarial} to address cross-view challenges. In the proposed end-to-end framework view-invariant features were learnt and comprised of three components \emph{i.e.} adversarial learning, drawing same features towards center and SIFT guidance. To improve the integration of these components attention mechanism was also adopted to produce the superior results.
     
    In \cite{lian2021attention} attention aligned network was presented that focuses on foreground information using channel wise multi-scale attention aware mechanism that had helped in learning the invariant views obtained from different cameras. To increase the capability of feature learning an improved triplet loss was also presented. Resulted in improved results by maximizing the inter-class distance and minimizing the intra-class distance.
     
    \subsubsection{\textbf{GAN-based Approaches:}}
    \label{subsubsec:ViewpointGANBasedApproaches}
    To learn the view-invariant features GAN and another contrastive learning module was combined into one training framework in \cite{chen2021joint}. Novel views are generated using mesh based view generator. Proposed method was flexible as the model does not rely on labeled source domain. Improved results are obtained as compared to existing fully unsupervised and unsupervised approaches.

    
    {\small
    \begin{longtable} {m{1.5em}  m{9em} m{10em} m{5em}  m{4.5em}}
    \caption{Results obtained on viewpoint variation challenge against each dataset. Results in bold are the highest.\label{table:ViewpointVariationResults}}\\
    
    \hline\noalign{\smallskip}
     Sr.No & Paper & Dataset & R1/mAP & Code availability\\
    \hline\noalign{\smallskip}\hline\noalign{\smallskip}
    1 & CAMEL, 2017, \cite{yu2017cross}  & Market-1501 & 54.5/26.3 & No\\
    &      & CUHK-01  & 61.9/57.3 &        \\
    &      & CUHK-03 & 39.4/31.9 &        \\
    &      & VIPeR & 30.9/--  &        \\
    \hline\noalign{\smallskip}
    2 & LOMO+GOG, 2017, \cite{liu2017end} & MARS & 23.9/-- & No\\
    &      & iLIDS & 41.7/-- &        \\
    &      &  PRID-2011  & 80.9/-- &  \\
    \hline\noalign{\smallskip}
    3 & DVDL, 2015, \cite{karanam2015person} &  iLIDS & 25.9/-- & No \\
    &      &  PRID-2011  & 40.6/-- &  \\
    \hline\noalign{\smallskip}
    4 & VKD, 2020, \cite{porrello2020robust} & MARS & 88.74/82.22 & Yes \\
    &      & DukeMTMC-VideoReId & 95.01/93.41  &        \\
    \hline\noalign{\smallskip}
    5 & GCL, 2021, \cite{liu2021spatial} & Market-1501 & 90.5/75.4 & Yes \\
    &      & DukeMTMC-ReId & 81.9/67.6  &        \\
    &      & MSMT-17 & 54.4/29.7  &        \\
    \hline\noalign{\smallskip}
    6 & PersonX, 2019, \cite{sun2019dissecting} & Market-1501 & 93.0/80.8 & No\\
    &      & DukeMTMC-ReId & 83.6/72.6  &        \\
    \hline\noalign{\smallskip}
    7 & ECN, 2019, \cite{zhong2019invariance} & Market-1501 & 63.3/40.4 & No\\
    &      & DukeMTMC-ReId & 63.3/40.4  &        \\
    &      & MSMT-17 & 30.2/10.2  &        \\
    \hline\noalign{\smallskip}
    8 & CASN, 2019, \cite{zheng2019re} & Market-1501 & 94.4/82.8 & No\\
    &      & DukeMTMC-ReId & 87.7/73.7  &        \\
    &       & CUHK-03(Labeled) & 73.7/68 &   \\
    &       & CUHK-03(Detected) & 71.5/64.4 &   \\
    \hline\noalign{\smallskip}
    9 & LOMO-XQDA, 2015, \cite{liao2015person} & CUHK-03(Labeled)  & 52.25/-- & Yes \\
    &      & CUHK-03(Detected)  & 46.25/--  &        \\
    &      & VIPeR   & 40/--  &        \\
    \hline\noalign{\smallskip}
    10 & FC2, 2018, \cite{wu2018structured}  & Market-1501  & 48.06/-- & No\\
    &      & CUHK-03 & 37.41/--  &        \\
    \hline\noalign{\smallskip}
    11 & ICV-ECCL, 2018, \cite{feng2018learning} & Market-1501 & 90.6/77.3 & Yes \\
    &      & CUHK-01 & 83.5/--  &    \\
    &      & CUHK-03 & 88.6/--  &    \\
    &      & VIPeR   & 51.9/--  &        \\
    \hline\noalign{\smallskip}
    12 & BINet, 2021, \cite{chen2021bidirectional} & Market-1501 & 95.3/88.7 & No \\
    &      & DukeMTMC-VideoReId & 91.4/81.3  &        \\
    &      & CUHK-03(Labeled) & 73.6/72.5  &    \\
    &      & CUHK-03(Detected) & 72.3/69.8  &    \\
    &      & MSMT-17 & 76.1/52.8  &    \\
    \hline\noalign{\smallskip}
    13 & SMC-ECD, 2018, \cite{borgia2018cross} & Market-1501 & 80.31/59.68 & No \\
    \hline\noalign{\smallskip}
    14 & DCIA, 2021, \cite{zhong2021progressive} & PRID-2011 & 32.5/-- & No \\
    &      & VIPeR & 64.78/--  &        \\
    \hline\noalign{\smallskip}
    15 & VRNNs, 2020, \cite{wu2020few} & MARS & 61.2/52.1 & No \\
    &      & iLIDS & 64.6/--  &        \\
    \hline\noalign{\smallskip}
    16 & MGN, 2020, \cite{huang2020improve} & Market-1501 & 95.8/88.7 & No \\
    &      & DukeMTMC-ReId & 90/79.9  &        \\
    &      & CUHK-03 (Labeled) & 78.8/74.4  &    \\
    &      & MSMT-17 & 84.0/62.4  &    \\
    \hline\noalign{\smallskip}
    17 & UDA, 2020, \cite{lin2020unsupervised} & Market-1501 & 73.3/38.0 & No \\
    &      & DukeMTMC-ReId & 56.1/30.6  &        \\
    \hline\noalign{\smallskip}
    18 & VIH-ReID, 2015, \cite{wu2015viewpoint} & VIPeR & 21.4/-- & No \\
    \hline\noalign{\smallskip}
    19 & CRAFT-MFA, 2018, \cite{chen2018person} & Market-1501 & 77.0/50.3 & No \\
    &      & CUHK-01 & 74.5/--  &    \\
    &      & CUHK-03 & 84.3/72.41  &    \\
    &      & VIPeR & 50.3/--  &    \\
    \hline\noalign{\smallskip}
    20 & DECAMEL, 2020, \cite{yu2020unsupervised} & Market-1501 & 60.24/32.44 & Yes \\
    &      & CUHK-01 & 65.81/--  &    \\
    &      & CUHK-03 & 38.27/--  &    \\
    &      & MSMT-17 & 30.34/11.13  &    \\
    \hline\noalign{\smallskip}
    21 & CSPL, 2018, \cite{dai2018cross} & CUHK-01 & 72.02/-- & No \\
    &      & CUHK-03 (Labeled) & 70.2/--  &    \\
    &      & CUHK-03 (Detected) & 66.8/--  &    \\
    &      & VIPeR & 51.3/--  &    \\
    &      & PRID-2011 & 69.20/--  &    \\
    \hline\noalign{\smallskip}
    22 & DAM, 2019, \cite{meng2019deep} & MARS & 74.65/-- & No \\
    &      & iLIDS & 77.3/--  &    \\
    &      & PRID-2011 & 87.0/--  &    \\
    \hline\noalign{\smallskip}
    23 & GMDA-RC, 2018, \cite{zhao2018maximal} & CUHK-01 & 75.69/-- & No \\
    &      & VIPeR & 67.09/--  &    \\
    \hline\noalign{\smallskip}
    24 & VS-SSL, 2020, \cite{jia2020view} & Market-1501 & 74.8/51.2 & No \\
    &      & CUHK-01 & 73.0/--  &    \\
    &      & VIPeR & 44.8/--  &    \\
    \hline\noalign{\smallskip}
    25 & CVDCA, 2016, \cite{chen2016asymmetric} & CUHK-01 & 47.8/-- & No \\
    &      & VIPeR & 47.78/--  &    \\
    &      & PRID-2011 & 57.6/--  &    \\
    \hline\noalign{\smallskip}
    26 & VCFL, 2021, \cite{zhang2021adversarial} & Market-1501 & 91.85/76.97 & No \\
    &      & DukeMTMC-ReID & 82.68/65.68  &    \\
    &      & CUHK-03 & 61.29/54.26  &    \\
    \hline\noalign{\smallskip}
    27 & MPML, 2019, \cite{hu2019adaptive} & Market-1501 & 55.61/27.58 & No \\
    &      & CUHK-01 & 64.98/--  &    \\
    &      & VIPeR & 44.72/--  &    \\
    \hline\noalign{\smallskip}
    28 & AANet, 2021, \cite{lian2021attention} & Market-1501 & 96.1/87.5 & No \\
    &      & DukeMTMC-ReID & 90.2/79.5  &    \\
    &      & CUHK-03 (Labeled) & 82.7/76.7  &    \\
    &      & CUHK-03 (Detected) & 77.2/70.5  &    \\
    \hline\noalign{\smallskip}
    29 & AVA-ReID, 2020, \cite{wu2020cross} & Market-1501 & 88.6/73.1 & No \\
    &      & CUHK-03 & 86.7/83.8  &    \\
    \hline\noalign{\smallskip}
    30 & CCM, 2021, \cite{wang2020exploiting} & DukeMTMC-VideoReID & 66.0/42.3 & No \\
    \hline\noalign{\smallskip}
    31 & RCN-ReID, 2019, \cite{zhang2018learning} & MARS & 48.0/-- & No \\
    &      & iLIDS & 65.0/--  &    \\
    \hline\noalign{\smallskip}
    32 & MLCPL, 2018, \cite{an2018multi} & CUHK-01 & 34.05/-- & No \\
    \end{longtable}
    }
\subsection{Low Resolution}
    \label{subsec:lowresolution}
    Mostly surveillance cameras are not capable to capture the high resolution images because of low resolution of cameras and vast distance between person and camera as shown Fig. \ref{fig:challenges}. Low resolution probe images and high resolution gallery images makes the re-id more challenging.
    Due to variations in pose, illumination and resolution appearance of same person may look very different in non-overlapping cameras. Technique of image restoration unable to produce efficient results in real time scenarios leveraging the domain gap in  self-supervised manner \emph{i.e.} without using extra cost of labelling. Therefore, it is important to address this problem and existing approaches address it by using appearance based methods. 
     
    These methods extract feature representation that contain high inter-class disparity between different subjects and low intra-class disparity for same subject. The intra-class disparity is often extensive than the inter-class disparity due to the significant appearance change across different cameras. Therefore, accurate matching become difficult.
    
    Additionally, illumination variance is one of the important challenge to address. As the light conditions play the most important role to match the query person from a large set of images captured by the non-overlapping cameras. Sometimes the color of body parts of a particular person perceives so different due to complex illumination variations as shown in Fig. \ref{fig:challenges}.
     
    Changes in visual appearance caused due to light variation is another challenge in person re-id. As variant light conditions also causes the change in pixel values. Lighting conditions significantly affects the performance of low level features such as texture and color. Color information of pedestrian like clothes become unidentifiable when lighting conditions changes (with or without lighting). Existing approaches are based on learning the depth information, color based features and joint representation learning. To extract the depth information using depth cameras such as Kinect is not a difficult task. Kinect obtain the depth value (distance to the camera) of each pixel by infrared, regardless of subject color and illumination.  But there are several limitations, depth images captured by depth cameras changes when the view point of person changes across camera views. To develop a model which efficiently learn the discriminative low level features is still challenging task in person re-identification. Fig. \ref{fig:progressOnillumination} shows the state of the art results on respective datasets achieved so far on illumination challenge. 
    
\begin{figure}[h]
        \includegraphics[width=\textwidth]{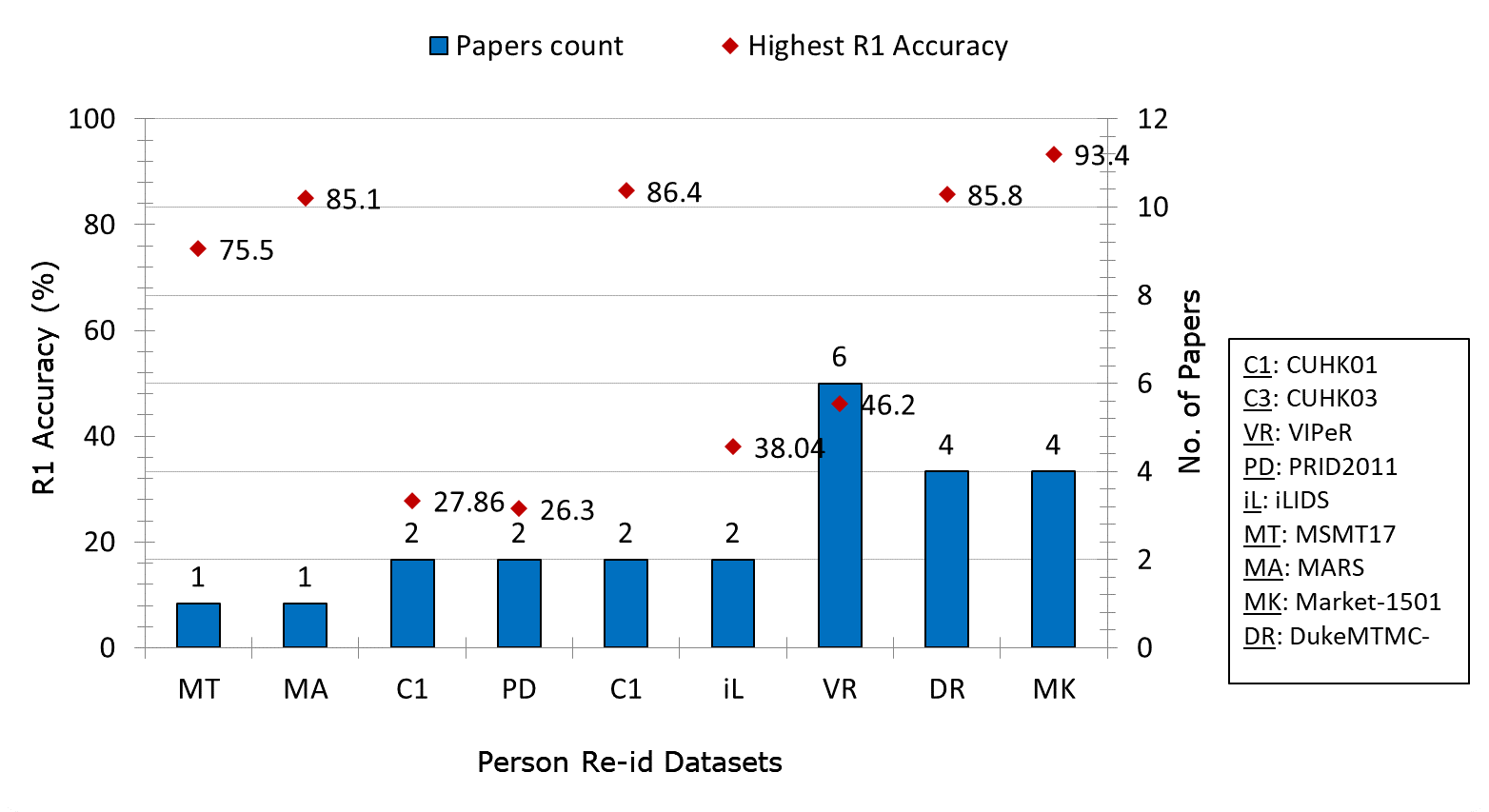}
        \caption{Progress on the challenges of low resolution and Illumination variance for person re-id benchmarks}
        \label{fig:progressOnlowresolution}
    \end{figure}

    Fig. \ref{fig:progressOnlowresolution} shows the progress of available papers on low resolution and illumination variance between 2015 to 2021.
    \subsubsection{CNN-based Approaches:}
    \label{subsubsec:LowResolutionCNNBasedApproaches}
    A supervised framework to resolve the challenge of resolution was proposed in \cite{wang2018resource}. They have combined embedding of multiple layers into single layer. In order to achieve efficient results lower layers with higher resolution are combined with higher layers having semantic information.
    Xiang li \emph{et al.} \cite{li2015multi} presented an attempt to resolve the low resolution that was beyond the scope of re-scaling and interpolation mechanism. They have presented the image features that are at same scale in a latent space and then performed distance metric modeling at each scale, this resulted in formation of shared space among low resolution and normal images of same person to obtain the effective results.
     
    In \cite{jing2015super} issue of no availability of super resolution images was challenged. In the presented procedure pair of high resolution and low resolution dictionaries and mappings functions are learned during training, due to this learned dictionary and mapping function low resolution images would be converted to high resolution discriminant features.
    Ke Han \emph{et al.} in \cite{han2020prediction} challenged the issue of low resolution. Presented model predicts and recovers the content aware details. Using self supervised approach the model also assigns  soft labels that are automatically generated. Important  thing is unique labels are assigned according to the optimal scale factor, it is to recover the issue of lacking ground truths. The probability of generated labels indicates the optimality of assigned scale level. Hence increasing the level of confidence against optimal scale for optimal resolution with context aware prediction.
     
    In \cite{han2020adaptive} an end-to-end adaptive feature fusion framework was proposed that has proved effective in handling resolution at different recovered body regions and at multiple scales. An adaptive feature integration module balances the relative importance of super resolved image content. In this manner adaptive weights were assigned to input features with super resolution.
    An extended version of \cite{jing2015super} was presented in \cite{jing2017super} to improve the results on resolution and they have showed analysis results on two more datasets with improved approach.Now  the structure was supporting multi-view. They have used to learn different mappings to convert low resolution images into discriminative high resolution features.
    In \cite{feng2021resolution} resolution-aware framework was proposed. They have used the knowledge transfer technique to minimize the variations in resolution of images. Teacher knowledge was exploited and transferred to the low resolution student network to narrow down the resolution differences.
    
    Mainly, challenge of illumination, camera-viewpoint and resolution were focused for adaptive cross-domain person re-id.
    In \cite{karianakis2018reinforced} illumination/lighting change was challenged. Reinforced temporal attention based end-to-end framework was proposed that was implemented at each frame level features to extract the temporal information accurately for depth based person re-id. LSTM was used after obtained features from CNN. LSTM models/learns the short term temporal dynamics/changes. Proposed approach is also useful for person with unseen clothes.
     
    For metric learning a multi-modality approach was presented in \cite{liu2018m3l}, it learns the changes in illumination via shift-invariant property. Proposed model also learn the sub-metric against each modality to reduce the role of bias in a global sense. Approach was validated on multiple datasets.
    
    In \cite{varior2016learning} illumination impact was studied and proposed a framework to resolve. It transform the pixels to invariant color space. Proposed framework learns the patterns and structures inherent in image pixels. It jointly learns the encoding and transformations among pixels pairs belonging to image pairs. This auto-encoder based approach transform the 3D-RGB pixel values to higher dimensional space and encode them using dictionary. Then mapping of pixels to invariant space is performed. These encoded pixel values are pooled over a particular regions and then integrated (or concatenated) to formulate the final representation. For higher level complementary feature learning, this framework can also be extended to multi-layers.
     
    \subsubsection{\textbf{GAN-based Approaches:}}
    \label{subsubsec:LowResolutionGANBasedApproaches}
    Cross resolution problem was considered in \cite{cheng2020inter} . In the proposed methodology existing GAN architecture was improved using high resolution images in an end-to-end fashion. Problem of low resolution was resolved by forming the association between super resolution images and re-id task. The formulation was due to parameterized sharing while training in end-to-end manner. 
     
    The feature mismatch problem caused due to the variations in resolution and illumination  might lead to poor representation learning and was resolved in \cite{huang2020real}.  In this self-supervised strategy identity related information was extracted to resolve the challenge of degradation occurs in real time. No extra computation was used instead model works on low resolution images. The approach was based on GAN, in which images were generated by switching the content of real world images with generated images. In this way domain gap between gallery images and real world images was reduced. Model effectively preserves the features related to identity information and remove the features that were related to degradation to achieve the effective results.

    {\small
    \begin{longtable} {m{1.5em}  m{9em} m{10em} m{5em}  m{4.5em}}
    \caption{Results obtained on low resolution and illumination variance challenge against each dataset. Results in bold are the highest.\label{table:LowResolutionResults}}\\
     Sr.No & Paper & Dataset & R1/mAP & Code availability\\
    \hline\noalign{\smallskip}\hline\noalign{\smallskip}
    1 & JUDEA, 2015, \cite{li2015multi}  & VIPeR & 25.87/-- & No\\
    \hline\noalign{\smallskip}
    2 & PRI, 2020, \cite{han2020prediction} & Market-1501 & 86.9/-- & No\\
      &      &  DukeMTMC-ReId & 82.1/-- &    \\
    \hline\noalign{\smallskip}
    3 & INTACT, 2020, \cite{cheng2020inter} & Market-1501 & 88.1/-- & No \\
      &      &  DukeMTMC-ReId & 81.2/-- &    \\
      &      &  CUHK-03 & 86.4/-- &  \\
      &      &  VIPeR & 46.2/-- &        \\
    \hline\noalign{\smallskip}
     4 & DaRe, 2018, \cite{wang2018resource} & Market-1501 & 90.9/86.7 & Yes \\
       &      & DukeMTMC-ReId & 84.4/80.0   &       \\
       &      & CUHK-03 (Labeled) & 73.8/74.7  &    \\
       &      & CUHK-03 (Detected) & 70.6/71.6  &    \\
       &      & MARS & 85.1/81.9  &    \\
     \hline\noalign{\smallskip}
     5 & SLD2L, 2015, \cite{jing2015super} & CUHK-01 & 24.48/-- & No\\
       &      & VIPeR & 16.86/--  &        \\
       &      & iLIDS & 33.33/--  &        \\
       &      & PRID-2011 & 22.6/--  &        \\
     \hline\noalign{\smallskip}
     6 & MCSLD2L, 2017, \cite{jing2017super} & CUHK-01 & 27.86/-- & No\\
        &      & VIPeR & 20.79/--  &        \\
        &      & iLIDS  & 38.04/-- &        \\
        &      & PRID-2011  & 26.30/-- &        \\
     \hline\noalign{\smallskip}
     7 & RKD, 2021, \cite{feng2021resolution} & Market-1501 & 93.4/83.7 & No\\
        &      & DukeMTMC-ReId & 85.8/73.0  &        \\
     \hline\noalign{\smallskip}
     8 & DI-REID, 2020, \cite{huang2020real}  & MSMT-17 & 75.5/-- & No\\
    \hline\noalign{\smallskip}
    9 & JLCF, 2016, \cite{varior2016learning} & VIPeR & 26.27/-- & No\\
    \hline\noalign{\smallskip}
    10 & M3L, 2017, \cite{liu2018m3l} &  VIPeR & 30.22/-- & No \\
    \hline\noalign{\smallskip}
    \end{longtable}
    }
\subsection{Cross-Domain/ Generalization}
    \label{subsec:generalization}
\begin{figure} [h]
        \includegraphics[width=\textwidth]{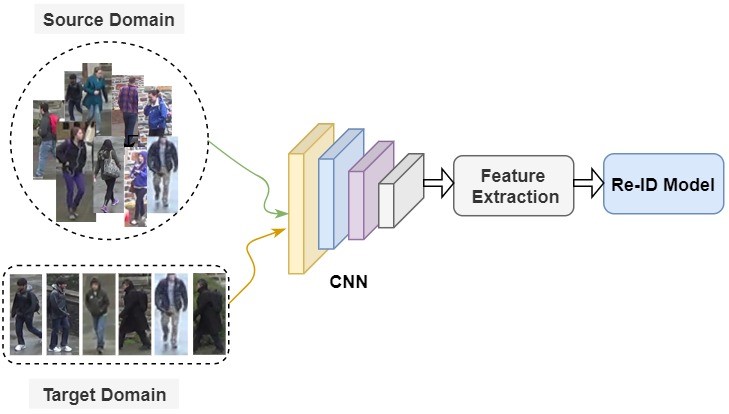}
        \caption{An approach to achieve generalization.}
        \label{fig:generalization}
    \end{figure}

\begin{figure} [h]
        \includegraphics[width=\textwidth]{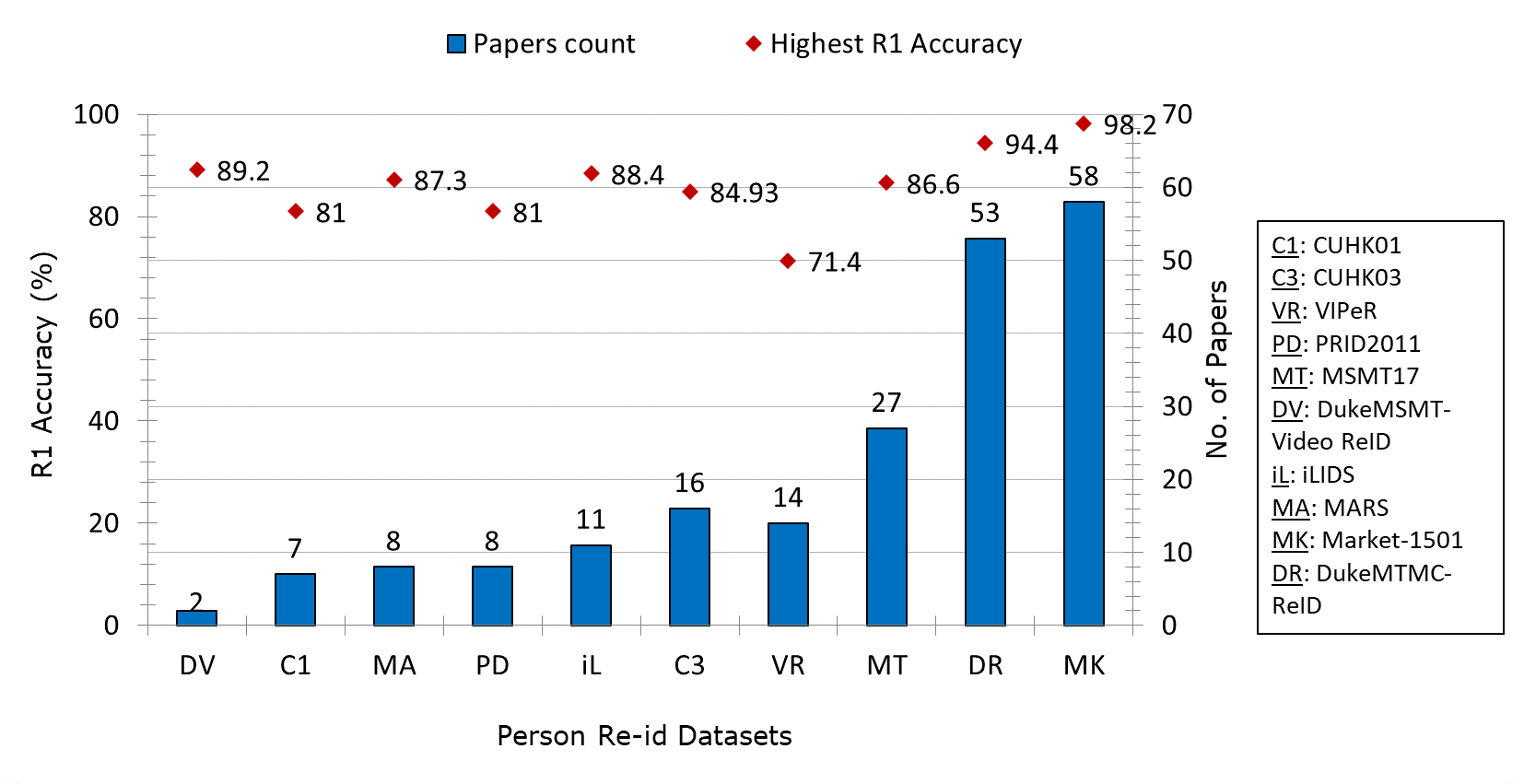}
        \caption{Progress on the challenge of cross-domain generalization for person re-id benchmarks}
        \label{fig:progressOnGeneralization}
    \end{figure}

    Person re-id models produce improved results when trained and tested on same dataset but perform poor when tested on different dataset due to different scenarios \emph{e.g.} changes in viewpoint, place, background, resolution and different visual appearance. For this, recently clustering, domain adaptation and image to image translation based approaches reported state of the art results. One such approach is shown in Fig. \ref{fig:generalization}.
    Aims of image to image translation is to develop a mapping function between two domains and it required paired training data which is difficult to manage. For domain adaptation, labelling a dataset is expensive and time consuming task. Due to these factors improving generalization is still challenging in person re-identification.
    Progress of Papers that has addressed the generalization challenge in top conferences and journals is shown in Fig. \ref{fig:progressOnGeneralization}

    \subsubsection{CNN-based Approaches:}
    \label{subsubsec:CrossDomainCNNBasedApproaches}
    Changes in background and poses results in challenge of intra-class variation and that is addressed in \cite{zheng2019joint}. In the proposed discriminative module primary features are learned along with generative module to generalize the data well. In \cite{kalayeh2018human} semantic segmentation with simple base line was used to address the challenge of local feature extraction from various human body parts that is an alternative to bounding box approach. In this framework, in order to integrate the body part classification, the identity attribute features was presented. ResNet-50 \cite{he2016deep} along with pre-trained ImageNet weights was used to resolve the partly attention-based challenge.
    Ruijie Quan \emph{et al.} \cite{quan2019auto} proposed an architecture that automatically searches for appropriate CNN that is suitable for efficient person re-id eliminating human effort. Body part information was used to capture the structure information of human body parts. Structure information was then embedded into flexible part-aware module to achieve state of the art results.
     
    Poor generalization of the model may lead to domain gaps that are needed to be covered. One of such approach to tackle this challenge was given in \cite{jin2020style} using unsupervised approach. Specifically a style normalization and restitution module was introduced that filters out the variations in style and color using instance normalization. Features related to identity are then refined and added to the network to achieve valuable discrimination results on unseen data as well.  For better feature disentanglement dual loss was also used in the proposed strategy.
    In \cite{huang2018eanet} Houjing Huang \emph{et al.} designed a cross-domain adaptive model without identity labels of target-domain. Part Aligned Pooling (PAP) was introduced to enhance alignment and hence results in model generalization. In the approach it is verified that part-alignment plays a vital role in achieving cross-domain generalization of re-id model. They have argued that training unlabelled target domain with part segmentation and training re-id on source domain, is an effective way to achieve cross-domain generalization or domain adaption.
     
    Weihua Chen \emph{et al.} in \cite{chen2017beyond} presented novel quadruplet ranking loss that was based on triplet loss; it was to increase the generality of the procedure for person re-id. It effectively achieves the capability to handle the intra and inter class variations. A quadruplet deep network was also presented that makes 4 hard samples of dataset and then proposed loss was applied to obtain the effective cross-domain results.
    To make the model domain invariant an efficient procedure was presented in \cite{xiao2016learning}. At first stage, for strong baseline they have merged multiple person re-id datasets and trained a CNN using single softmax loss. Standard dropout layer was replaced with domain guided dropout layer for training few more epochs, resulted in effective learning of weights that are more effective for each domain.
     
    In \cite{zhai2020multiple} domain adaptive person re-id was proposed using unsupervised target domain. Multiple expert brainstorming networks were learned with multiple architectures for optimal re-id. In proposed network models were first pre-trained on multiple expert models. To adjust the heterogeneity of multiple models a regularization scheme was added that modulates the expert models according to their feature distribution in the target domain. In this way significant performance was gained due to the increase in discrimination capability of re-id model.
    Another framework to achieve generlized results in unseen scenarios as well was proposed in \cite{zhuang2020rethinking}. In order to reduce the domain gap in the presented mechanism all camera images were assigned the common subspace. This resulted in achieving the generalized results even on unseen images.
     
    In another framework presented in \cite{luo2020generalizing} interpolation mechanism was used to obtain cross-domain generality. Vanilla neighbourhood approach was improved by restricting the camera-aware manner.
    In \cite{zou2020joint} a joint learning framework was presented for generalized person re-id. Representation space was purified in  a way that it can only learn id related feature space. Disentangling module encodes cross-domain images into appearance space and structure space that was shared for reduced domain gap.
    Recently a new dataset Person30K is proposed in \cite{bai2021person30k} to support generalization. Proposed DMG-Net strengthens the model to attain generalization on unseen data without the need to fine tune further.  Specifically, feature based centers were computed that represent identities and then matching was done between support centers and query samples.
     
    In \cite{choi2021meta} Meta batch instance normalization (MetaBIN) technique was used in an unsupervised manner to achieve the generalization on unseen person re-identification domains. The MetaBIN approach prevents the model from over-fitting. It does so by learning the scenarios before hand in a phase called meta-learning. Moreover meta-train loss accompanied by a cyclic inner-updating manner helped in boosting the generalization capability. It is noted that no additional augmentation technique used that makes the model more complicated.
    Another method was proposed in \cite{dai2021generalizable} named as relevance-aware mixture of experts (RaMoE). In the method voting based technique was used to dynamically integrate the diverse characteristics of source domains that eventually helped in increasing the generalization capability of the model.
     
    Anguo Zhang \emph{et al.} in \cite{zhang2021coarse} an issue of intra-class variation was challenged. Second order information bottleneck was integrated into the network. It optimizes the network at inference time resulted in reduction in computation overhead. Proposed framework has reduced the impact of intra-class variations and superior performance was achieved.
    To generalize the results on unseen domains experiments on synthetic data were presented in \cite{zhang2021unrealperson}. Usage of unreal data have significantly reduced the cost of both training and testing and have also improved the results via using the transfer knowledge technique on real datasets. The approach has also provided the basis for unsupervised domain adaptations. The pre-trained model obtained via proposed technique can be easily plugged into existing methods to achieve better results.
     
    An optimization strategy was presented in \cite{zheng2021online} to achieve the generalization. Specifically label banks were formed in hierarchical manner, mini-batches were updated using iterative approach. Hierarchical clusters form and split on the basis of computed or propagated label information. Effective labeling resulted in achieving the generalization of the proposed model.
    \cite{khatun2020joint} generalization was improved that was based on their previous work \cite{khatun2018deep}. An additional term was added in the quartet loss that ensures the distance between matching pairs should be less than specified threshold, that resulted in less intra-class distance and more than inter-class distance with different probe images. It also ensured the closeness of intra-class features to each other. Moreover, model has also optimised both identification and classification tasks.
     
    In order to reduce the difference in domains due to background inferences and to attain generalization, an unsupervised approach was presented in \cite{huang2021unsupervised}. They have suppressed the background information to have the focus on learning the person's information present in the foreground. They have used human body information and associated ID related information present in the environment. Model keeps on updating using the virtual labels of target domain.
    In an effort to learn the efficient feature embedding a joint learning framework was presented in \cite{feng2021complementary}. In order to predict a non-ambiguous label information subgroups were formed by merging neighbours. For higher subgroup influence similarity-aggregating loss was introduced that has pulled the similar embedding closer to each other.
     
    Using multi-tasking reinforcement learning both spatial and temporal embedding were learnt by a framework presented in \cite{song2019context}. The framework was comprised of two branches. First branch determines the optimal spatial region along with temporal range of frames to obtain the contextual information. While second branch has focused on learning the identity information based on the information obtained from the first branch. Collaborative interaction among pedestrians and the context were exploited to achieve outstanding performance.
    Another approach to mitigate the challenges of unsupervised domain adaptation was proposed in \cite{dai2021dual}. In the proposed dual-refinement approach pseudo labels were refined at off-line clustering  while feature refinement was performed at on-line training phase. This dual-refinement has minimized the influence of noisy labels and hence superior results were obtained.
    A model to learn the domain invariant features was designed in \cite{lin2020multi}. Adversarial auto-encoder was the model base. And maximum mean discrepancy measure was used to align the learned distribution across variant domains.
     
    A 3D guided network was developed in \cite{zhang20213d}. In order to support the formulation on unknown domains as well the proposed model have the ability to transfer knowledge from domain training data. They have performed image-to-image translation to form the possible synthetic poses and viewpoints of a person while preserving the identity information. In a way higher accuracy results were achieved even on unknown domains.
    Aiming to enhance the generalization with low computational complexity, another framework was presented in \cite{zhao2018person} comprised of two related modules \emph{i.e.} patch-based metric learning and local salience learning. CNN was used to extract the features of person patches then just two positive patch-pairs were selected to learn the patch based metric matrices. k-means clustering was then used as a salience learning algorithm to learn the patch weights. Finally, on the basis of these learned weights patch-wise similarity score was computed.
     
    An approach to address cross-dataset person re-id was developed in \cite{li2020structure}. In the proposed dictionary learning approach visual-attribute embedding was learned and then transferred from source to target domain. Finally, pseudo-labels were fine-tuned by choosing few samples from the target domain to produce outperforming results.
    The challenge of part-based learning was addressed in a model named PatchNet \cite{yang2019patch}. In order to learn the patch features discriminative model measures similarity between images in unsupervised manner.
    In \cite{yao2019deep} a deep representation learning procedure named as Part Loss Network (PL-Net) was proposed. Part loss network minimizes the empirical classification risk (ECR) on training images and representation learning risk (RLR) on unseen images. RLR was evaluated via part loss. It was evaluated for each body part separately. This part loss computation helped in gaining the discriminative power on unseen images as compared to traditional global classification loss.
     
    A semi-supervised approach based on deep attribute learning framework was presented in \cite{su2016deep} to achieve cross-domain generalization. At first stage fully supervised training is performed, resulted in learned attribute labels against target dataset. Second stage involved the fine tuning of learned labels on the basis of attribute triplet loss. At every iteration, attribute triplet loss ensures that similar attributes belong to same person. Seconds stage predicts attributes for target dataset, these are then combined with initial labels obtained at first stage and then fine tuned again. At final stage,  more discriminative deep attributes were obtained and that have performed impressively well. The approach was semi-supervised because it includes one dataset with labels and other without any attribute label.
     
    In \cite{shi2015transferring} context representation was learned and transferred to target re-id dataset via Bayesian adaptation. Proposed approach can handle both labelled and weakly labelled data in the training process. Learned representation was view invariant but person variant, that is an ideal case for efficient person re-id.
    To make re-id task more scalable a domain transfer unsupervised technique was presented in \cite{peng2016unsupervised}. In the approach they have used target data with no labelled matching pairs. Model transfers the view invariant representation from labelled datasets to an unlabelled target dataset. Dictionary learning was adopted assuming that multiple appearances of a person can be represented as linear combination of vectors.
     
    An unsupervised approach to achieve the effective outcome in real world scenarios was presented in\cite{jin2020global}. Domain gaps were managed using Gaussian distribution. Positive and negative samples were pushed apart using unsupervised setting. A momentum update mechanism was applied to clearly separate the negative and positive samples distribution. Hence significant improvement over baselines was achieved.
    In \cite{chen2020deep} potential of cross domain model was explored for real world application of person re-id systems. System was developed using unsupervised domain adaptation. Proposed model has adaptively learned the credible samples for training by avoiding noisy labels. An instance margin loss was introduced that has increased the margin of instance. Resulted in increased instance level discrimination of learned features.
     
    Djebril Mekhazni \emph{et al.} in \cite{mekhazni2020unsupervised} presented an unsupervised scheme for unseen domain adaptation for person re-id. Pair-wise distances were optimized by using gradient descent and comparatively smaller batches. They have introduced novel dissimilarity based discrepancy loss that  made the source and target distribution similar to make the results more efficient.
    Multi-source domain training to make the model generalize on multiple unseen domain was a way adopted in \cite{bai2021unsupervised}. For this purpose, there comes a problem of domain gaps. To resolve such issues two strategies were adopted. First was domain-specific batch normalization and second was the infusion of multi-domain information. These strategies had helped in reducing the domain distances via fusing the features of multiple domains. Proposed techniques have provided the comparable results to supervised approaches. Attained results were obtained without using any post-processing techniques.
    In order to address the limited scale issue of existing re-id datasets, a new dataset named as Large scale Unsupervised Person re-id (LUPerson) was introuduced in \cite{fu2021unsupervised}. They have investigated the impact of factors in learning \emph{i.e.} data augmentation and temperature usage in a contrastive learning framework.
     
    Degraded label assignment caused due to distribution discrepancy among cameras was a key issue and addressed in \cite{xuan2021intra}. To obtain the possible accuracy on pseudo-label's generation the proposed framework was divided into two modules \emph{i.e.} inter-camera and intra-camera computation. In inter-camera computation a new feature vector was computed for different cameras. This new feature vector results in obtaining the reliable pseudo-labels and hence removes the issue of distribution discrepancy among different cameras. While in intra-camera computation similarity measure was computed by comparing features obtained from CNN. Then the model was trained in two stages \emph{i.e.} inter-camera and intra-camera pseudo-labels to obtain the outperforming results.
    In \cite{yang2021joint} problem of discriminative learning with unlabeled data for unsupervised person re-id is targeted. A Dynamic and Symmetric Cross-Entropy (DSCE) loss was introduced to cope the challenge of negative effect caused due to noisy labels. And for handling the camera shift issue a meta-learning (MetaCam) technique that splits the training data into meta-train and meta-test was used. MetaCam resulted in achieving the interactive gradient impact for each meta set that would enforce the model to learn the camera-invariant features for better results. 
     
    To tackle the challenge of pseudo label noise an approach was presented in\cite{zhang2021refining}. They have refined the pseudo labels based on clustering consensus. Pseudo labels were refined dynamically with temporally generated ensembles pseudo labels. This refinery approach can also be integrated into existing clustering based methods to obtain the simple yet effective results.
    A method with only few labeled information was presented in \cite{zhao2021learning}. They have presented a multi-domain generalization framework that can perform well on unseen domains with no need to train a new model. Specifically, a meta learning strategy was adopted to perform the like train-test process resulted in learning more generalized model. A memory based generalization loss and a meta batch normalization layer was also added to diversify the advantages of meta-learning.
    In \cite{zheng2021group} a Group-aware label transfer algorithm was proposed. It promotes the pseudo-labels via online interaction. It not only uses the pseudo-labels but at the same time it also refines them based on an online clustering algorithm. The effectiveness of the approach was tested on large-scale re-id datasets and hence it has helped to reduce the gap among supervised and unsupervised approaches.
     
    In order to bridge the domain gap collection of unsupervised schemes were adopted in \cite{isobe2021towards} that have resulted in learning the unique representation of features. First a clustering algorithm optimizes the features in iterative manner to ensure that learnt features are noise free. Second was the progressive domain adaptation and third was the Fourier augmentation in which extra constraints were deployed to increase the chances of class separability.
    In \cite{zhao2020joint} patch-wise features were learnt from unlabelled patches of person's image. A novel loss function was then used that guides the model to mine the discriminative information. Same features were pulled closer while features belonging to different instances were pulled away depending on the computation of loss function.
     
    Yan Bai \emph{et al.} in \cite{bai2021hierarchical} a reliable approach to generate pseudo labels was presented for unsupervised domain adaptation. In the proposed hierarchical scheme graph convolutions were used to learn the complicated structure of each cluster. Connection among samples were estimated in a hierarchical way and then refined progressively.
    In video person re-id \cite{liu2020iterative} a joint global video and local frame information was considered to obtain the diverse information for better estimation of pseudo labels. Moreover novel loss term induce the model to not focus on undesirable factors of identity. A dynamic strategy was also adopted to choose the pseudo label with higher confidence score that keeps on updating during training process until the higher confidence score met.
     
    An end-to-end self-supervised learning algorithm was proposed in \cite{jiang2020self} for unsupervised person re-id. Domain discrepancy was minimized using agent learning mechanism. Due to learnt discriminative representation model has performed well on unseen domains as well as compared to existing unsupervised approaches.
    Hang Zhang \emph{et al.} has presented an unsupervised view-invariant approach to tackle the person re-id at multi-scale levels in \cite{zhang2021self}. Proposed framework was self-trained and can be implemented on unseen domains as well. Local and global representations of pedestrian images were learnt.And then used as pseudo-labels that were improved progressively using iterative approach.
    For unsupervised domain adaptation a multi-loss optimization learning model was proposed in \cite{sun2021unsupervised}. They have estimated the pseudo labels via clustering mechanism applied in a supervised way. For similarity and adversarial learning two losses were introduced that had helped in model optimization. Together these loss terms has also benefited in exploring the intra-domain relation to evaluate and estimate the improved final outcome.
     
    In \cite{li2020unsupervised} discriminative representation was obtained from tracklet data in an end-to-end formulation. Proposed framework learned the discrimination and association within-camera and cross-camera respectively. Superior performance on eight benchmark datasets proves the effectiveness of the proposed framework.
    In \cite{li2019attribute} a self-supervised learning algorithm was proposed to achieve generalization. In order to bridge the gap among source and target attribute-identity embedding were used as a base to optimize the model. A prediction-training cycle was also implemented with a purpose to fine-tune the model variables so that the model become more adaptive to target domain.
    In \cite{zheng2016mars} video tracklets were used to achieve the generalized results at large scale, hence proposed new ideal dataset named MARS. They employed motion features and CNNs to learn discriminative embedding. According to their results motion features are less effective in real scenarios because of complex challenges like occlusion, pose and background clutter. However CNN based features remarkably performed well as there we have large training data hence a good generalization ability was achieved.
     
    Challenge of cross-view and intra-bag alignment was addressed using weakly supervised approach \cite{meng2019weakly}. To reduce the labelling cost presence of humans was labelled only however the information of what and where was not considered.
    Density based clustering is a technique adopted by \cite{zhai2020ad} to form a model that is domain adaptive. Proposed model was made discriminative in the target domain using GAN’s min-max strategy. Sample clusters were first predicted from the target domain, and then sample features were extracted using re-id model. The model was already pre-trained on the source domain. Cluster formation was then improved via iterative process. Feature encoder increases the inter-class distance and decreases the intra-class distance. Image generator and feature encoder competes in adversarial min-max manner and hence helped to optimize the models effectively.
     
    Liao1 \emph{et al.} in \cite{liao2019interpretable} also considered the generalization as a challenge and proposed a framework to resolve it.
    In order to bridge the domain-gap ‘divide-and-conquer’ strategy was adapted for factor-wise style transfer \cite{liu2019adaptive} based on CycleGAN \cite{zhou2017point}.
    An unsupervised domain-to-domain translation method was presented in \cite{tang2020cgan}. The method keeps the pedestrian identity information and have used maximum mean discrepancy as a base to pull the alike distribution closer.They have used CycleGAN to transfer the label information to unlabeled domain.
    Weijian Deng \emph{et al.} \cite{deng2018image} has improved the baseline of “learning via translation”, in their proposed work they have tried to generalize the domain by preserving the similarity of image and dissimilarity of domain in unsupervised manner. Their proposed framework was based on Saimese Structure and Cycle-GAN.
     
    In \cite{pu2021lifelong} generalization was targeted using lifelong learning scenario. In the proposed framework learn-able knowledge graph was maintained that updates the previously learned knowledge in an adaptive manner.In order to improve the generalization learned knowledge was transferred on unseen domains. Promising improvement were reported over existing SOTA results.
    A self-training scheme  for optimized unsupervised domain adaptation was designed in \cite{song2020unsupervised}. An encoder was trained on the basis of guessed labels for unlabeled target data to achieve the effective results.
    New open environment larger dataset was introduced in \cite{zheng2015scalable} named "Market" to achieve generalized results due to its large scale. They have also presented an unsupervised Bag of Words descriptor that takes re-id task as image search and this is beyond the scope of this paper and hence not further discussed.
    
    \subsubsection{Attention-based Approaches:}
    \label{subsubsec:CrossDomainAttentionBasedApproaches}
    A self-critical attention based learning mechanism was proposed in \cite{chen2019self}. In the proposed design unified modules self-critic and self-correctness are introduced to guide the attention agent learn the correct attention maps based on information provided by critic module. 
    In \cite{chen2019mixed} high-order attention module was proposed to resolve the challenge of part-based modelling of pedestrians in person re-id task to generate more discriminative and powerful attention proposals. High order relationship among human parts was obtained using polynomial predictor of high-order. In this way discriminative attention maps are obtained with subtle differences. Proposed model works well on unseen person images as well due to learning at multiple diverse levels so that attention of all sides could be preserved.
     
    In \cite{tay2019aanet} classification based attribute aware re-id approach was proposed. In the proposed model attention mechanism was used to identify the specified body parts in a unified learning framework. Architecture integrates the identity information with attribute features and body parts. In this manner discriminative feature space was learned and model becomes more generic.
    An end-to-end supervised approach \cite{si2018dual} to make the model context free was based on attention mechanism. In the proposed Siamese network intra sequence and inter-sequence attention mechanism was used for feature refinement and alignment accordingly.
    A novel cross-correlated attention module was presented in \cite{zhou2020cross} that effectively learns the discriminative representation that was based on inherent spatial relation of different regions of a person image.
     
    In \cite{li2020scalable} large-scale re-id scenarios are targeted. A novel harmonious attention network was deployed that jointly learns the attention based pixel representation of soft and hard regions. Hence resulted in having more discriminative features that has helped in more efficient re-id searching and matching.
    An attentional aggregation formulation was designed in \cite{fu2020improving} to handle the changing representation of an identity in a query image. Proposed scheme flexibly incorporates the similarity metrics along with multiple representations.
    Videos captured from different cameras might end up a video with different camera views, and this view variation is challenge handled in \cite{xu2017jointly}. Their end-to-end framework accounts the inter-dependencies among video sequences. They used Recurrent Convolutional Network to extract features and then learns similarity among them. These similarity scores are then used to form the attention network in spatial as well as temporal dimensions. Obtained attention vectors are forwarded for pooling. At final stage, Siamese structure is deployed at attention vectors to make the solution more generalised.
    
    {\small
    \begin{longtable} {m{1.5em}  m{9em} m{10em} m{5em}  m{4.5em}}
    \caption{Results obtained on Cross-domain/Generalization challenge against each dataset. Results in bold are the highest.\label{table:CrossDomainResults}}\\
    
    \hline\noalign{\smallskip}
     Sr.No & Paper & Dataset & R1/mAP & Code availability\\
    \hline\noalign{\smallskip}\hline\noalign{\smallskip}
    1 & CCL-PDA-FA, 2021, \cite{isobe2021towards}  & Market-1501 & 94.2/83.4 & No\\
      &      &  DukeMTMC-ReId & 83.5/70.8 &    \\
      &      &  MSMT-17 & 66.6/36.3  &        \\
    \hline\noalign{\smallskip}
    2 & OPLG, 2021, \cite{zheng2021online} & Market-1501 & 91.5/80.0 & No\\
      &      &  DukeMTMC-ReId & 82.2/70.1 &    \\
      &      &  MSMT-17 & 56.1/29.3  &        \\
    \hline\noalign{\smallskip}
    3 & SCAL, 2019, \cite{chen2019self} & Market-1501 & 95.8/89.3 & No \\
      &      &  DukeMTMC-ReId & 88.9/79.1 &    \\
      &      &  CUHK-03 (Labeled)  & 74.8/72.3 &  \\
      &      &  CUHK-03 (Detected)  & 71.1/68.6 &  \\
    \hline\noalign{\smallskip}
     4 & NAS, 2019, \cite{quan2019auto} & Market-1501 & 95.4/94.2 & No \\
       &      & CUHK-03 (labeled) & 77.9/73  &    \\
       &      & CUHK-03 (Detected) & 73.3/69.3  &    \\
       &      & MSMT-17 & 78.2/52.5  &    \\
     \hline\noalign{\smallskip}
     5 & MHN, 2019, \cite{chen2019mixed} & Market-1501 & 95.1/85.0 & Yes\\
     &      & DukeMTMC-ReID & 89.1/77.2  &        \\
     &      & CUHK-03 & 71.7/65.4  &        \\
     \hline\noalign{\smallskip}
     6 & ASTPN, 2017, \cite{xu2017jointly} & MARS & 44.0/-- & Yes\\
       &      & PRID-2011 & 30.0/--  &        \\
     \hline\noalign{\smallskip}
     7 & DPM, 2015, \cite{g2015scalable} & Market-1501 & 42.64/19.47 & No\\
        &      & CUHK-03 & 22.95/22.7  &        \\
        &      & VIPeR & 21.74/26.55  &        \\
     \hline\noalign{\smallskip}
     8 & MEB-Net, 2020, \cite{zhai2020multiple} & Market-1501 & 89.9/76.0 & Yes\\
        &       & DukeMTMC-ReID & 79.6/66.1 &   \\
     \hline\noalign{\smallskip}
     9 & GDS, 2020, \cite{jin2020global} & Market-1501  & 81.1/61.2 & No \\
       &       & DukeMTMC-ReID & 73.1/55.1 &   \\
     \hline\noalign{\smallskip}
     10 & DCML, 2020, \cite{chen2020deep}  & Market-1501  & 88.2/72.3 & No\\
     &      & DukeMTMC-ReId & 79.3/63.5  &        \\
     \hline\noalign{\smallskip}
     11 & NRMT, 2020, \cite{huang2021unsupervised} & Market-1501 & 87.8/71.7 & No \\
     &      & DukeMTMC-ReId & 77.8/62.2  &        \\
     &      & MSMT-17 & 45.2/20.6  &        \\
     \hline\noalign{\smallskip}
     12 & CBN, 2020, \cite{zhuang2020rethinking} & Market-1501 & 94.3/83.6 & Yes\\
     &      & DukeMTMC-ReId & 84.8/70.1  &        \\
     \hline\noalign{\smallskip}
     13 & CD-ReID, 2020, \cite{luo2020generalizing} & Market-1501 & 88.1/71.5 & Yes \\
     &      & DukeMTMC-ReId  & 79.5/65.2    &       \\
     \hline\noalign{\smallskip}
     14 & DG-Net++, 2020, \cite{zou2020joint} & Market-1501 & 83.1/64.6 & Yes \\
     &      & DukeMTMC-ReId & 78.9/63.8  &        \\
     &      & MSMT-17 & 48.8/22.1  &        \\
     \hline\noalign{\smallskip}
     15 & TLift, 2020, \cite{liao2019interpretable} & Market-1501 & 88.4/76.0 & Yes \\
     &      & DukeMTMC-ReId & 82.2/78.4  &        \\
     \hline\noalign{\smallskip}
     16 & D-MMD, 2020, \cite{mekhazni2020unsupervised}  & Market-1501 & 72.8/50.8 & Yes\\
     &      & DukeMTMC-ReId      & 68.8/51.6    &        \\
     &      & MSMT-17      & 34.4/15.3    &        \\
     \hline\noalign{\smallskip}
     17 & SSDAL, 2016, \cite{su2016deep} & Market-1501  & 49.0/25.8 & No\\
     &      & VIPeR      & 43.5/--    &        \\
     &      & PRID-2011      & 22.6/--    &        \\
     \hline\noalign{\smallskip}
     18 & MARS, 2016, \cite{zheng2016mars} & MARS & 68.3/49.3 & No\\
     &      & iLIDS   & 53.0/--    &       \\
     &      & PRID-2011  & 77.3/--  &        \\
     \hline\noalign{\smallskip}
     19 & DMG-Net, 2021, \cite{bai2021person30k} & Person-30K & 84.23/72.19 & No\\
     \hline\noalign{\smallskip}
     20 & RDSBN, 2021, \cite{bai2021unsupervised} & Market-1501 & 94.8/86.0 & No\\
     &      & DukeMTMC-ReId  & 82.1/68.9     &        \\
     &      & MSMT-17  & 64.7/34.9     &        \\
    \hline\noalign{\smallskip}
     21 & MetaBIN, 2021, \cite{choi2021meta} & Market-1501 & 69.2/35.9 & yes\\
     &      & DukeMTMC-ReId  & 55.2/33.1     &        \\
     &      & VIPeR  & 59.9/68.6     &        \\
     &      & iLIDS  & 81.3/87.0     &   \\
     &      & PRID-2011  & 81.0/72.4     & \\
     \hline\noalign{\smallskip}
     22 & RaMoE, 2021, \cite{dai2021generalizable} & Market-1501 & 82.0/56.5 & No \\
     &      & DukeMTMC-ReId  & 73.6/56.9     &        \\
     &      & CUHK-03  & 36.6/35.5     & \\
     &      & VIPeR  & 63.4/72.2     & \\
     &      & MSMT-17  & 34.1/13.5     &        \\
     &      & iLIDS  & 88.4/92.3     &        \\
     &      & PRID-2011  & 56.9/66.8     &        \\
     \hline\noalign{\smallskip}
     23 & LUPerson, 2021, \cite{fu2021unsupervised} & Market-1501 & 97.0/92.0 & No\\
     &      & DukeMTMC-ReId  & 91.9/84.1     &        \\
     &      & CUHK-03  & 81.9/79.6     & \\
     &      & MSMT-17  & 86.6/68.8     & \\
     \hline\noalign{\smallskip}
     24 & LReID, 2021, \cite{pu2021lifelong} & Market-1501 & 87.0/74.8 & Yes \\
     &      & DukeMTMC-ReId  & 80.1/68.3     &        \\
     &      & CUHK-03  & 56.6/50.8     & \\
     &      & MSMT-17  & 54.1/27.9     & \\
     \hline\noalign{\smallskip}
     25 & IICS, 2021, \cite{xuan2021intra} & Market-1501 & 89.5/72.9 & Yes \\
     &      & DukeMTMC-ReId  & 80.0/64.4     &        \\
     &      & MSMT-17  & 56.4/26.9     & \\
     \hline\noalign{\smallskip}
     26 & DSCE, 2021, \cite{yang2021joint} & Market-1501 & 83.9/61.7 & Yes \\
     &      & DukeMTMC-ReId  & 73.8/53.8     &        \\
     &      & MSMT-17  & 35.2/15.5     &   \\
     \hline\noalign{\smallskip}
     27 & ADC-2OIB, 2021, \cite{zhang2021coarse} & Market-1501 & 94.8/87.7 & No \\
     &      & DukeMTMC-ReId  & 87.4/74.9     &        \\
     &      & CUHK-03(Labeled)  & 80.6/79.3     & \\
     &      & CUHK-03(Detected)  & 81.3/84.1     & \\
     \hline\noalign{\smallskip}
     28 & RLCC, 2021, \cite{zhang2021refining} & Market-1501 & 90.8/77.7 & No \\
     &      & DukeMTMC-ReId  & 83.2/69.2     &        \\
     &      & MSMT-17  & 56.5/27.9     &   \\
     \hline\noalign{\smallskip}
     29  & UnrealPerson, 2021, \cite{zhang2021unrealperson} & Market-1501 & 93.0/80.2 & Yes \\
     &      & DukeMTMC-ReId  & 88.3/75.2     &        \\
     &      & MSMT-17  & 68.2/34.8     &   \\
     \hline\noalign{\smallskip}
     30  & M3L, 2021, \cite{zhao2021learning} & Market-1501 & 75.9/50.2 & Yes \\
     &      & DukeMTMC-ReId  & 69.2/51.1     &        \\
     &      & CUHK-03  & 33.1/32.1     & \\
     &      & MSMT-17  & 36.9/14.7     &   \\
     \hline\noalign{\smallskip}
     31  & GLT, 2021, \cite{zheng2021group} & Market-1501 & 92.2/79.5 & No \\
     &      & DukeMTMC-ReId  & 82.0/69.2     &        \\
     &      & MSMT-17  & 59.5/27.7     &   \\
     \hline\noalign{\smallskip}
     32  & SNR, 2020, \cite{jin2020style} & Market-1501 & 85.5/65.9 & No \\
     &      & DukeMTMC-ReId  & 78.2/61.6     &        \\
     \hline\noalign{\smallskip}
     33  & AD-Cluster, 2020, \cite{zhai2020ad} & Market-1501 & 86.7/68.3 & No \\
     &      & DukeMTMC-ReId  & 72.6/54.1     &        \\
     \hline\noalign{\smallskip}
     34  & ATNet, 2019, \cite{liu2019adaptive} & Market-1501 & 45.1/24.9 & No \\
     &      & DukeMTMC-ReId  & 55.7/25.6     &        \\
     \hline\noalign{\smallskip}
     35  & AANet, 2019, \cite{tay2019aanet} & Market-1501 & 95.1/92.38 & No \\
     &      & DukeMTMC-ReId  & 90.36/36.87     &        \\
     \hline\noalign{\smallskip}
     36  & PatchNet, 2019, \cite{yang2019patch} & Market-1501 & 68.5/40.1 & Yes \\
     &      & DukeMTMC-ReId  & 72.0/53.2     &        \\
     \hline\noalign{\smallskip}
     37  & EANet, 2019, \cite{huang2018eanet} & Market-1501 & 94.5/85.6 & Yes \\
     &      & DukeMTMC-ReId  & 87.5/74.6     &        \\
     &      & CUHK-03  & 72.5/66.8     &        \\
     \hline\noalign{\smallskip}
     38  & DG-Net, 2019, \cite{zheng2019joint} & Market-1501 & 94.8/86.0 & No \\
     &      & DukeMTMC-ReId  & 86.6/74.8     &        \\
     &      & MSMT-17  & 77.2/52.3     &        \\
     \hline\noalign{\smallskip}
     39  & CV-MIML, 2019, \cite{meng2019weakly} & DukeMTMC-VideoReId & 78.05/59.53 & No \\
     &      & MARS  & 66.88/55.16     &        \\
     &      & iLIDS  & 60.0/56.01     &        \\
     &      & PRID-2011  & 72.0/70.78     &        \\
     \hline\noalign{\smallskip}
     40  & SPGAN, 2018, \cite{deng2018image} & Market-1501 & 58.1/26.9 & No \\
     &      & DukeMTMC-ReId  & 46.9/26.4     &        \\
     \hline\noalign{\smallskip}
     41  & DuATM, 2018, \cite{si2018dual} & Market-1501 & 91.42/76.62 & No \\
     &      & DukeMTMC-ReId  & 81.82/64.58     &        \\
     &      & MARS  & 78.74/62.26     &        \\
     \hline\noalign{\smallskip}
     42  & QDNet, 2017, \cite{chen2017beyond} & CUHK-01 & 81.0/-- & No \\
     &      & CUHK-03  & 75.53/--     &        \\
     &      & VIPeR  & 49.05/--     &        \\
     \hline\noalign{\smallskip}
     43  & DGD, 2016, \cite{xiao2016learning} & CUHK-01 & 66.6/-- & Yes \\
     &      & CUHK-03  & 75.3/--     &        \\
     &      & VIPeR  & 38.6/--     &        \\
     &      & iLIDS  & 64.6/--     &        \\
     &      & PRID-2011  & 64.0/--     &        \\
     \hline\noalign{\smallskip}
     44  & UMDL, 2016, \cite{peng2016unsupervised} & CUHK-01 & 27.1/-- & Yes \\
     &      & VIPeR  & 31.5/--     &        \\
     &      & iLIDS  & 49.3/--     &        \\
     &      & PRID-2011  & 24.2/--     &        \\
     \hline\noalign{\smallskip}
     45  & SAL, 2015, \cite{shi2015transferring} & CUHK-01 & 22.4/-- & No \\
     &      & CUHK-03  & 29.3/--     &        \\
     \hline\noalign{\smallskip}
     46  & 4S-Net, 2020, \cite{khatun2020joint} & Market-1501 & 91.6/75.7 & No \\
     &      & DukeMTMC-ReId  & 82.4/77.3     &        \\
     &      & VIPeR  & \textbf{71.4}/--     &        \\
     &      & iLIDS  & 84.8/--     &        \\
     \hline\noalign{\smallskip}
     47  & PL-Net, 2019, \cite{yao2019deep} & Market-1501 & 88.2/69.3 & No \\
     &      & CUHK-03  & 82.75/--     &        \\
     &      & VIPeR  & 56.65/--     &        \\
     \hline\noalign{\smallskip}
     48  & kLDFA, 2016, \cite{chen2016deep} & CUHK-01 & 57.28/-- & No \\
     &      & VIPeR  & 38.37/--     &        \\
     \hline\noalign{\smallskip}
     49  & CGAN-TM, 2020, \cite{tang2020cgan} & Market-1501 & 64.43/31.33 & No \\
     &      & DukeMTMC-ReId  & 54.85/32.85     &        \\
     \hline\noalign{\smallskip}
     50  & GPP-ReID, 2021, \cite{feng2021complementary} & Market-1501 & 90.6/78.6 & No \\
     &      & DukeMTMC-ReId  & 81.3/67.9     &        \\
     &      & MSMT-17  & 53.5/24.6     &        \\
     \hline\noalign{\smallskip}
     51  & CI-CNN, 2020, \cite{song2019context} & Market-1501 & 94.26/89.54 & No\\
     &      & DukeMTMC-ReId  & 87.6/81.3     &       \\
     &      & MARS  & 87.3/78.8     &        \\
     \hline\noalign{\smallskip}
     52  & DRM, 2021, \cite{dai2021dual} & Market-1501 & 90.9/78.0 & No\\
     &      & DukeMTMC-ReId  & 82.1/67.7     &       \\
     &      & MSMT-17  & 55.0/26.9     &        \\
     \hline\noalign{\smallskip}
     53  & HCC-GCNs, 2021, \cite{bai2021hierarchical} & Market-1501 & 91.2/78.9 & No\\
     &      & DukeMTMC-ReId  & 81.2/67.5     &       \\
     &      & MSMT-17  & 57.4/28.4     &        \\
     \hline\noalign{\smallskip}
     54  & IIA, 2020, \cite{fu2020improving} & Market-1501 & 98.2/96.0 & No\\
     &      & DukeMTMC-ReId  & 94.4/91.8     &       \\
     &      & CUHK-03 (Labeled)  & 84.93/86.58     &    \\
     &      & CUHK-03 (Detected)  & 80.14/82.72     &    \\
     \hline\noalign{\smallskip}
     55  & VOLTA, 2020, \cite{liu2020iterative} & MARS & 66.7/51.9 & No\\
     &      & DukeMTMC-VideoReId  & 89.2/85.9     &       \\
     \hline\noalign{\smallskip}
     56  & MMFA-AAE, 2021, \cite{lin2020multi} & VIPeR & 58.4/-- & No\\
     &      & MSMT-17  & 46.0/20.7     &       \\
     &      & iLIDS  & 84.8/--     &       \\
     \hline\noalign{\smallskip}
     57  & SAL, 2020, \cite{jiang2020self} & Market-1501 & 68.7/41.9 & No\\
     &      & DukeMTMC-ReId  & 70.8/51.4     &       \\
     &      & CUHK-03 (Labeled)  & 31.0/35.7     &    \\
     &      & CUHK-03 (Detected)  & 28.7/34.4     &    \\
     \hline\noalign{\smallskip}
     58  & PREST, 2021, \cite{zhang2021self} & Market-1501 & 82.5/62.4 & No\\
     &      & DukeMTMC-ReId  & 74.4/56.1     &       \\
     \hline\noalign{\smallskip}
     59  & MLOL, 2021, \cite{sun2021unsupervised} & Market-1501 & 86.6/70.9 & No\\
     &      & DukeMTMC-ReId  & 83.1/69.8     &       \\
     &      & MSMT-17  & 48.3/22.4     &       \\
     \hline\noalign{\smallskip}
     60  & CCAN, 2020, \cite{zhou2020cross} & Market-1501 & 94.6/87.0 & Yes\\
     &      & DukeMTMC-ReId  & 87.2/76.8     &       \\
     &      & CUHK-03 (Labeled)  & 75.2/72.9     &    \\
     &      & CUHK-03 (Detected)  & 73.0/70.7     &    \\
     &      & MSMT-17  & 76.3/53.6     &       \\
     \hline\noalign{\smallskip}
     61  & JPIL, 2020, \cite{zhao2020joint} & Market-1501 & 73.5/48.2 & No\\
     &      & DukeMTMC-ReId  & 74.7/55.8     &       \\
     \hline\noalign{\smallskip}
     62  & CDL, 2018, \cite{li2017person} & CUHK-01 & 78.17/-- & Yes\\
     &      & VIPeR  & 66.39/--     &       \\
     &      & iLIDS  & 45.1/--     &       \\
     \hline\noalign{\smallskip}
     63  & DECAMEL, 2020, \cite{yu2020unsupervised} & Market-1501 & 60.24/32.44 & No\\
     &      & CUHK-01  & 65.81/--     &       \\
     &      & CUHK-03  & 38.27/--     &    \\
     &      & MSMT-17  & 90.34/11.13     &       \\
     \hline\noalign{\smallskip}
     64  & UTAL, 2020, \cite{li2020unsupervised} & Market-1501 & 69.2/46.2 & No\\
     &      & DukeMTMC-ReId  & 62.3/44.6     &       \\
     &      & CUHK-03  & 56.3/42.3     &    \\
     &      & MSMT-17  & 31.4/13.1     &       \\
     &      & MARS  & 49.9/35.2     &       \\
     &      & iLIDS  & 35.1/--     &       \\
     \hline\noalign{\smallskip}
     65  & HAN, 2019, \cite{li2020scalable} & Market-1501 & 94.2/83.4 & No\\
     &      & DukeMTMC-ReId  & 80.6/64.1     &       \\
     &      & CUHK-03 (Labeled)  & 46.5/46.1     &    \\
     &      & CUHK-03 (Detected)  & 47.5/45.5     &    \\
     &      & MSMT-17  & 60.1/32.6     &       \\
     \hline\noalign{\smallskip}
     66  & SBSGAN, 2021, \cite{huang2021unsupervised} & Market-1501 & 87.9/80.0 & No\\
     &      & DukeMTMC-ReId  & 79.7/71.5     &       \\
     \hline\noalign{\smallskip}
     67  & 3D-GAT, 2021, \cite{zhang20213d} & Market-1501 & 94.1/81.5 & No\\
     &      & DukeMTMC-ReId  & 85.5/71.2     &       \\
     \hline\noalign{\smallskip}
     68  & pLMNN, 2018, \cite{zhao2018person} & CUHK-01 & 53.5/-- & No\\
     &      & VIPeR  & 46.5/--     &       \\
     \hline\noalign{\smallskip}
     69  & SADL, 2020, \cite{li2020structure} & Market-1501 & 60.7/26.6 & No\\
     &      & DukeMTMC-ReId  & 50.2/28.1     &       \\
     &      & CUHK-03   & 75.42/--     &    \\
     &      & MSMT-17  & 30.6/--    &       \\
     \hline\noalign{\smallskip}
     70  & UDAM, 2020, \cite{song2020unsupervised} & Market-1501 & 75.8/53.7 & Yes\\
     &      & DukeMTMC-ReId  & 68.4/49.0     &       \\
     \hline\noalign{\smallskip}
     71  & LADL, 2020, \cite{li2019attribute} & CUHK-01 & 57.67/-- & No\\
     \hline\noalign{\smallskip}
     72  & IVD-ReID, 2019, \cite{zhang2018learning} & MARS & 48.0/-- & No\\
     &      & iLIDS  & 65.0/--     &       \\
    \end{longtable}
    }
 
 Table \ref{table:SOTAresultsOnEachDataset} comprehensively shows the SOTA results achieved on each dataset.
\begin{table*} 
\caption{SOTA results obtained on each challenge against each dataset.}
\label{table:SOTAresultsOnEachDataset}       
\begin{center}
    \resizebox{\textwidth}{!}{%
    \begin{tabular}{l l l l l l}
    \hline\noalign{\smallskip}
    Sr.No & Dataset & SOTA R1-results & Paper cited & Challenge addressed & Venue\\
    \hline\noalign{\smallskip}\hline\noalign{\smallskip}
    1 & Market-1501 & 98.3 & \cite{ning2020feature} & Background & TCSV-2020 \\
    2 & DukeMTMC-ReID & 94.7 & \cite{ning2020feature} & Background & TCSV-2020 \\
    3 & CUHK01 & 98.73 & \cite{qian2020leader} & Scale & PAMI-2020 \\
    4 & CUHK03 & 97.3 & \cite{shen2021person} & Pose & PAMI-2021 \\ 
    5 & VIPeR & 71.4 & \cite{khatun2020joint} & Generalization & CVIU-2020 \\
    6 & MSMT-17 & 87.7 & \cite{zhang2021part} & Misalignment & PR-2021 \\
    7 & PRID-2011 & 95.9 & \cite{hu2021hypergraph} & Pose & PR-2021 \\
    \hline\noalign{\smallskip}\hline\noalign{\smallskip}
    8 & MARS & 91.5 & \cite{wang2021pyramid} & Occlusion & ICCV-2021 \\
    9 & DukeMTMC-Video ReID & 98.3 & \cite{wang2021pyramid} & Occlusion & ICCV-2021 \\
    10 & iLIDS & 92 & \cite{he2021dense}(arxiv) & Misalignment & ICCV-2021 \\
    \hline\noalign{\smallskip}
    \end{tabular}
    }
\end{center}
\end{table*}

\section{Discussion \& Future Trends}
\label{sec:discussion}

In this systematic review, 230+ articles are reviewed that were published from January 2015 till October 2021 focused on the challenges faced for person re-id. In all these papers, the specified challenges have been addressed and the achieved state-of-the-results have been summarized in this review article. We have grouped the articles into few categories and have critically analysed their impact on the obtained results. The limitations along with the datasets used in the published articles have also been reviewed against each challenge. 

\subsection{Impact of Automated Person Re-id on Society}
\label{sec:impactofreid}
Generally, due to the scarcity of security resources as well as the lack of technological advancements in third world countries, the traditional law and order system does not meet the needs of the public, hence failing to build the people’s confidence. The automated surveillance systems \emph{i.e.} person re-id aims to improve the quality of the lives of common people by providing a sustainable living environment to them. Since the assurance of security and implementation of law and order are from the basic needs of human beings, the person re-id solutions can assist the law enforcement authorities in providing enhanced preemptive security and quick implementation of the law and order. Moreover, in case of any adverse happening, the re-id solutions can greatly assist the security officials in rapid response and quick resolution of the security issues and can prevent the delays caused by manual video forensic analytic. In order to trace people in a camera network, faces can only be used for recognition only if the subject is close enough and facing towards the camera. But usually in CCTV footage this is not the case, people are captures at variant poses and viewpoints where their faces are not clear. Therefore, the features found in a person's entire body (like clothing, height etc) are more useful to identify a person across different cameras of the network.

\subsection{Deep Learning Conjecture}
\label{sec:discussionOverview}

Since the evolution of deep learning methods in 2015, the computer vision research community immediately shifted from hand-crafted machine learning research to the deep learning based algorithms. Soon after the availability of a very large scale vision based dataset \emph{i.e.} ImageNet and its pre-trained weights, like many other research domain, deep person re-id solutions came into existence. Meanwhile, the medium to large scale re-id benchmarks were proposed so that the specialized custom re-id  solutions could be developed. Beginning from the transfer learning methods using the generic deep architectures like AlexNet, ResNet \emph{etc}, the re-id research quickly got independence in designing much sophisticated solutions due to the availability of large scale person re-id benchmarks. For many years, the convolution neural networks served as a strong backbone for re-id solutions. The CNNs are used to learn a variety of person representations,\emph{i.e.} the global person representations, local parts based person representations, semantics based person representations, attributes driven representations \emph{etc}. The CNN architectures with single stream/ branch was common in the start, however with the passage of time, multi-streamed architectures are proposed for person re-id, where each stream targets a different perspective.

Later on, with the development of attention based mechanism for vision problems, the same were extensively explored to develop the re-id solutions. Most of the attention based re-id solutions are developed on the backbone of CNNs and are multi-steam architectures to capture the various types of attention features \emph{i.e.} spatial attention, temporal attention, channel wise attention etc. The attention based re-id solutions performed really well for various re-id challenges in comparison with the methods that do not involve computation of the attention. 

Since the deep-learning is still evolving and new SOTA backbone architectures are being developed by the research community with every passing year, it drives the whole research community to new dimensions. Transformers and its variants are SOTA for the language problems since long but due to certain limitations, these were not used for vision problems in a holistic way. Recently, the development of vision transformer (in year 2021) was a great break-through in the vision research, and opened up the new ways in vision research. The vision transformer outperforms the counterpart CNN baseline deep architectures with a great margin for various vision problems. Since then, the re-id solutions with transformers backbone are gaining popularity among the research community.


This comprehensive study on person re-identification research unfolds few interesting aspects for the future re-id researches. While exploring the re-id solutions for the foremost common re-id challenge, i.e., the occlusion, it has been observed that the top three re-id solutions \cite{wang2021pyramid}, \cite{aich2021spatio}, \cite{eom2021video} that optimally re-identify the occluded persons are all attention based approaches. These attention based mechanisms are strengthened by the use of 3D convolutional architecture, pyramid architecture and the memory units. The different perspectives of the spatial and temporal features are learnt to capture the the dynamic and static information of a person. The temporal features compensate for the occluded spatial regions and enhance the performances of the re-id solutions.

The pose and view point variations make the person re-id quite challenging especially in case of inter-class differences (where people of different classes appear similar under different camera acquisitions) due to the similar appearances. This study highlights that for this particular challenge, both the CNN and attention based approaches are comparable. For instance, among the top three re-id solutions \cite{wang2021horeid}, \cite{zhou2020fine} and \cite{zhang2020relation}, only \cite{wang2021horeid} presented the approach which is based on learning attention weights while \cite{zhou2020fine} employed the higher level semantic information to generate multi-level feature maps and \cite{zhang2020relation} presented a novel Kronecker product matching operation to perform the re-identification. Generally, these approaches work to handle the misalignment by learning aligned and misaligned parts of person images. Local parts based dynamic feature learning addresses the misalignment issues either by focusing on motion specific and joint specific information in person images.  

The cluttered background, if not efficiently removed or suppressed, inversely effects the performance of re-id solution. Since the attention based mechanism inherently focus and highlight the attentive parts/ regions on a person image, the top three re-id solutions which perform the best even in case of cluttered background also employed attention based mechanisms to exclude the background information \cite{zhou2019discriminative} and learns multi-level attention in different ways \cite{sun2021memf} and fuse the information learnt from various levels and branches \cite{ning2020feature}. Generally, these approaches capture the foreground attention features using either the encoder decoder architecture or the multi-level attention modules. Learning the binary masks is common practice to exclude the cluttered background from the person foreground. However, these approaches do not aid the scenario where the background information might play vital role in the identification of a person.

The orientation of cameras and different viewing angles result in the misalignment of person images. To tackle Either the re-id solutions perform images alignment by proposing part-based re-id solutions or align the image regions by proposing various sophisticated algorithms. Top three approaches to address this issue are both the CNN based and the attention based. \cite{he2021dense}, an encoder-decoder based architectures, is a hybrid approach that takes the advantage of both CNN and attention based mechanism to handle the misalignment efficiently and outperformed all other solutions for video-based re-id benchmarks. \cite{chen2019abd} is an attention mechanism that handled the misalignment by learning the channel wise and position/spatial information and \cite{zhang2021part} handled the misalignment by part-guided graph convolution network. Generally, the local parts/ patches based learning in addition to global feature learning aids to address the misalignment issues in person re-id.

The differences in scales of captured images is generally seen in CCTV footage due to the variations in the distances of cameras from the targets. The scale differences are handled by various re-id solutions, however, the attention based approaches outperformed the rest of re-id solutions. The top three solutions either used the multi-scale attention pyramid \cite{qian2020leader} or divided the image into multiple local parts and then learnt the attention \cite{chen2021person}, or \cite{huang2020multiscale} extracted the holistc and local feature maps using multi-scale omni-bearing attention network. The re-id solutions that support multi-scale re-id learn the person features at multiple scales through multi-sized convolutional layers or branches. And then aggregate the learned information into final person descriptor. For video based benchmarks, the temporal information alleviate the multi-scale re-id solutions.  

An effective re-id solution handles the illumination variations efficiently such that the changes in the color of a person's dress due to variant lightening conditions may not result in false re-identification of an entity. Similarly the view-point variations need significant attention by the underlying re-id solution. The best performance is attained by an attention based mechanism \cite{zheng2019re} to address these re-id challenges. However, few other convolutions based architectures \cite{tao2017deep} and \cite{chen2021bidirectional} performed well to address the view-point variations. 

Since the surveillance cameras work 24/7 and capture the person images from a distance, generally the low image resolution is observed in CCTV footage. This results in another re-id challenge i.e. to identify a person correctly in low resolution images. Recently, \cite{feng2021resolution} proposed a resolution-aware re-id framework that works very well and follows the teacher-student learning mechanism.

Lastly, the cross-domain person re-identification is quite challenging with a huge room for improvement. Due to numerous re-id challenges, already discussed in previous sections, it is difficult to re-identify the people of totally different camera network. A well generalized re-id solution is highly desirable for cross-domain re-identification. A lot of research has been carried out to solve this problem and recently a generalized re-id solution \cite{fu2021unsupervised} used the temperature features to propose a contrastive learning framework. Another attention aggregation formulation was designed by \cite{fu2020improving} for a generalized re-id solution.

\subsection{Concluding Remarks}
\label{sec:futureResearch}

In general, we can emphasize that attention based re-id solutions are gaining more interest in the research community with their promising performances. Also, since the self-attention based vision research is just in its beginning phase, the full strength of these approaches are yet to be researched and analyzed.  

During the recent years, the re-id research is at its best for few initially proposed re-id datasets \emph{ i.e.} Market1501, DukeMTMC-Reid etc, which were captured from the public places with controlled environment, hence do not depict the real world scenario. The customized re-id algorithms addressed most of the re-id challenges effectively due to the medium level of complexity for these benchmarks. Among various re-id challenges, the pose variations remained most popular among research community, as it is the most common re-id issue with significant impact on the performance of re-id solutions. The newly proposed re-id benchmarks (\emph{i.e.} MSMT17 \emph{etc}) , captured from the complex scenes with a large number of indoor and outdoor cameras and closer to the real world complex scenarios, therefore these need more sophisticated re-id solutions to solve the real world problems.

Since the start of the re-id research journey, the majority of re-id solutions work for re-id of people from a single benchmark for unseen identities. However, far less research is done to propose the re-id solution doe cross-domain benchmarks. The major reason is that, the re-id research in still evolving and needs sophisticated research solutions even for the independent re-id benchmarks. For the recently proposed complex and large-scale re-id benchmarks, the performance of the re-id solutions need significant improvement.

\section*{Declarations}
The authors declare no conflict of interest.


\bibliography{main_article}
	\bibliographystyle{unsrt}


\end{document}